\documentclass[opre,nonblindrev]{informs3}

\OneAndAHalfSpacedXI

\usepackage{mystylefile}
\usepackage[size=small]{subcaption}
\usepackage{anyfontsize}
\usepackage{natbib}
 \bibpunct[, ]{(}{)}{,}{a}{}{,}%
 \def\bibfont{\small}%
\TheoremsNumberedThrough     
\ECRepeatTheorems

\EquationsNumberedThrough    

\MANUSCRIPTNO{MS-0001-1922.65}
\counterwithin{equation}{section}
\begin{document}



\RUNTITLE{Efficient Offline and Online Learning in Matching Markets}

\TITLE{Match Made with Matrix Completion: Efficient Learning under Matching Interference}

\ARTICLEAUTHORS{%
\AUTHOR{Zhiyuan Tang}
\AFF{Naveen Jindal School of Management, University of Texas at Dallas, 
\EMAIL{zhiyuan.tang@utdallas.edu}} 
\AUTHOR{Wanning Chen\thanks{Authors are listed in alphabetical order.}}
\AFF{Foster School of Business, University of Washington,  
\EMAIL{wnchen@uw.edu}} 
\AUTHOR{Kan Xu\footnotemark[1]}
\AFF{W. P. Carey School of Business, Arizona State University, 
\EMAIL{kanxu1@asu.edu}} 
} 

\ABSTRACT{Matching markets face increasing needs to learn the matching qualities between demand and supply for effective design of matching policies. In practice, the matching rewards are high-dimensional due to the growing diversity of participants. We leverage a natural low-rank matrix structure of the matching rewards in these two-sided markets, and propose to utilize matrix completion to accelerate reward learning with limited offline data. A unique property for matrix completion in this setting is that the entries of the reward matrix are observed with \emph{matching interference} --- i.e., the entries are not observed independently but dependently due to matching or budget constraints. Such matching dependence renders unique technical challenges, such as sub-optimality or inapplicability of the existing analytical tools in the matrix completion literature, since they typically rely on sample independence. In this paper, we first show that standard nuclear norm regularization remains theoretically effective under matching interference. We provide a near-optimal Frobenius norm guarantee in this setting, coupled with a new analytical technique. Next, to guide certain matching decisions, we develop a novel ``double-enhanced'' estimator, based off the nuclear norm estimator, with a near-optimal entry-wise guarantee. Our double-enhancement procedure can apply to broader sampling schemes even with dependence, which may be of independent interest. Additionally, we extend our approach to online learning settings with matching constraints such as optimal matching and stable matching, and present improved regret bounds in matrix dimensions. Finally, we demonstrate the practical value of our methods using both synthetic data and real data of labor markets.}


\maketitle

\section{Introduction}
\label{sec: intro}
Matching markets have become increasingly essential to facilitate the matching efficiency across many domains. 
For instance, freelance service platforms such as Upwork and Taskrabbit create new opportunities for businesses to secure temporary labor \citep{belavina2020matching}; U.S. emergency departments match physicians to patients to enable timely diagnosis and treatment \citep{gowrisankaran2023physician,jameson2025impact}; the charter school systems allocate students to schools by lotteries, offering opportunities to study education matching mechanisms \citep{deming2014using,chabrier2016can}; many host countries assign refugees across resettlement locations, creating new employment opportunities to support integration \citep{bansak2018improving}.
The majority of the matching markets exhibit a two-sided structure, e.g., jobs and workers in the labor market, or patients and physicians in the healthcare sector. For convenience, we use the aforementioned labor markets as our primary setting for illustration, and denote the two sides of the market as \emph{jobs} and \emph{workers} respectively.

While the matching literature typically focuses on matching policy design given a limited number of known matching qualities \citep{chen2023feature}, we consider the problem of learning many \emph{unknown} matching rewards through noisy \emph{offline} data, such as ratings of worker performance \citep{belavina2020matching} or quality measure of productivity \citep{kaynar2023estimating}. In practice, learning the matching qualities or rewards can be a high-dimensional problem, given the increasing variety of both job and worker types. For example, Upwork has more than a hundred job categories according to their website description; therefore, even tens of worker types result in thousands of matching qualities to learn in such a labor market. Consequently, learning accurate matching qualities to allow for precise downstream matching decisions generally requires a large number of matched samples, which are often not readily available.

\begin{figure}[htbp]
     \centering
     \begin{subfigure}[b]{0.38\textwidth}
         \centering
         \includegraphics[width=\textwidth]{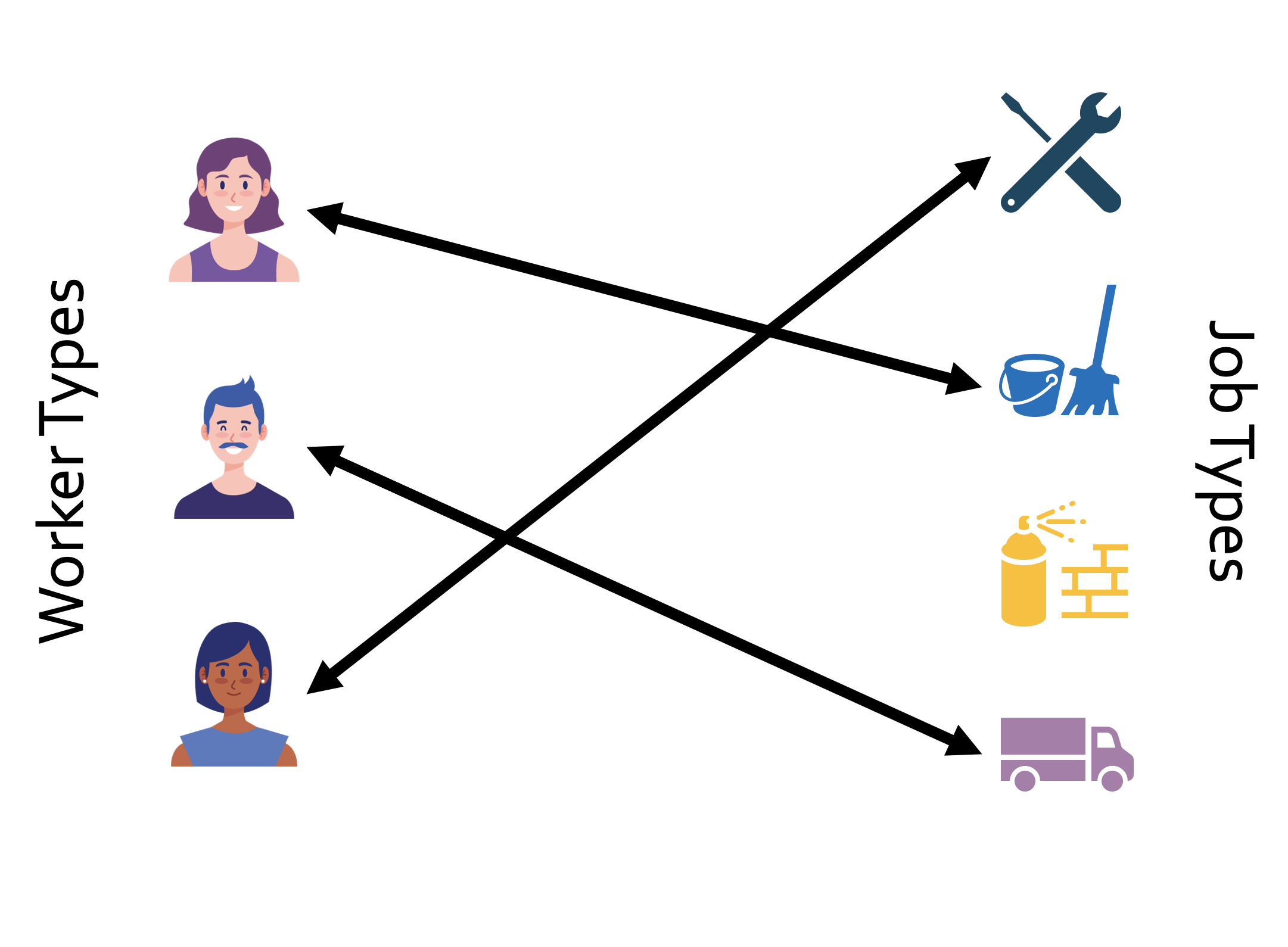}
         \caption{One-to-one Match}
         \label{fig:ex_match}
     \end{subfigure}
     \hspace{30pt}
     \begin{subfigure}[b]{0.38\textwidth}
         \centering
         \includegraphics[width=\textwidth]{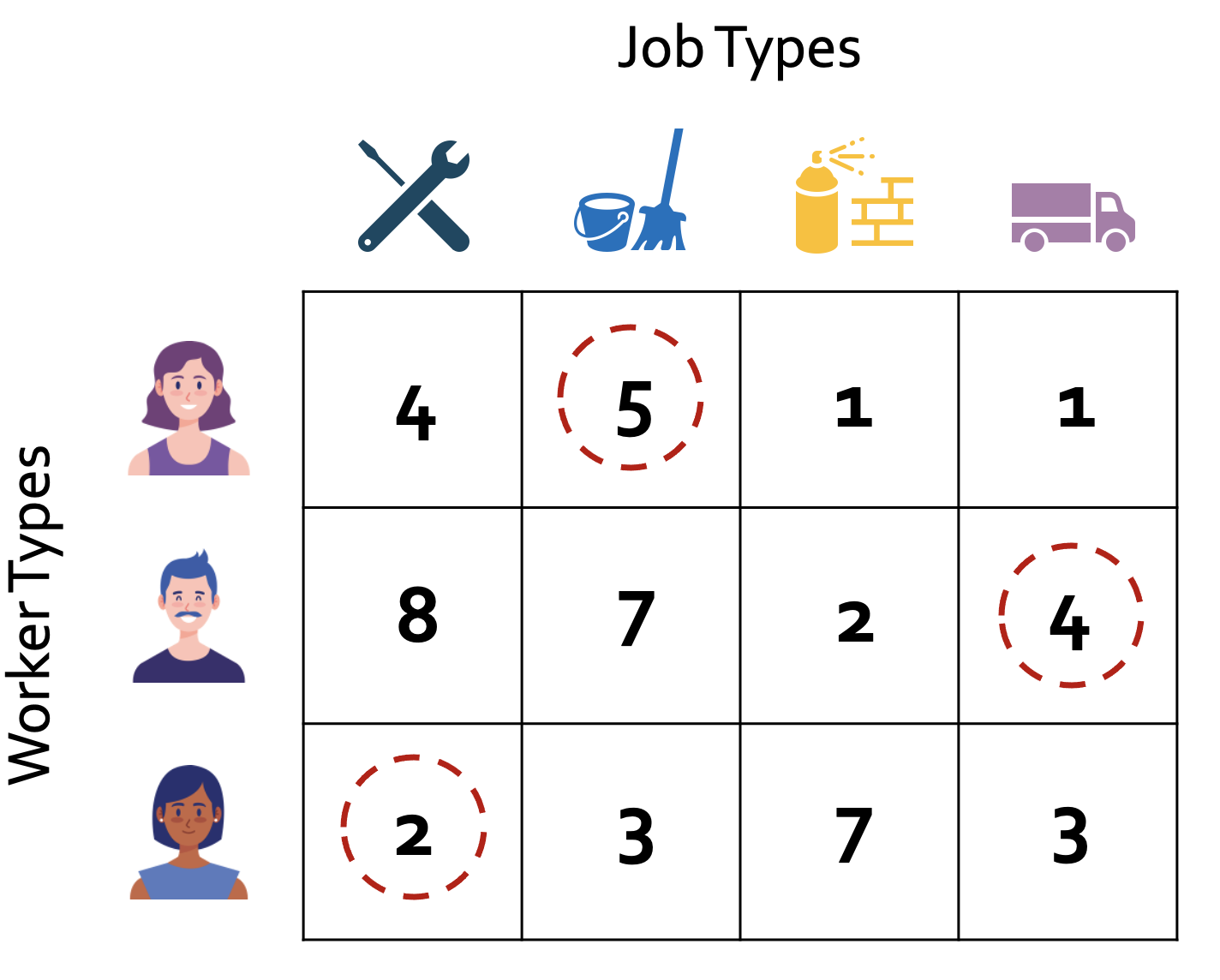}
         \caption{Matrix Representation}
         \label{fig:ex_matrix}
     \end{subfigure}
        \caption{Toy example of one-to-one matching for $3$ worker types and $4$ job types. (a) presents one matching of this labor market. (b) shows the matrix representation of the matching problem, where each entry denotes the reward of matching the corresponding worker and job pair; the matched pairs in (a) are circled in red. }
        \label{fig:ex_labor}
\end{figure}

In this paper, we propose to facilitate the learning process of these high-dimensional matching qualities using \emph{matrix completion}, specifically the state-of-the-art \emph{nuclear norm} regularization approach \citep{candes2008exact,candes2010matrix,negahban2012restricted}. The two-sided matching markets possess a natural matrix structure (see our illustrations in Figure \ref{fig:ex_labor}). Motivated by our data and existing applications across different domains {\citep{bell2007lessons,pennington2014glove,schuler2016discovering}}, we impose a \emph{low-rank} structure on the reward matrix, which enables us to learn a large reward matrix using only few matching outcomes. 
Formally, consider a matching market of $N$ worker types and $K$ job types with a reward matrix $\bTheta^* \in \mathbb{R}^{N\times K}$. Its $(i, j)^{\tth}$ entry $\bTheta^{*(i, j)}$ denotes the true reward of matching workers of type $i\in\{1, \cdots, N\}$ and jobs of type $j\in\{1, \cdots, K\}$. $\bTheta^*$ has a low rank $r=\rank(\bTheta^*)$ much smaller than the matrix dimensions $d=\max\{N, K\}$ (i.e., $r \ll d$); that is, there exist two latent feature matrices $\bU^* \in \mathbb{R}^{N\times r}$ and $\bV^*\in \mathbb{R}^{K\times r}$ such that $\bTheta^*= \bU^*\bV^{*\top}$. 
Intuitively, nuclear norm regularization (which we will introduce in Section \ref{sec: matrix_completion}) only requires few samples to identify the sparse set of matrix singular values accurately and hence estimate the low-rank matrix efficiently. 

\begin{table}[htbp]
\centering
\scalebox{0.92}{
\begin{tabular}{lccc}
\toprule
\textbf{Literature} & \textbf{\makecell[c]{Sampling \\ Assumption}} & \textbf{\makecell[c]{Guarantee \\ Type}} & \textbf{\makecell[c]{Error Rate \\ (under our setting)}} \\
\midrule
{Naive sample-average estimator} & N/A & Frob \& Inf & \makecell[l]{Frob: \(\widetilde{\mathcal{O}}\left({\sqrt{\frac{d^3}n}}\right)\), Inf: \(\widetilde{\mathcal{O}}(\sqrt{\frac{d}{n}})\)
}\\
\midrule
\makecell[l]{\cite{negahban2012restricted}\\ \cite{klopp2014noisy}}& \makecell[c]{Non-uniform \\ independent} & \multirow{2}{*}{Frob only} & \multirow{2}{*}{\makecell[c]{Adaptable\\Frob: \(\widetilde{\mathcal{O}}\left(\sqrt{\frac{rd^3}{n}}\right)\)}} \\
\cmidrule{1-2}
\cite{athey2021matrix} & \makecell[c]{Row-wise \\ independent} & & \\
\midrule
\makecell[l]{\cite{koltchinskii2011nuclear}\\ \cite{jang2024efficient}} & \makecell[c]{Non-uniform \\ independent, \\ known distribution} & Frob only & Not adaptable\\
\midrule
\cite{chen2020noisy} & i.i.d. & Frob \& Inf & Not adaptable \\
\midrule
\textbf{This paper} & \makecell[c]{Matching \\ dependent} & Frob \& Inf & \makecell[l]{Frob: \(\widetilde{\mathcal{O}}\left(\frac{rd}{\sqrt{n}}\right)\), Inf: \(\widetilde{\mathcal{O}}\left(\frac{r^2}{\sqrt{n}}\right)\)} \\
\bottomrule
\end{tabular}
}
\caption{Comparison of matrix completion literature.
The column of ``Assumption" lists the assumptions on the matrix entry sampling pattern; ``Guarantee Type" lists whether the error bounds are provided in Frobenius norm (``Frob") and/or entry-wise infinity norm (``Inf"); ``Error Rate (under our setting)" lists whether the analytical technique is adaptable to matching interference and, if so, the corresponding error rate. $d=\max\{N,K\}$ represents the matrix dimensions, $r$ is the matrix rank, and $n$ is the number of observed matchings. Note that the matrix rank is much smaller than the matrix dimension, i.e., $r\ll d$. The minimax lower bound for Frobenius norm is $\widetilde{\Omega}\left(\sqrt{\frac{rd^2}{n}}\right)$.}
\label{tab:matching-sampling-bound}
\end{table}

\textbf{Matching Interference.} However, we face unique technical challenges to establish optimal theoretical guarantees for nuclear norm regularization due to a distinctive property of the matching markets, which we call \emph{``matching interference"}. Matching interference refers to a dependent sampling pattern introduced by matching or budget constraints of the market participants. Particularly, the entries of the reward matrix are observed \emph{dependently}, instead of independently as in the literature \citep{candes2008exact,negahban2012restricted}. To illustrate, consider a one-to-one matching problem as in Figure \ref{fig:ex_labor}, which shows one matching of the market. The three observed entries, circled in red in Figure \ref{fig:ex_matrix}, are not independently observed, since the same worker cannot choose multiple jobs and vice versa --- i.e., there can be at most one observed entry in each row and column (detailed descriptions of our setting are provided in Section \ref{sec: problem_formulation}). Thus, matching interference introduces additional correlation among observations. 

Such matching dependent structure leads to either sub-optimality or inapplicability of the current analytical techniques in the nuclear norm regularization literature. We summarize the assumptions and results of the existing literature in Table~\ref{tab:matching-sampling-bound} for a comparison with our work. 

First, the typical analytical techniques \citep{negahban2012restricted,klopp2014noisy,athey2021matrix} all yield \emph{sub-optimal} error bounds in Frobenius norm regarding matrix dimension $d$ under our matching interference setting\footnote{Note that all these techniques only provide Frobenius norm guarantees for nuclear norm regularization, which describe the average accuracy over all matrix entries. We will discuss about the more difficult entry-wise guarantees in the infinity norm shortly.}. Particularly, these analyses lead to an additional factor of $\sqrt{d/r} (\gg 1)$ compared to the error rate derived using our techniques in our setting, resulting in rate no better than that of a naive entry-wise sample average estimator (see our Table~\ref{tab:matching-sampling-bound}). This performance gap arises because prior works predominantly rely on the independent sampling assumption and do not account for matching dependence. Alternatively, \cite{koltchinskii2011nuclear,jang2024efficient} propose new estimators but require the full knowledge of entry sampling distributions in addition to data independence. So our first research question is: \emph{can we show that nuclear norm regularization remains theoretically effective under matching interference?} 

Second, the more challenging entry-wise guarantee is required to inform certain matching decisions downstream (e.g., stable matching); however, the state-of-the-art leave-one-out technique {\citep{ma2018implicit,chen2020noisy,cai2022nonconvex}} is \emph{not adaptable} to our matching interference setting\footnote{Note that the nuclear norm estimator does not enjoy an entry-wise guarantee automatically given a Frobenius norm guarantee. Thus, additional analytical techniques such as leave-one-out is required for entry-wise guarantee.}. This is because these works heavily relies on the assumption of entry-wise i.i.d. sampling. Therefore, their analytical techniques cannot be extended to any data dependence structure like matching interference. So our second research question is: \emph{can we find a new estimator that provides a sharp entry-wise infinity norm guarantee even under matching interference?}

\textbf{Contributions.} Our work makes contributions to the matrix completion literature by providing the first optimal theoretical guarantees under matching interference and addressing these technical challenges through new analytical techniques. 

First, given the sub-optimality of existing techniques, we develop new analytical techniques to handle carefully the matching dependence; we prove that nuclear norm regularization can still achieve a near-optimal Frobenius norm bound under matching interference. Most proof techniques \citep[see, e.g.,][]{klopp2014noisy,hamidi2022low} rely on some contraction inequalities, which gives rise to deteriorated error rates in our setting. To overcome this issue, we propose a new linearization technique that leverages the sampling property of matching interference. We demonstrate that nuclear norm regularization can remain theoretically effective under dependent sampling schemes such as matching interference. 

Second, we design a novel ``double-enhanced" estimator atop the nuclear norm estimator, which provides a near-optimal entry-wise error bound in the infinity norm for the reward matrix under matching interference. Specifically, our double enhancement procedure estimates more precisely each row of the latent feature matrices $\bU^*$ and $\bV^*$ respectively --- and thus each entry of $\bTheta^*$ --- using linear regression based on the initial nuclear norm estimate of $\bTheta^*$. As such, our estimator provides the first theoretical result for entry-wise guarantee under matching dependence, while the existing techniques only work for independent sampling. We believe such a technique can also apply to a broader class of sampling distributions, which may be of independent interest.

Furthermore, we extend our theories to online learning problems with matching interference, such as optimal matching or stable matching \citep{gai2010learning,liu2020competing}. Analogous to our offline setting, our bandit algorithm leverages the low-rank reward structure to accelerate exploration. Particularly, we show that our algorithms can efficiently learn the reward matrix under matching interference. We derive regret upper bounds with improved dependence on the matrix dimensions in the data-poor regime for both optimal matching and stable matching problems. This highlights the strength of matrix completion in an online learning setting when data is limited, even under the more challenging matching dependence. 

Finally, we empirically evaluate the performance of our approaches using both synthetic and real data in the labor market. Our findings indicate that our approaches based on matrix completion significantly improve the learning process in the offline setting and facilitate the overall matching performance in the online scenario. 

\subsection{Related Literature}
\label{subsec:lit}

Our work relates to the literature of matrix completion, and bandits for matching problems. 

\paragraph{Matrix Completion.} Matrix completion involves learning missing entries of a matrix from a small sample of observed ones, particularly for high-dimensional data that naturally forms a matrix structure. Due to its efficiency, there has been significant interest from the operations research and machine learning communities in applying matrix completion to various domains, such as learning preference information in recommendation systems
{\citep{bi2017group,farias2019learning}},{learning user transition behaviors on digital platforms \citep{liu2022learning,jiang2024play}}, personalizing assortment planning \citep{kallus2020dynamic}, facilitating experimentation \citep{bayati2022speed,farias2022synthetically} and enhancing textual analytics \citep{xu2024group}. To the best of our knowledge, our paper makes the first investigation into adopting matrix completion for matching problems. 

However, the low-rank matrix completion theory faces a unique technical challenge in the matching setting, which we call ``matching interference". Typically, the matrix completion literature assume the entries of the unknown matrix $\bTheta^*$ are sampled with the same probability independently (i.e., uniformly at random) \citep{candes2008exact,candes2010power}. These studies prove that the nuclear norm regularization achieves a near-optimal error rate up to logarithmic terms in Frobenius norm. {\cite{negahban2012restricted,klopp2014noisy,hamidi2022low} }later generalize the results and consider the sampling schemes where the matrix entries are sampled with different probabilities but still independently (i.e., non-uniformly at random). \cite{koltchinskii2011nuclear, jang2024efficient} propose different estimators that require the knowledge of sampling distributions. \cite{athey2021matrix} consider a different sampling scheme with row independence given the specific panel data structure in a causal inference problem. In contrast, the matching interference problem in our setting introduces  unique data dependence of the entry observations, where we can at most observe one entry from each row and column in a one-to-one matching (see Figure~\ref{fig:ex_matrix}). As a result, existing proof techniques \citep{klopp2014noisy,athey2021matrix,hamidi2022low} lead to sub-optimal error bounds in Frobenius norm in our setting. Rather, we build on a linearization technique that leverages the sampling property of matrix completion, and show that we can achieve a near-optimal error bound with improved dependence on matrix dimensions (i.e., the number of worker/job types).

For certain downstream decision making problem such as stable matching, it is also important to quantify the statistical uncertainty of each entry estimate and provide entry-wise error control in \emph{$\ell_\infty$ norm}, which is a harder problem than controlling Frobenius norm error. \cite{chen2020noisy,chen2021bridging} are among the first to provide a near-optimal entry-wise error bound for matrix completion; however, their leave-one-out analyses rely on the assumption that the entries are sampled i.i.d., and thus do not apply to our sampling scheme with matching interference. \cite{hamidi2019personalizing} propose a row-enhancement technique for a different matrix factorization problem with contexts, which only provides row-wise error guarantees\footnote{Row-wise error control is simpler than entry-wise analysis, but harder than Frobenius norm error control.}. We develop a novel double-enhancement algorithm that improves the standard nuclear norm regularized estimator and ensures an entry-wise accuracy guarantee in our setting. 

\paragraph{Multi-Armed Bandits.} {The multi-armed bandit framework studies sequential decision-making problems in which optimal actions are learned through an exploration-exploitation trade-off, with broad applications in operations management \citep{chen2021dynamic,zheng2024online,simchi2025blind}}. Our online learning algorithms contribute to the bandit literature for matching problems such as optimal matching and stable matching. 
In the bandit problem for optimal matching, the goal is to learn the unknown rewards of matching workers with jobs over time and maximize the cumulative rewards of matched pairs. This can be formulated as a combinatorial semi-bandit problem, where each arm is one pair of worker and job types in our setting and a set of arms that form one matching is played simultaneously in each round \citep{gai2010learning,chen2013combinatorial,kveton2015tight}. 
\cite{chen2013combinatorial} and \cite{wang2018thompson} propose Combinatorial Upper Confidence Bound (\textsf{CUCB}) and Combinatorial Thomspon Sampling (\textsf{CTS}), adapted from the classic bandit literature \citep{auer2002finite,chapelle2011empirical}. \cite{kveton2015tight} improve the \textsf{CUCB} algorithm; they derive near-optimal upper bounds of their algorithm, matching the lower bounds. 
Recently there has also been growing interest in learning stable matching \citep{gale1962college} from bandit feedback in a centralized platform \citep{liu2020competing,jagadeesan2021learning}. \cite{liu2020competing} first introduce this bandit problem, where one side of the market (i.e., agents) has no prior knowledge about its ranking preference over the other side (i.e., arms) and needs to learn online. They propose an Explore-then-Commit (ETC) algorithm to minimize the agent-optimal stable regret --- i.e., to find an optimal policy that is a stable matching and optimal for the specific agent side. 
\cite{jagadeesan2021learning} consider a different matching with transfers problem \citep{shapley1971assignment}, and introduce a new notion of regret that captures the deviation of a market outcome from equilibrium. 
We adopt the bandit setting from \cite{liu2020competing} for our stable matching problem, as its regret definition aligns closely with the traditional bandit literature.

Our online learning bandit algorithms are also related to the low-rank bandit literature based on matrix completion/factorization \citep[see, e.g.,][]{jun2019bilinear, lu2020lowrank, jang2024efficient}. However, the existing literature study full-bandit feedback problem with a single observed reward each time, while our online learning part studies a semi-bandit feedback problem with a combinatorial nature defined by the matching set. Thus, their problems differ from ours, and do not involve the challenging matching interference structure. 
For instance, \cite{trinh2020solving,lattimore2021bandit} restrict their attention to rank-one reward matrix only. Particularly, \cite{trinh2020solving} fully exploit the rank-one structure and explore through learning the specific column and row of the leading entry; this approach cannot be generalized into general low-rank structure. \cite{lattimore2021bandit} consider a quadratic form reward model, which is equivalent to a rank-one symmetric reward matrix. Besides, their action space is the continuous unit sphere. \cite{lu2020lowrank} also employs a nuclear norm estimator, but their approach is similar to the offline matrix completion literature \citep{candes2008exact,negahban2012restricted} and requires independent sampling. \cite{kang2022efficient} propose an offline estimator analogous to \cite{koltchinskii2011nuclear} but require the exact knowledge of the sampling distribution and sampling independence over observations, while our matching interference setting involves sampling dependence and the sampling distribution of our offline data is unknown.

\paragraph{Dynamic Matching.} Our online learning part is also related to the dynamic matching literature in operations research that attempts to incorporate learning in a two-sided market \citep{massoulie2016capacity,bimpikis2019learning,shah2020adaptive,johari2021matching,hsu2022integrated}. The majority of this literature focus on learning unknown types of one market side (e.g., worker/job side), and thus requires knowledge of type-dependent matching rewards. \cite{hsu2022integrated} provide the first study with both unknown matching payoffs and unknown types of one side, but consider a different problem with queueing and do not aim to improve the dependence on market size (e.g., the number of worker/job types) as we do. \cite{saure2013optimal} propose an ETC algorithm and present improved regret dependence on market size (i.e., the number of products); however, they address a different demand learning problem in dynamic assortment and hence their technique cannot be extended to our setting. In contrast, we focus on improving the performance bounds regarding their dependence on market size given unknown rewards and known types of participants. 

\textbf{Outline.} The remainder of the paper is organized as follows. Section~\ref{sec: problem_formulation} introduces the problem setup for matrix completion under matching sampling. Section~\ref{sec: matrix_completion} establishes the Frobenius norm error bound for the nuclear norm regularization method and derives a corresponding minimax lower bound under matching interference. Section~\ref{sec:double-enhance} presents our double-enhanced estimator, which provides a near-optimal entry-wise error bound for our reward matrix in our setting. In Section~\ref{sec: online_learning}, we extend our theories to an online learning setting for stable and optimal matching, and establish regret bounds for the proposed bandit algorithms. Section~\ref{sec: experiments} presents empirical results using both synthetic and real-world data. 

\section{Problem Formulation}
\label{sec: problem_formulation}

This section formalizes the problem of learning the matching rewards between workers and jobs using offline matching data as a matrix completion problem. In Section \ref{subsec:matrix}, we formulate the unknown rewards as a low-rank matrix and introduce the observation model for this reward matrix. In Section \ref{subsec:sampling}, we highlight the key challenge of applying matrix completion to matching problems --- i.e., dependent entry-sampling due to matching interference, in contrast to canonical independent sampling.

\textbf{Notation.} We use regular capital letters for vectors, lowercase letters for scalars, and bold capital letters for matrices, unless otherwise specified. For any positive integer $k,$ let $[k]$ denote the index set $\{1, 2, \cdots, k\}$. For any vector $V$, let $V^{(i)}$ denote its $i^{\tth}$ entry, and $\|V\|$ denote its $\ell_2$ norm. 
For any matrix $\bTheta\in\mathbb{R}^{d_1\times d_2}$, we use $\bTheta^{(i,j)}$ to represent entry $(i,j)$ of a matrix $\bTheta$, $\bTheta^{(i, \cdot)}$ the $i^{\tth}$ row of the matrix, and  $\bTheta^{(\cdot, j)}$ the $j^\tth$ column. For a matrix $\bTheta$ of rank $r$, we denote its non-zero singular values by $\sigma_{\max}(\bTheta) = \sigma_1(\bTheta) \ge \sigma_2(\bTheta) \ge \cdots \ge \sigma_r(\bTheta) = \sigma_{\min}(\bTheta) > 0$, its Frobenius norm by $\|\bTheta\|_F = \sqrt{\sum_{i=1}^r \sigma_i^2(\bTheta)}$, its operator norm by $\|\bTheta\|_{\op} = \sigma_1(\bTheta)$, its nuclear norm by $\|\bTheta\|_*=\sum_{i=1}^r\sigma_i(\bTheta)$, its $\ell_{2,\infty}$ norm by $\|\bTheta\|_{2,\infty}=\max_{i\in[d_1]}\|\bTheta^{(i,\cdot)}\|$, and its vector $\ell_\infty$ norm by $\|\bTheta\|_{\infty}=\max_{i,j} |\bTheta^{(i,j)}|$. Given any two matrices $\bTheta,\widetilde\bTheta\in\mathbb{R}^{d_1\times d_2}$, we denote their trace inner product by $\langle\bTheta,\widetilde\bTheta\rangle=\sum_{i=1}^{d_1}\sum_{j=1}^{d_2}\bTheta^{(i,j)}\widetilde\bTheta^{(i,j)}$, and their Hadamard product by $\bTheta \circ \widetilde\bTheta \in\mathbb{R}^{d_1 \times d_2}$ with $(\bTheta \circ \widetilde\bTheta)^{(i, j)} = \bTheta^{(i, j)} \cdot \widetilde\bTheta^{(i, j)}$. Let $e_i(d)\in\mathbb{R}^d$ denote a basis vector with value 1 in its $i^{\tth}$ entry and 0 otherwise, i.e., $e_i(d)^{(j)}=1$ for $j=i$ and 0 otherwise.

\subsection{Matching in a Two-Sided Market}
\label{subsec:matrix}

Consider a two-sided online labor platform with $N$ available types of workers to be matched with $K$ unfilled types of jobs. We assume $N\leq K$ without loss of generality. The platform is centralized --- i.e., it has full control over job assignments. 

\textbf{Reward Matrix.} The true qualities or rewards of matching the worker and job sides can be naturally collected into a matrix form, according to the two-sided structure of the market (see, e.g., Figure \ref{fig:ex_labor}).
We use $\bTheta^*\in\mathbb{R}^{N\times K}$ to denote the reward matrix, where the worker types $i\in[N]$ and job types $j\in[K]$ define one dimension of the matrix respectively. Particularly, each row $i$ of the matrix corresponds to the matching rewards of one worker type $i$, and each column $j$ corresponds to one job type $j$; the value of each entry $(i, j)$, i.e., $\bTheta^{*(i, j)}$, indicates the expected reward the platform receives when a worker of type $i$ is matched with a job of type $j$, for $i\in[N]$ and $j\in[K].$ 

We make two assumptions on the reward matrix $\bTheta^*$, which are standard in the matrix completion literature \citep{koltchinskii2011nuclear, negahban2012restricted,klopp2014noisy,farias2019learning,chen2020noisy,athey2021matrix}. 
First, the true reward matrix $\bTheta^*$ is \emph{entry-wise bounded} by 1, i.e., $\|\bTheta^{*}\|_{\infty} \le 1$. Note that we choose an upper bound of 1 just for simplicity --- our results hold for any constant upper bound. 
Second, $\bTheta^*$ has \emph{low rank}; that is, the rank of the matrix $\bTheta^*$ has $\rank(\bTheta^*)=r\ll \min\{N, K\}$. In detail, there exist two low-dimensional matrices $\bU^* \in \mathbb{R}^{N\times r}$ and $\bV^*\in \mathbb{R}^{K\times r}$ such that $\bTheta^*= \bU^*\bV^{*\top}$. The matching reward of worker type $i$ and job type $j$ is then jointly determined by their latent features, i.e., $\bTheta^{*(i,j)} = \bU^{*(i, \cdot)} \bV^{*(j, \cdot)\top}$. Intuitively, low-rankness suggests that the true reward matrix depends on very few parameters, which helps reduce the number of parameters to learn from $NK$ of $\bTheta^*$ to $r(N+K)$ of $\bU^*$ and $\bV^*$.
\begin{remark}
\cite{udell2019big} demonstrate theoretically that any sufficiently large matrix has low-rank property in general, while typical matching markets face a large number of worker and job types (i.e., the matrix dimensions $N$ and $K$ are large) in practice. 
\end{remark}

\textbf{Matching.} We consider the \emph{one-to-one matching} scheme for simplicity, i.e., each worker can be matched with at most one job and vice versa. Our insight and results can be extended to the general many-to-many matching setting. Let $\mathbb{M}$ denote one such \emph{matching}, defined as a set of pairs of worker and job types:
\begin{align*}
    \mathbb{M} = \{(i,j(i)) \mid i\in[N];\, j(i) \ne j(i'),\, \forall i\ne i'\},
\end{align*}
where, with slight abuse of notation, we let $j(i)$ denote the job type matched with the $i^{\tth}$ worker type, and no two worker types $i\ne i'$ share the same job types $j(i) \ne j(i')$ and vice versa. Note that the platform clears the market in any matching. 

Equivalently, we can denote a matching $\mathbb{M}$ using a matrix $\bX\in\{0,1\}^{N\times K}$, where the $(i, j)^{\tth}$ entry $\bX^{(i,j)}$ takes value 1 if $(i,j)\in \mathbb{M}$ and $0$ otherwise. Any matching should belong to the following set of matchings
\begin{align}\label{eq:x_matchset}
\mathcal{M} = \left\{\bX\in \{0, 1\}^{N\times K} \,\middle|\, \sum_{j=1}^K \bX^{(i, j)} = 1,\forall i\in[N]; \sum_{i=1}^N \bX^{(i, j)} \le 1, \forall j\in[K]\right\}.
\end{align}
Particularly, $\sum_{i=1}^N \bX^{(i, j)}\le1$ for a job type $j$ since $j$ can be assigned to at most one worker type; in other words, the $j^{\tth}$ column of $\bX$, i.e., $\bX^{(\cdot, j)}$, contains at most one entry of 1 and elsewhere 0. Similarly, $\sum_{j=1}^K \bX^{(i, j)} = 1$ since the market clears and each worker type can take exactly one job type. When no ambiguity arises, we will use $\mathbb{M}$ and $\bX$ interchangeably in the subsequent sections. 

We further define the \emph{matched pair} of worker type $i$ in a matching $\bX$ using a superscript $i$
\begin{align}\label{eq:matchedpair}
\bX^i = e_i(N)e_{j(i)}(K)^\top \in \mathbb{R}^{N\times K},
\end{align}
that is, a basis matrix with the $(i,j(i))^{\tth}$ entry being 1 and 0 otherwise (recall that $e_i(d)\in\mathbb{R}^d$ is a basis vector with value 1 in its $i^{\tth}$ entry and 0 otherwise). By our definition, the matched pairs of all worker types make up the matching $\bX$:
\begin{align}\label{eq:matcheqv}
\sum_{i=1}^N \bX^i = \bX \in \mathcal{M}.
\end{align}

\textbf{Observation Model.}
Our offline matching data consists of $n$ observed matchings from the platform. For each matching indexed by $t\in [n]$, let $\mathbb{M}_t$ (and $\bX_t$) denote the matching  and let $(i,j_t(i))\in\mathbb{M}_t$ represent a matched pair of worker type $i$ and its corresponding job type.
Each observed matching $\bX_t$ is independently observed from the set $\mathcal{M}$ defined in \eqref{eq:x_matchset} according to a sampling distribution $\Pi_t$ over $\mathcal{M}$. Note that $\Pi_t$ can be heterogenous and non-uniform across $t\in[n]$; additionally, it is unknown. For every $(i,j)\in [N]\times [K],$ let $\pi_t^{(i,j)}=\mathbb{P}_{\bX_t\sim \Pi_t}\left\{\bX_t^{(i,j)}=1\right\}$ denote the probability of the $(i,j)$ pair being matched in matching $\bX_t.$ Similar to \cite{negahban2012restricted,klopp2014noisy}, we assume there exists a positive $\pmin$ such that $\pi_t^{(i,j)}\geq (\pmin K)^{-1},\forall t\in[T], \forall (i,j)\in [N]\times [K].$

For each matching $\mathbb{M}_t$, the platform observes only a noisy signal of the true reward $\bTheta^{*(i, j_t(i))}$ for any pair $(i,j_t(i))\in\mathbb{M}_t$. We concatenate the $N$ noisy rewards of the $N$ matched pairs in $\mathbb{M}_t$ into a vector $Y_t\in\mathbb{R}^N$, and denote the $N$ corresponding noises as $\varepsilon_t\in\mathbb{R}^N$. Specifically, the $i^{\tth}$ entry of $Y_t$, denoted by $Y_t^{(i)}$, corresponds to the observed reward of the pair $(i,j_t(i))$; $\varepsilon_t^{(i)}$ represents the unobserved noise of that pair. Then, each reward $Y_t^{(i)}$ of worker type $i$ in a matching $\mathbb{M}_t$ has
\begin{equation} \label{model: data-generating}
    Y_t^{(i)} = \langle\bX_t^i, \bTheta^*\rangle +\varepsilon_t^{(i)}
\end{equation}
for $t\in[\samplesize]$ and $i\in[N]$. The noises $\varepsilon_t^{(i)}$ are $\sigma$-subgaussian (see Definition \ref{def:subgaussian}) and independent across matchings $t\in[\samplesize]$; we want to mention that the noises are not required to be independent across matched pairs for $i\in[N]$ within the same matching $t\in[n]$.
\begin{definition}
    A random variable $X\in\mathbb{R}$ is $\sigma$-subgaussian if $\mathbb{E}[X]=0$ and $\mathbb{E}[\exp(sX)]\leq \exp\left(\frac{\sigma^2s^2}{2}\right),\, \forall s\in\mathbb{R}$.
    \label{def:subgaussian}
\end{definition}
To simplify notation, we define an \emph{observation operator} $\mathcal{X}_t: \mathbb{R}^{N\times K} \rightarrow \mathbb{R}^{N}$ such that
\begin{align} \label{def:obs-opt}
    \mathcal{X}_t(\bTheta)=\begin{bmatrix}
        \langle\bX_t^1,\bTheta\rangle,\cdots,\langle \bX_t^N,\bTheta\rangle
    \end{bmatrix}^\top
\end{align}
for any $\bTheta\in\mathbb{R}^{N\times K}$ and $t\in[\samplesize]$. 
\begin{remark}\label{rmk:xdep}
As we will discuss in Section \ref{subsec:sampling}, the entries of $\bTheta^*$ are not independently sampled, i.e., $\{\bX_t^i\}_{i\in[N]}$ are correlated given each $t\in[n]$ due to the one-to-one matching constraints. Thus, $\{\bX_t^i\}_{i\in[N], t\in[n]}$ are not independent. 
\end{remark}
 
\textbf{Estimation.} Our problem of recovering the unknown low-rank matrix $\bTheta^*$ using entry-wise observations is an instance of the matrix completion problem \citep{candes2008exact,candes2010power}. 
In total, we observe $nN$ samples for the model \eqref{model: data-generating} from $n$ matchings and $N$ matched pairs in each matching, i.e., $\{(\bX_t^i, Y_t^{(i)}) \mid t\in[\samplesize],i\in[N]\}$. 
We might face a high-dimensional problem where the sample size $nN$ could be much less than the total number of unknown parameters $NK$ in $\bTheta^*$, i.e., $n \ll K$. As a result, we propose to exploit the low-rank structure of $\bTheta^*$ and efficiently learn the unknown matrix using matrix completion. We describe the details of the estimation procedure in Section \ref{sec: nonconvex}. 
We measure the estimation accuracy of an estimator $\widehat{\bTheta}$ by its Frobenius norm, i.e., $\|\widehat{\bTheta} - \bTheta^*\|_F$, in Section \ref{sec: matrix_completion}. 
As aforementioned, we require an entry-wise error guarantee for certain downstream matching problems. Thus, we also measure an entry-wise estimation error of $\widehat{\bTheta}$ by its $\ell_\infty$ norm, i.e., $\|\widehat{\bTheta} - \bTheta^*\|_\infty$, in Section \ref{sec:double-enhance}.

\subsection{Sampling with Matching Interference}
\label{subsec:sampling}

A key technical challenge of our matrix completion problem is to establish optimal theoretical guarantees under entry sampling dependence due to interference in the matching markets. In contrast, the existing literature primarily rely on sampling independence to derive error bounds, and thus their techniques either yield sub-optimal results or are inapplicable to our setting (see Table~\ref{tab:matching-sampling-bound}). In the following, we describe our matching interference setting, and compare with two standard independent sampling schemes in the literature, i.e., independent sampling and independent row sampling. For simplicity, we take the one-to-one matching setting as an example; our argument applies to the many-to-many matching setting as well.

\begin{figure}[htbp]
     \centering
     \begin{subfigure}[b]{0.3\textwidth}
         \centering
         \includegraphics[width=\textwidth]{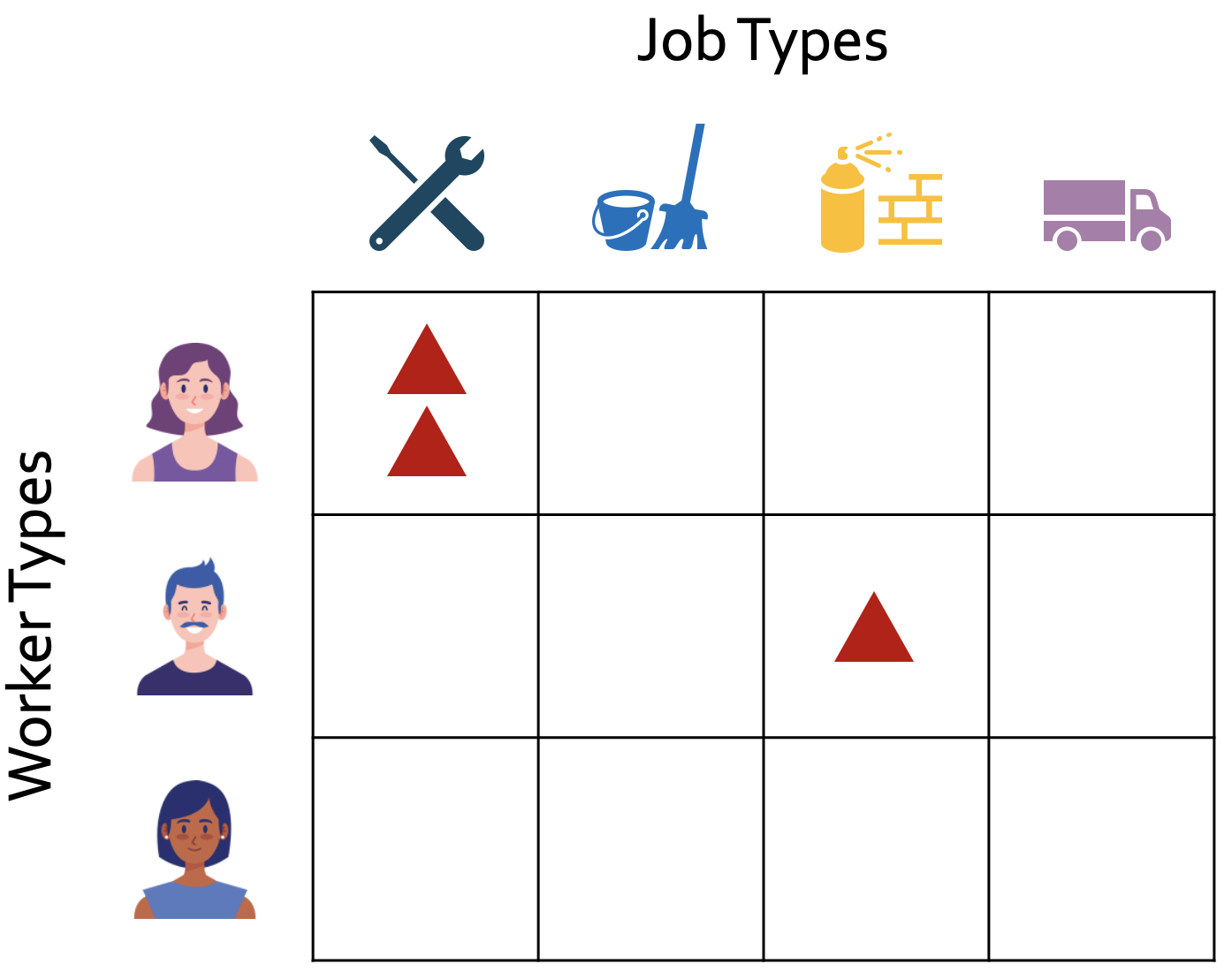}
         \caption{Independent Sampling}
         \label{fig:sample_iid}
     \end{subfigure}
     \begin{subfigure}[b]{0.3\textwidth}
         \centering
         \includegraphics[width=\textwidth]{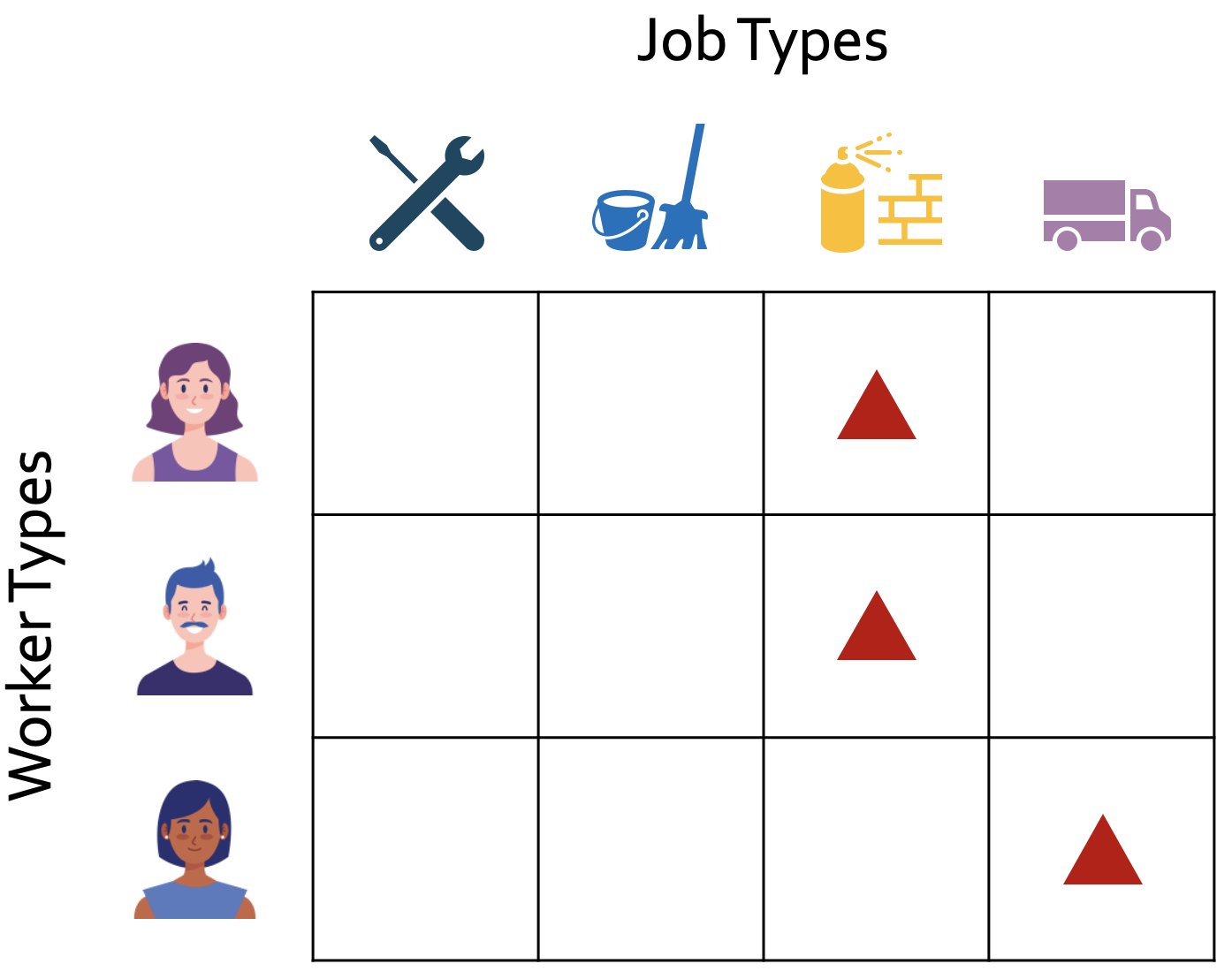}
         \caption{Indepedent Row Sampling}
         \label{fig:sample_row}
     \end{subfigure}
     \begin{subfigure}[b]{0.3\textwidth}
         \centering
         \includegraphics[width=\textwidth]{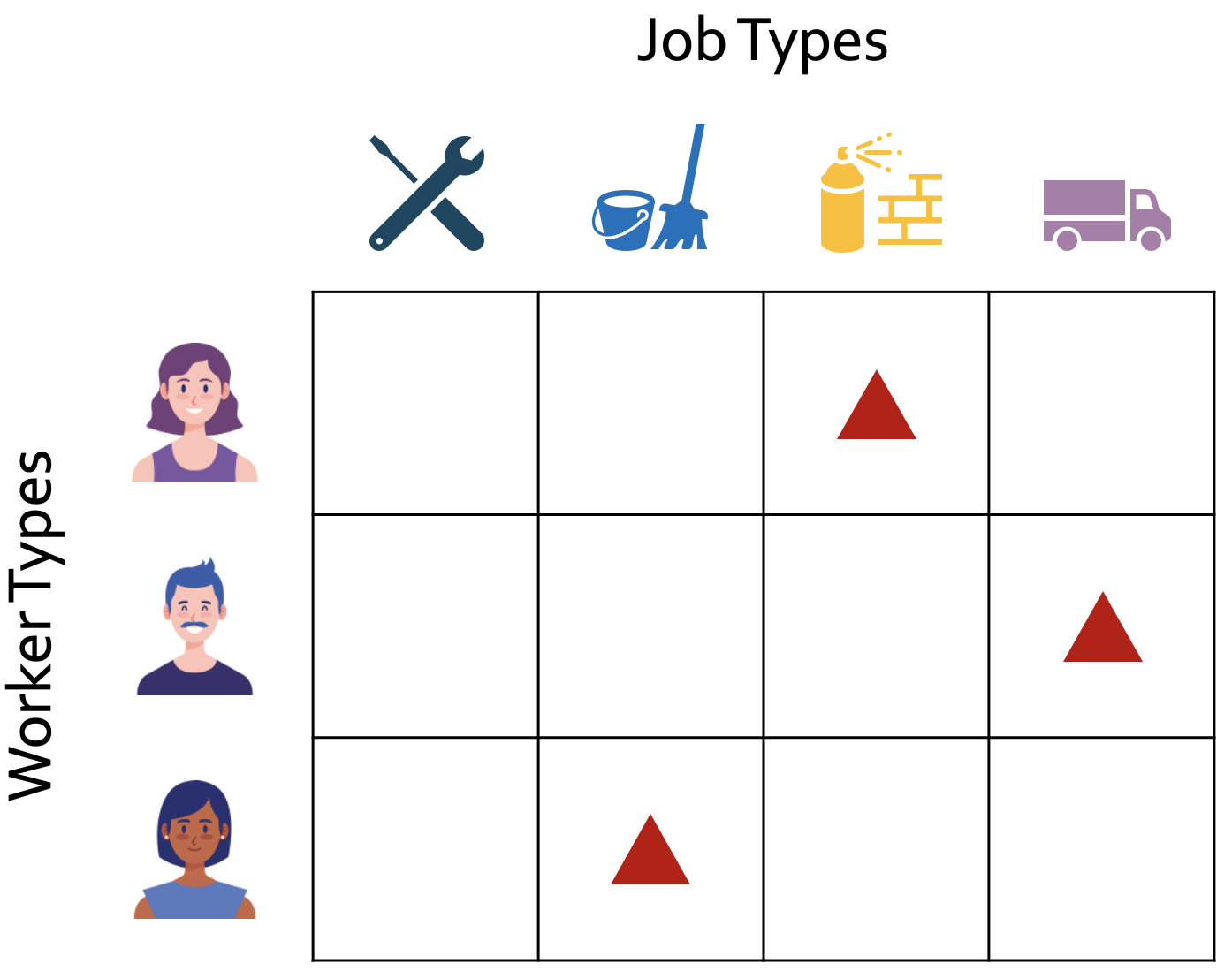}
         \caption{Matching Sampling}
         \label{fig:sample_match}
     \end{subfigure}
        \caption{Toy example of different sampling schemes for $N=3$ worker types and $K=4$ job types. One red triangle indicates that the corresponding entry is sampled once. (a) shows an independent sampling scheme, where all entries are sampled at random with replacement. (b) shows an independent row sampling scheme, where the column is selected at random for each row independently. (c) shows a dependent sampling scheme with matching interference in our matching setting. 
        }
        \label{fig:ex_sampling}
\end{figure}

\textbf{Independent Sampling.} Independent sampling is the most widely studied sampling scheme in the matrix completion literature. 
Typically, it refers to the setting where the entries of a matrix are sampled at random either uniformly \citep[see, e.g.,][]{candes2008exact,candes2010power} or non-uniformly \citep[see, e.g.,][]{negahban2012restricted,klopp2014noisy}. 
The majority of the literature derives an improved estimation error bound for an unknown low-rank matrix under this assumption, given its tractability. 
For instance, consider sampling uniformly at random in the observation model \eqref{model: data-generating}. In this case, the entry sampling or the matched pair $\bX_t^i$ has
\begin{align}\label{eq:sampleiid}
\bX_t^i = e_{\iota}(N)e_{j}(K)^\top,
\end{align}
where $\iota$ and $j$ are randomly sampled from the index sets $[N]$ and $[K]$ respectively for any $i\in[N]$ and $t\in[n]$. Note that here both $\iota$ and $j$ do not depend on the indices $i$ and $t$. Therefore, $\{\bX_t^i\}_{i\in[N], t\in[n]}$ are all independent from each other.

We illustrate that in Figure \ref{fig:sample_iid}; for simplicity, we consider the three samples of a specific $t$ but of three different $i$'s with $i\in[N]$ and $N=3$, i.e., $\{(\bX_t^i, Y_t^{(i)}) \mid i\in[3]\}$. 
Particularly, one red triangle indicates that the corresponding entry is sampled once (or the corresponding worker-job pair is matched once). 
Then, all entries of the matrix are sampled three times at random with replacement given our definition in \eqref{eq:sampleiid}; since the location of one red triangle does not affect that of another red triangle, the entries are sampled independently. Consequently, the same pair of worker and job types might be observed multiple times, while some worker or job types might not be matched through the sampling process. Apparently, independent sampling cannot guarantee a one-to-one matching for our matching setting. 

\textbf{Independent Row Sampling.} An alternative sampling scheme is the independent row sampling \citep{jain2022online,baby2024online}. Particularly, sampling takes place within each row of the matrix, and a column is randomly selected for each row independently. Now, consider independent row sampling in the observation model \eqref{model: data-generating}. In such a case, the entry sampling or the matched pair $\bX_t^i$ has
\begin{align*}
\bX_t^i = e_{i}(N)e_{j}(K)^\top,
\end{align*}
where $j$ is randomly sampled from the index set $[K]$ for any $i\in[N]$ and $t\in[n]$. Different from our example of independent sampling, there is no longer randomness in sampling rows for a given $i\in[N]$. Yet, $\{\bX_t^i\}_{i\in[N], t\in[n]}$ are still independent from each other, since the columns are uniformly sampled at random regardless of the row index $i$. The independent row sampling is closely related to the independent sampling, since the entry samplings or matched pairs $\bX_t^i$'s are still independent from each other.

This is illustrated in Figure \ref{fig:sample_row}. Similarly, we consider three samples of a specific $t$ but of three different $i$'s so matrix entries are observed exactly three times through $\{\bX_t^i\}_{i\in[N]}$. Under independent row sampling, 
the entries in each row are randomly sampled once; the entry sampling is independent across rows, since the column of one red triangle in a given row does not interfere with those of different rows. As a result, multiple worker types might be matched with the same job type, though no worker type remains unmatched. The independent row sampling still cannot ensure a one-to-one matching for our matching purpose. 

\textbf{Matching Sampling.} In contrast, the one-to-one matching constraints in our matching problem raise a unique challenge for adopting matrix completion techniques. Specifically, our matching sampling scheme introduces dependence of entry sampling or matched pairs within a matching due to matching interference. Recall that $\{\bX_t^i\}_{i\in[N], t\in[n]}$ are not independent from each other in our observation model \eqref{model: data-generating}. In particular, the $N$ matched pairs $\{\bX_t^i\}_{i\in[N]}$ that comprise any matching $\bX_t$ are correlated, since they have to satisfy
\begin{align*}
1 = \sum_{j=1}^K \bX_t^{(i, j)} = \sum_{j=1}^K \bX_t^{i(i, j)},\, \forall i\in[N]\,,
\quad \text{and}\quad 1 \ge \sum_{i=1}^N \bX_t^{(i, j)} = \sum_{i=1}^N \bX_t^{i(i, j)},\, \forall j\in[K]
\end{align*}
according to \eqref{eq:x_matchset} and \eqref{eq:matcheqv} (recall that the superscript $(i, j)$ indicates the entry $(i, j)$ of a matrix). Basically, these inequalities enforce the one-to-one matching constraints, ensuring that two worker types $i\ne i'$ cannot be matched with a same job type $j$ in the same matching $\bX_t$, i.e., $j_t(i) \ne j_t(i')$, and vice versa. Nevertheless, any matched pairs $\bX_t^i$ and $\bX_{t'}^{i'}$ across different matchings $\bX_t$ and $\bX_{t'}$ with $t\ne t'$ are still independent for any $i, i'\in[N]$, since the matchings are independently sampled from the distribution $\Pi$ given our model setup in Section \ref{subsec:matrix}. 

In Figure~\ref{fig:sample_match}, we present one possible matching $\bX_t$ of three observed entries or matched pairs given a specific $t$ and $i\in[N]$. The entries are sampled according to the one-to-one matching constraints; thus, the entry sampling is dependent in our case, as any two of the three red triangles cannot fall into the same row or column. In other words, no two worker types share the same job type and vice versa, and all worker types are fully occupied. 

In summary, our matching sampling scheme involves dependence across sampled entry observations due to matching interference, and is hence different from both the independent sampling and independent row sampling schemes in the prior literature. The current analytical techniques fully exploit the property of independence and thus induce sub-optimality in the error bounds or are not adaptable in our setting (see our Table~\ref{tab:matching-sampling-bound}). As a result, new proof techniques are required to achieve optimal sample efficiency for our matching sampling. In the next section, we first specify our matrix completion approach using nuclear norm regularization, and then provide a new analytical technique to demonstrate the optimality of matrix completion under matching interference.

\section{Matrix Completion for Offline Matching Data}
\label{sec: matrix_completion}

We first describe the standard matrix completion approach of nuclear norm regularization, and utilize it to estimate the reward matrix in our matching setting in Section \ref{sec: nonconvex}. Then, in Section \ref{sec:theorems}, we prove that this estimator can achieve a near-optimal error rate in Frobenius norm under dependent sampling with matching interference. Finally, we provide a minimax lower bound for our estimator to show its sample efficiency in Section \ref{subsec:lower}.
 
\subsection{Estimation via Nuclear Norm Regularization}
\label{sec: nonconvex}

We estimate the unknown reward matrix $\bTheta^*$ using the standard nuclear norm regularization. We note that one main focus in this paper is to provide a theoretical guarantee for nuclear norm regularization under matching interference; thus, the offline algorithm here is similar to those in the matrix completion literature. The nuclear norm regularization approach is defined as follows:
\begin{equation}
\label{program:nonconvex} 
\widehat{\bTheta} = \argmin_{\substack{\|\bTheta\|_{\infty}\leq 1,\,\rank(\bTheta)\leq r}}\bigg\{\frac{1}{\sizeone}\sum_{t=1}^{\sizeone}\|Y_t-\mathcal{X}_t(\bTheta)\|^2+\lambda\|\bTheta\|_*\bigg\},
\end{equation}
where $\mathcal{X}_t$ is the observation operator constructed with respect to the $N$ matched pairs $\{\bX_t^i\}_{i\in[N]}$ as defined in \eqref{def:obs-opt}, $Y_t$ is a vector of $N$ noisy rewards observed for the $N$ matched pairs in one matching $\mathbb{M}_t$, and $\lambda$ is a hyperparameter. In total, we have $nN$ samples of entries observed through $n$ matchings with $N$ matched pairs in each matching. We follow \cite{chen2015fast,ma2018implicit} and search our estimate in $\rank(\bTheta)\leq r$ to ensure that our estimate has rank no greater than $r$. Previously, \cite{chen2015fast,ma2018implicit} demonstrate that such a rank constraint can lead to improved estimation accuracy. Indeed, enforcing the low-rank structure is crucial for obtaining a near-optimal error rate in our Theorem \ref{theorem:errorbound}.

Our objective function in \eqref{program:nonconvex} consists of two parts. The first part measures how well we fit the observed data by calculating the mean squared error of their predictions. Its notation slightly departs from the literature since we calculate the loss in two steps -- we first calculate the estimation errors of the $N$ matched pairs within each observed matching, and then sum across all $n$ observed matchings. The second part is a nuclear norm penalty that provides additional regularization beyond the rank constraint, where the hyperparameter $\lambda$ regulates the degree of this penalty. Our main result in Theorem~\ref{theorem:errorbound} will provide a theoretically optimal value of $\lambda$ for minimizing the estimation error.

\subsection{Error Bound in Frobenius Norm}
\label{sec:theorems}

We provide an estimation error bound in Frobenius norm in Theorem~\ref{theorem:errorbound} for \eqref{program:nonconvex}. In essence, our result demonstrates that the standard nuclear norm regularization approach can still achieve an optimal performance even under a harder matching sampling scheme such as matching interference than the canonical independent sampling. We will further show its minimax optimality (up to logarithmic terms) in the upcoming Section \ref{subsec:lower}. 

\begin{theorem}\label{theorem:errorbound}
The estimator $\widehat{\bTheta}$ in \eqref{program:nonconvex} satisfies
\begin{equation*}
{\big\|\widehat{\bTheta}-\bTheta^*\big\|_F}
=\widetilde{\mathcal{O}}\left(\frac{rK}{\sqrt{{\samplesize}}}\right)
\end{equation*}
with probability at least $1-4\exp(-\alpha)$ for any $\alpha>0$, where
$\lambda=c_{\lambda} \sigma (\alpha+\log(N+K))/\sqrt{\sizeone}$ for a positive constant $c_\lambda$.
\end{theorem}
We provide the proof in Appendix \ref{sec: proof_of_f_norm}. We show that the nuclear norm regularized estimator yields an estimation error of $\widetilde{\mathcal{O}}(rK/\sqrt{n})$ in the Frobenius norm in our setting, which is near-optimal (see our lower bound in Theorem~\ref{theorem: lower_bound}). 
In contrast, the naive entry-wise sample average estimator obtains a worse rate of $\mathcal{O}(K\sqrt{N/n})$. Our error bound is comparable with the results under independent sampling; we have an additional factor of $\sqrt{r}$, which is insignificant given $r\ll \min\{N, K\}$, due to the complexity of matching interference. Our result shows that matrix completion can be especially helpful in the high-dimensional setting when $N$ and $K$ are much larger than the rank $r$. 

Yet, our Theorem~\ref{theorem:errorbound} does not trivially hold without new analytical techniques for matrix completion to handle matching interference. Intuitively, a key step in proving Theorem~\ref{theorem:errorbound} is to bound an error term that captures the degree of non-convexity of the loss \eqref{program:nonconvex}.
Specifically, this small error term is 
\begin{align}\label{eq:termsmall}
\mathbb{E}\bigg[\sup_{\bold{\Delta}\in \widetilde{\mathcal{C}}}\sum_{t=1}^{\sizeone}\xi_t (\sum_{i=1}^N\langle \bX_t^i,\bold{\Delta}\rangle^2)\bigg]
\end{align}
for some restricted set $\widetilde{\mathcal{C}}$ ($\subseteq\mathcal{C}_{\alpha}(r)$, defined in \eqref{eq: restricted}), where $\bX_t^i$ is defined in \eqref{eq:matchedpair} and $\{\xi_t\}_{t\in[n]}$ is a sequence of Rademacher random variables. The existing analytical techniques \citep[see, e.g.,][]{negahban2012restricted,klopp2014noisy,athey2021matrix} rely on a contraction inequality to bound \eqref{eq:termsmall}, which is well-suited for the independent sampling scheme but raises an additional $\sqrt{N}$ factor due to our matching dependent structure. As a result, applying their techniques to our setting leads to a worse Frobenius norm error bound of $\widetilde{\mathcal{O}}(K\sqrt{rN/n})$ by factor of $\sqrt{N/r}$ compared to our bound, as summarized in Table~\ref{tab:matching-sampling-bound}. To that end, we propose a new linearization technique to refine the upper bound of \eqref{eq:termsmall} and thus obtain a near-optimal guarantee. Intuitively, 
we leverage the sampling property of matching in matrix completion carefully and analyze an equivalent small term using Hadamard product
\[
\mathbb{E}\bigg[\sup_{\bold{\Delta}\in \widetilde{\mathcal{C}}} \frac{1}{\sizeone}\sum_{t=1}^{\sizeone} \xi_{t} \sum_{i=1}^N \langle \bX_t^i,\bold{\Delta} \circ \bold{\Delta}\rangle\bigg],
\]
where $\bold{\Delta}\circ\bold{\Delta}\in\mathbb{R}^{N\times K}$ is the Hadamard square of $\bold{\Delta}$ with $(\bold{\Delta} \circ \bold{\Delta})^{(i, j)} = (\bold{\Delta}^{(i, j)})^2$. We further manage to control this error term without introducing additional significant factors by using a property of Hadamard square, i.e., the Hadamard square of a low-rank matrix is also low-rank.
Our technique then provides a tighter guarantee for the convexity of our loss function in the following restricted strong convexity (RSC) result, which eventually leads to Theorem~\ref{theorem:errorbound}. 

Define the $L^2(\Pi)$ norm of any matrix $\bTheta\in\mathbb{R}^{N\times K}$ as 
\[
\|\bTheta\|_{L^2(\Pi)}=\sqrt{\mathbb{E}\left[\sum_{t=1}^T\sum_{i=1}^N \langle\bX_t^i,\bTheta\rangle^2\right]},
\]
where the expectation is taken over $\{\bX_t^i\}_{i\in[N]}$ defined in \eqref{eq:matchedpair} (recall that $\bX_t$ is sampled from $\Pi_t$), and $\Pi = \{\Pi_t\}_{t=1}^T$ denotes the collection of all sampling distributions. Then, we have the following proposition.

\begin{proposition}\label{lem:step3}
Define for any $\alpha>0$ 
\begin{align}
\label{eq: restricted}
    \mathcal{C}_{\alpha}(r)=\left\{\bold{\Delta}\in\mathbb{R}^{N\times K} \,\middle|\, \|\bold{\Delta}\|_{\infty}\leq 1,\|\bold{\Delta}\|_{L^2(\Pi)}^2>{{c_0N\alpha}},\rank(\bold{\Delta})\leq r\right\}.
\end{align}
Then, for any $\bold{\Delta}\in\mathcal{C}_\alpha(r),$ we have
\begin{equation}
\label{eq: rsc}
    \sum_{t=1}^{\sizeone}\sum_{i=1}^N\langle \bX_t^i, \bold{\Delta}\rangle^2 \ge c_2\|\bold{\Delta}\|_{L^2(\Pi)}^2
    -c_3\left({\pmin r^2K\log[(N+K)\sizeone]}\right)
\end{equation}
with probability greater than $1-\exp(-\alpha)$, where $c_0,c_2$ and $c_3$ are positive constants.
\end{proposition}

The proof is provided in Appendix~\ref{sec: proof_of_rsc}. Essentially, our RSC condition shows that the loss function in the first part of \eqref{program:nonconvex} is almost strongly convex, since the last term on the right hand side of \eqref{eq: rsc} is minor with moderate size of $n$. 
Intuitively, our matching sampling scheme, though limited by the one-to-one matching constraints, still captures and reveals a substantial proportion of the true matrix entries. 

\subsection{Minimax Lower Bound}
\label{subsec:lower}

We establish the minimax lower bound in Frobenius norm for our matrix completion approach for matching problems. We show that our nuclear norm regularized estimator is minimax optimal since its upper bound in Theorem \ref{theorem:errorbound} matches the lower bound we provide below in Theorem~\ref{theorem: lower_bound} up to insignificant terms. 

We first define the minimax risk of our problem in the Frobenius norm as 
\begin{align*}
\ell(\bTheta^*, \|\cdot\|_F) = \inf_{\widetilde{\bTheta}}\sup_{\bTheta^*\in\mathcal{C}}\mathbb{E}\left[\|\widetilde{\bTheta}-\bTheta^*\|_F \right], 
\end{align*}
where $\mathcal{C}=\{\bTheta\in\mathbb{R}^{N\times K} \mid \rank(\bTheta)=r,\|\bTheta\|_\infty\leq 1\}$ and the infimum ranges over all possible estimators. The minimax risk measures in principle the complexity of
estimating any unknown matrix $\bTheta^*$ satisfying our assumptions. We provide a lower bound for the minimax risk as follows.

\begin{theorem}\label{theorem: lower_bound}
The minimax risk of $\bTheta^*$ satisfies 
\begin{equation*}
\ell(\bTheta^*, \|\cdot\|_F) =\Omega\left(K\sqrt{\frac{r}{\samplesize}}\right).
\end{equation*}
\end{theorem}
We provide the proof in Appendix~\ref{sec: prrof_of_lower_bound}.
The proof strategy is similar to that of Theorem 3 in \cite{negahban2012restricted} and Theorem 5 in \cite{koltchinskii2011nuclear}, which provide a minimax lower bound under the independent sampling scheme. The result shows that the minimax lower bound of the matrix completion problem in our matching setting scales as $\Omega(K\sqrt{r/n})$. Compared to our upper bound in Theorem \ref{theorem:errorbound}, this lower bound suggests that our estimator in \eqref{program:nonconvex} is minimax optimal up to logarithmic terms and insignificant factors such as $\sqrt{r}$ ($\ll \min\{N, K\}$).

\section{Double-Enhancement Procedure}
\label{sec:double-enhance}

In this section, we show how to obtain a desired entry-wise guarantee, which is essential to some downstream matching decision-making processes such as online stable matching discussed in Section~\ref{sec: stable-matching}. We first design a double-enhancement procedure to produce an estimator with sharp entry-wise guarantee (Section \ref{sec: enhancement}). We then prove an entry-wise error bound for this enhanced estimator (Section \ref{subsec:entry-wise-analysis}), which is typically harder than the Frobenius error bound in Section \ref{sec: matrix_completion}. 

\subsection{Double-Enhancement Design}
\label{sec: enhancement}

As aforementioned, to guide certain downstream decision making such as stable matching, it is important to control the statistical uncertainty of the entry-wise estimates in $\ell_\infty$ norm. For instance, entry-wise estimates enable learning the preference rankings of the market sides, which can be crucial for decision making of matching discussed in Section \ref{sec: online_learning}. However, the nuclear norm estimators do not enjoy an entry-wise guarantee automatically given a Frobenius norm guarantee. To that end, we provide a double-enhanced estimator, atop the nuclear norm regularization, with an entry-wise guarantee. The procedure is summarized in Algorithm \ref{alg: db_enhancement}.

\begin{algorithm}\caption{Double-Enhancement}
\label{alg: db_enhancement}
\hspace*{\algorithmicindent} \textbf{Inputs:} $\lambda$ 
\begin{algorithmic}[H]
\State Set $\sizefirst=\lfloor \samplesize/2\rfloor$, $\mathcal{J}_1 = [\sizefirst]$, $\mathcal{J}_2 = [n]\setminus [n_1]$
\State Calculate $\widehat{\bTheta}$ in \eqref{program:nonconvex} using the data $\{(\bX_t^i, Y_t^{(i)}) \mid t\in\mathcal{J}_1,i\in [N]\}$
\State Compute the SVD $\nuclearest=\widehat{\bU}\widehat{\bD}\widehat{\bV}^{\top}$
\For {$i \in [N]$}
\State Set $\mathcal{R}_i=\{(\bX_{t}^i,Y_{t}^{(i)})\mid t\in\mathcal{J}_2\}$
\State Compute $\widetilde{\beta}_i = \argmin_{\gamma\in\mathbb{R}^{r}}\Big\{\sum\limits_{(\bX,y)\in \mathcal{R}_i} (y - \bX^{(i,\cdot)}{\nuclearestV}\gamma)^2\Big\}$
\EndFor
\For {each $j\in[K]$}
\State Set $\mathcal{C}_j=\{(\bX_{t}^i,Y_{t}^{(i)}) \mid t\in\mathcal{J}_2,i\in[N],j_t(i) = j\}$
\State Compute $\widetilde{\alpha}_j = \argmin_{\gamma\in\mathbb{R}^{r}}\Big\{\sum\limits_{(\bold{X},y)\in\mathcal{C}_j} (y-{\nuclearestU}^\top \bX^{(\cdot, j)}\gamma^{\top})^2\Big\}$ \label{alg_step: regression_U}
\EndFor
\State Let $\widetilde{\bU}=\begin{bmatrix}
\widetilde{\beta}_1 & \widetilde{\beta}_2 & \cdots &\widetilde{\beta}_N
\end{bmatrix}^\top$, 
$\widetilde{\bV}=\begin{bmatrix}
\widetilde{\alpha}_1 & \widetilde{\alpha}_2 & \cdots & \widetilde{\alpha}_K
\end{bmatrix}^\top$
\State Compute the SVD $\widetilde{\bU}=\bU_1\bold{D}_1\bold{Q}_1$
\State Compute $\enhancedest=\bU_1\bold{Q}_1\widetilde{\bV}^\top$
\end{algorithmic}
\hspace*{\algorithmicindent} \textbf{Outputs:} $\enhancedest$
\end{algorithm}

Particularly, our double-enhancement procedure builds on the nuclear norm–regularized estimator and hence achieves entry-wise error control through the following steps. First, we split the whole $n$ matchings into two subsets $\mathcal{J}_1$ and $\mathcal{J}_2$, where we estimate a nuclear norm regularized estimator ${\nuclearest}$ via \eqref{program:nonconvex} using the samples in $\mathcal{J}_1$. 
Next, we refine the estimation of the row and column spaces of $\bTheta^*$ alternatively using the remaining samples in $\mathcal{J}_2$. Let $\bTheta^*=\bU^*\bD^*\bV^{*\top}$ be the singular value decomposition (SVD) of the true matrix $\bTheta^*$ with $\bU^*$ and $\bV^*$ being two orthogonal matrices; then, the row and column spaces of $\bTheta^*$ refer to the subspace spanned by the columns of $\bV^*$ and columns of $\bU^*$ respectively.
Note that our matching model \eqref{model: data-generating}
\begin{align*}
Y_t^{(i)} = \langle \bX_t^i, \bTheta^*\rangle+\varepsilon_t^{(i)}
= \langle \bX_t^{i(i, \cdot)}, \bTheta^{*(i, \cdot)}\rangle+\varepsilon_t^{(i)}
= \underbrace{\bX_t^{i(i, \cdot)}\bV^{*}}_{\text{features}} \underbrace{(\bU^{*(i, \cdot)}\bD^*)^\top}_{\text{parameters}}  + \varepsilon_t^{(i)}
\end{align*}
can be represented by a standard linear regression model, where $\bX_t^{i(i, \cdot)}\bV^{*}$ is the feature vector and $\bU^{*(i, \cdot)}\bD^{*}$ is the unknown parameter vector. Since we have no direct access to $\bV^*$ in the feature vector, we approximate it with the orthogonal matrix $\nuclearestV$ from the SVD of the estimator $\widehat{\bTheta}=\widehat{\bU}\widehat{\bD}\widehat{\bV}^{\top}$. 
Now, we can enhance (i.e., first enhancement) the row space of the nuclear norm regularized estimate $\widehat\bTheta$ by estimating $\bU^{*(i, \cdot)}\bD^{*}$ using linear regression;
the corresponding least square estimate $\widetilde{\bU}$ enjoys a tighter row-wise guarantee in $\ell_{2, \infty}$ norm, i.e., $\|\widetilde{\bU}-\bU^*\bD^*\|_{2,\infty} \approx \|\widehat{\bTheta}-\bTheta^*\|_F/\sqrt{N}.$ 
Similarly, we can also enhance (i.e., second enhancement) the column space with an estimate $\widetilde{\bV}$ that satisfies $\|\widetilde{\bV}-\bV^*\bD^*\|_{2,\infty}\approx \|\widehat{\bTheta}-\bTheta^*\|_F/\sqrt{K}.$ 
Finally, our double-enhanced estimator $\widetilde{\bTheta}$ is built upon $\widetilde{\bU}$ and $\widetilde{\bV}$, obtained through our double-enhancement procedure. 
The specific design of $\widetilde{\bTheta}$ allows us to obtain an entry-wise error bound through the row-wise bounds of $\widetilde{\bU}$ and $\widetilde{\bV}$, considering that $\|\bold{A}\bold{B}^\top\|_{\infty} \le \|\bold{A}\|_{2,\infty}\cdot\|\bold{B}\|_{2,\infty}$ for any matrices $\bold{A}, \bold{B}$. In the next section, we provide a theoretical guarantee on the entry-wise error bound for $\widetilde{\bTheta}$;
we will show that the entry-wise error bound scales as $1/\sqrt{NK}$ of the Frobenius norm error $\|\nuclearest-\bTheta\|_F$ provided in Theorem \ref{theorem:errorbound}.

The work most closely related to ours is the row-enhancement design in \cite{hamidi2019personalizing}. However, our matching context differs from their problem in two aspects. First, \cite{hamidi2019personalizing} provide only row-wise error controls, whereas our application requires entry-wise error guarantees. This is considerably more challenging than row-wise or Frobenius norm error bounds in matrix completion \citep[see, e.g.,][]{chen2020noisy}. To that end, we need to not only simultaneously enhance both the row and column spaces, but also establish a new matrix perturbation inequality that connects subspace estimation errors to entry-wise deviations. This is distinct from the techniques used in the prior literature such as \cite{hamidi2019personalizing}.
Second, we consider a matrix completion problem, where our observation operator in \eqref{def:obs-opt} only reveals entry-wise information of matching rewards. In contrast, \cite{hamidi2019personalizing} consider a matrix factorization problem for contextual bandits; their observation operator builds on Gaussian random vectors and thus provides substantially richer information. Such observation operator difference makes our analysis different from and harder than theirs. Particularly, their analytical technique specifically applies to Gaussian contexts and leads to sub-optimal rates in our setting. To attain optimal error rates, our approach explicitly leverages the low-dimensional structure of $\bU^*$ and $\bV^*$ and performs convergence analysis on $\bU^*$ and $\bV^*$. This is neither employed nor required for the Gaussian setting considered by \cite{hamidi2019personalizing}.

\subsection{Entry-Wise Estimation Error Bound}
\label{subsec:entry-wise-analysis}

Our entry-wise estimation error bound of the double-enhanced estimator $\widetilde{\bTheta}$, introduced in Section \ref{sec: enhancement}, holds under a standard \emph{spikiness} condition in the matrix completion literature \citep{negahban2012restricted,hamidi2022low}. 
\begin{assumption}[Spikiness]\label{assump: spikiness}
    There exists a constant $\eta\geq 1$ such that
    \begin{equation*}
        \frac{\sqrt{NK}\|\bTheta^*\|_\infty}{\|\bTheta^*\|_F}\leq \eta.
    \end{equation*}
\end{assumption}

The spikiness condition is standard in the literature, which captures the impact of entry-wise large values on sample complexity for matrix completion.
Intuitively, the spikiness condition excludes matrices with overly large values in a few entries. If an unknown matrix is spiky, i.e., it has a few entries with extremely large values, then we cannot accurately estimate all matrix entries without observing all the entries \citep{candes2010matrix,negahban2012restricted,hamidi2022low}. For example, consider a matrix $\bTheta^*$ with $\bTheta^{*(1, 1)}=1$ and other entries 0. This matrix does not satisfy our assumption since $\sqrt{NK}\|\bTheta^*\|_\infty/\|\bTheta^*\|_F=\sqrt{NK}$, which cannot be bounded by a constant. Note that it is impossible to recover the entry $\bTheta^{*(1, 1)}$ and hence maintain small entry-wise error without observing rewards from this entry. 
In other words, the spikiness condition ensures the entry-wise identifiability of the matrix given any random data samples. 

\begin{remark}
The spikiness condition is related to another popular incoherence condition \citep{candes2008exact, candes2010matrix,chen2020noisy}, defined as
\begin{align*}
\|\bU^*\|_{2,\infty}\leq \mu \sqrt{\frac{{r}}{N}}, \quad\|\bV^*\|_{2,\infty}\leq\mu \sqrt{\frac{{r}}{K}},
\end{align*}
for some constant $\mu$ in our setting, where $\bU^*$ and $\bV^*$ are from the SVD $\bTheta^*=\bU^*\bD^*\bV^{*\top}.$ Intuitively, the incoherence condition implies that all rows of $\bU^*$ and $\bV^*$ are of similar scales, and thus prevents $\bTheta^*$ from being spiky.
We note that our results still hold under the incoherence condition.
\end{remark}

Next, we state the following result of an entry-wise error bound of the double-enhanced estimator ${\enhancedest}$ under the spikiness assumption. 

\begin{theorem}\label{theorem: second_stage_error_bound}
Suppose $n=\widetilde{\Omega}\left(\max\{(r^4K)/N, (K/N)^2\}\right)$.
The double-enhanced estimator ${\enhancedest}$ in Algorithm~\ref{alg: db_enhancement} satisfies
\begin{align*}
\|{\enhancedest}-\bTheta^*\|_\infty   
&=\widetilde{\mathcal{O}}\left(r^2\sqrt{\frac{K}{N\samplesize}}\right),
\end{align*}
with probability at least $1-(3N+3K+5)\exp(-\alpha)$ for any $\alpha>0$, where $\lambda=c_\lambda (\alpha+\log(N+K))/\sqrt{\sizeone}$ for some positive constant $c_\lambda$.
\end{theorem}

The proof is provided in Appendix~\ref{sec: proof_of_second_stage}. According to Theorem~\ref{theorem: second_stage_error_bound}, the entry-wise estimation error $\|{\enhancedest}-\bTheta^*\|_{\infty}$ for the double-enhanced estimator ${\enhancedest}$ in our setting is of order $\widetilde{\mathcal{O}}(r^2\sqrt{K/(nN)})$. In comparison, the nuclear norm regularized estimator $\widehat{\bTheta}$ has no entry-wise guarantee other than a trivial one using Theorem \ref{theorem:errorbound}, i.e., $\|{\nuclearest}-\bTheta^*\|_\infty \le \|{\nuclearest}-\bTheta^*\|_F = \widetilde{\mathcal{O}}(rK/\sqrt{n})$. Thus, our double-enhancement design saves us a significant factor of $\sqrt{NK}/r$ ($\ll 1$) in the entry-wise error control.
As aforementioned in Section \ref{subsec:lit}, some existing literature, such as \cite{chen2020noisy,chen2021bridging}, have also derived an entry-wise error bound. However, their approach exploits the unique property of the independent sampling scheme, and thus does not apply to our setting with sampling interference. Our double-enhancement procedure in Algorithm~\ref{alg: db_enhancement} might be of independent interest; it can be readily applied to existing estimators and obtain entry-wise guarantees in broader sampling schemes, including the independent sampling scheme in \cite{chen2020noisy,chen2021bridging}.

\begin{remark}
We note that our entry-wise error bound in Theorem~\ref{theorem: second_stage_error_bound} also matches its minimax lower bound up to logarithmic terms and insignificant factors. The proof of an entry-wise minimax risk is similar to that of our Theorem \ref{theorem: lower_bound}.
\end{remark}

\section{Online Learning in the Matching Market}\label{sec: online_learning}

In this section, we extend our offline matrix completion approach to the online learning setting, where a centralized platform needs to learn from adaptively collected data and make sequential matching decisions with no prior information. We discuss both the optimal matching (Section \ref{sec: online_optimal}) and the stable matching (Section \ref{sec: stable-matching}). 
We propose two algorithms respectively that speed up the learning process by reducing exploration cost, which can be especially useful under short horizons and in large matching markets. 

\subsection{Online Optimal Matching}
\label{sec: online_optimal}

We first discuss an online optimal matching problem, which is usually formulated as a combinatorial semi-bandit problem in the literature \citep{gai2010learning,chen2013combinatorial,kveton2015tight}.

\paragraph{Problem Formulation.} Analogous to our offline setting in Section \ref{subsec:matrix}, we consider a two-sided matching platform with $N$ worker types and $K$ job types, and $\bTheta^*\in\mathbb{R}^{N\times K}$ represents their reward matrix. In each step $t$ of a time horizon $T$, the platform chooses a matching (an arm) $\pi_t = \bX_t$ from the set of all matchings $\mathcal{M}$ (defined in \eqref{eq:x_matchset}) based on all historical information, and receives noisy rewards $Y_t$ from all $N$ matched pairs (defined in \eqref{model: data-generating}). We want to learn the unknown reward matrix $\bTheta^*$ and maximize the total expected reward in each matching. We compare our policy $\pi$ to an optimal matching $\bX^*$ that obtains the maximum total reward among all matchings, i.e., $\bX^*=\argmax_{\bX\in\mathcal{M}}\langle\bX, \bTheta^*\rangle$\footnote{Given $\bTheta^*$, $\bX^*$ can be efficiently calculated by many well-established algorithms such as Hungarian algorithm \citep{kuhn1955hungarian} and Munkres algorithm \citep{munkres1957algorithms}.}.
Our goal is to learn a policy $\pi$ to minimize the cumulative regret over time
\begin{equation}
\label{eq: regret_optimal}
R(T)=\sum_{t=1}^T \Big(\langle \bX^*,\bTheta^*\rangle-\langle \bX_t,\bTheta^*\rangle\Big)\,,
\end{equation}
where $\langle \bX^*,\bTheta^*\rangle-\langle \bX_t,\bTheta^*\rangle$ is the regret at time $t$.

\paragraph{Algorithm Design.} We propose a \textbf{Comb}inatorial \textbf{L}ow-\textbf{R}ank \textbf{B}andit (\textsf{CombLRB}) algorithm in  Algorithm~\ref{alg: lowrank_bandit} for the online optimal matching problem. \textsf{CombLRB} exploits the low-rank structure of the reward matrix $\bTheta^*$ to accelerate reward learning in the exploration phase; it incorporates the nuclear norm regularization approach formulated in Section \ref{sec: nonconvex}. 
Our algorithm has a two-stage design of exploration and exploitation.
First, it explores for $E_h$ time periods in the early stage. For each $t\in [E_h]$, it draws a matching $\bX_t$ following the uniform distribution $\Pi$ over $\mathcal{M}$ (defined in Section \ref{subsec:matrix}). 
Then, at the end of the exploration, we calculate a nuclear norm regularized estimator $\widehat{\bTheta}$ based on the data we have collected in the exploration phase, i.e., $\{(\bX_t^i, Y_t^{(i)}) \mid t\in[E_h], i\in[N]\}$. Our algorithm identifies a matching $\bX_c\in\mathcal{M}$ that maximizes the total reward using ${\nuclearest}$ as a surrogate for $\bTheta^*$, i.e., $\bX_c=\argmax_{\bX\in\mathcal{M}}\langle\bX, \nuclearest\rangle$. Finally, our algorithm commits to the matching $\bX_c$ and keeps playing this arm for the remaining time periods.

\begin{algorithm}\caption{Combinatorial Low-Rank Bandit (\textsf{CombLRB})}\label{alg: lowrank_bandit}
\hspace*{\algorithmicindent} \textbf{Inputs:} $E_h$, $\lambda$
\begin{algorithmic}[H]
\For{$t\in [E_h]$}
    \State Choose matching $\pi_t=\bX_t \sim \Pi$
    \State Observe rewards $Y_t^{(i)} = \langle\bX_t^i, \bTheta^*\rangle +\varepsilon_t^{(i)}$ for all $i\in [N]$
\EndFor
\State Calculate $\widehat{\bTheta}$ in \eqref{program:nonconvex} using the data $\{(\bX_t^i, Y_t^{(i)}) \mid t\in[E_h],i\in [N]\}$
\State Compute $\bX_c =\argmax_{{\bX}\in\mathcal{M}}\langle \bX, \widehat{\bTheta}\rangle$
\For{$t \in [T]\setminus [E_h] $}
\State Choose matching $\pi_t = {\bX}_c$
\EndFor
\end{algorithmic}
\end{algorithm}

Our next theorem provides a regret upper bound on our \textsf{CombLRB} Algorithm in the online optimal matching setting. 

\begin{theorem}\label{theorem: optimal_matching_bandit}
The regret of \textsf{CombLRB} in Algorithm~\ref{alg: lowrank_bandit} has
\begin{equation*}
\mathbb{E}[R(T)]=\widetilde{\mathcal{O}}\left(r(N+K)T^{2/3}\right),
\end{equation*}
where $E_h= \mathcal{O}(rT^{2/3})$ and $\lambda=c_{\lambda}\sigma\log(N(N+K)T)/\sqrt{E_h}$ for some positive constant $c_\lambda$.
\end{theorem}

We provide a proof in Appendix~\ref{app: proof_regret_optimal}. We compare the regret bound of our approach with that of a state-of-the-art \textsf{CUCB} algorithm \cite{kveton2015tight}. \textsf{CUCB} builds on the upper confidence bound insight from the bandit literature and estimates each matrix entry using its sample average individually. In other words, \textsf{CUCB} does not exploit the low-rank structure of the rewards and thus cannot gain efficiency through information sharing as we do. This approach yields a regret upper bound of $\widetilde{\mathcal{O}}(\sqrt{N^2KT})$. 
In contrast, our \textsf{CombLRB} algorithm achieves an improved regret bound for short time horizon $T$ or large market with large values of $N$ and $K$. Specifically, when $T=\mathcal{O}(K^3)$, our bound is strictly better than that of \textsf{CUCB} since we obtain a favorable dependence on the matrix dimensions $N$ and $K$. Indeed, the low-rankness of the reward matrix facilitates the reward learning and hence the matching decisions through only few explorations. It can be especially helpful for short horizons or high-dimensional contexts, when reward learning is very costly. 

It is also worth noting that our algorithm can run for any extremely short time horizon $T < K$ as well, while \textsf{CUCB} requires at least $K$ initial random matchings for exploration and only works for relatively long time horizons.

\subsection{Online Stable Matching}
\label{sec: stable-matching}

Next, we discuss an online stable matching problem; our problem setting is the same as that formulated by \cite{liu2020competing}.

\paragraph{Problem Formulation.} Consider a two-sided platform with $N$ worker types and $K$ job types. Unlike all previous settings, now the matrix $\bTheta^*$ represents the rewards of the worker side, which further implies the worker preference rankings over jobs. Particularly, its $(i,j)^{\tth}$ entry $\bTheta^{*(i, j)}$ denotes the reward received by a worker of type $i$ if they are matched with a job of type $j$. Then, the preference ranking of worker type $i$ over the $K$ job types is determined by the $i^{\tth}$ row of $\bTheta^*$; that is, they prefer a job of type $j$ over $j'$ if $\bTheta^{*(i, j)}>\bTheta^{*(i, j')}$. We encode the job side preferences in columns of another matrix $\bPhi^*\in \mathbb{R}^{N\times K}$. The platform initially does not know the worker rewards (i.e., $\bTheta^*$), hence has to learn their preferences online, whereas job preferences (i.e., $\bPhi^*$) are known in advance. 

At each time $t$ of a time horizon $T$, the platform chooses a matching $\pi_t = \bX_t\in\mathcal{M}$ and observes the noisy rewards $Y_t$ received by the $N$ worker types formulated by \eqref{model: data-generating}. We want to learn the unknown matrix $\bTheta^*$ of worker preferences and find the worker-optimal stable matching $\bX^*$; particularly, $\bX^*$ is the stable matching returned by the Gale-Shapley (GS) Algorithm \citep{gale1962college} when the workers are the proposing side. We note that $\bX^*$ is optimal among all stable matchings $\mathcal{S}$ ($\subseteq\mathcal{M}$) for all worker types \citep{knuth1997stable}, i.e., $\langle\bX^{*i}, \bTheta^*\rangle \ge \langle\bX^{i}, \bTheta^*\rangle, \forall\, \bX\in\mathcal{S}$ with $\bX^i$ defined in \eqref{eq:matchedpair}. We design a policy $\pi$ to minimize the worker-optimal stable regret
\begin{equation}\label{def: stable_regret}
R_i(T)=\sum_{t=1}^T\Big(\langle \bX^{*i},\bTheta^*\rangle-\langle \bX_t^i,\bTheta^*\rangle\Big),
\end{equation}
for every worker $i\in[N]$, where $\langle \bX^{*i},\bTheta^*\rangle-\langle \bX_t^i,\bTheta^*\rangle$ is the regret for workers of type $i$ at time $t$. 

\paragraph{Algorithm Design.} We develope a \textbf{Comp}eting \textbf{L}ow-\textbf{R}ank \textbf{B}andit (\textsf{CompLRB}) algorithm as in Algorithm~\ref{alg: lowrank_competing_bandit}. The design of \textsf{CompLRB} closely follows our \textsf{CombLRB} algorithm for online optimal matching but with two distinctions. Note that the regret \eqref{def: stable_regret} captures the entry-wise value difference of $\bTheta^*$, and thus can be bounded more tightly using the double-enhanced estimator $\widetilde{\bTheta}$ from our double-enhancement procedure in Section \ref{sec: enhancement}. Additionally, since our goal is to find an optimal policy among stable matchings, our algorithm identifies a worker-optimal stable matching $\bX_c$ using GS algorithm based on $\widetilde{\bTheta}$ and $\bPhi^*$. 

\begin{algorithm}
\hspace*{\algorithmicindent} \textbf{Input:} $E_h$, $\lambda$, $\bPhi^*$
\begin{algorithmic}[H]\caption{Competing Low-Rank Bandit (\textsf{CompLRB})}
\label{alg: lowrank_competing_bandit}
\For{$t \in [E_h] $}
\State Choose matching $\pi_t=\bX_t\sim \Pi$
\State Observe rewards $Y_t^{(i)} = \langle\bX_t^i, \bTheta^*\rangle +\varepsilon_t^{(i)}$ for all $i\in [N]$
\EndFor
\State Calculate $\widetilde{\bTheta}$ in Algorithm~\ref{alg: db_enhancement} using the data $\{(\bX_t^i, Y_t^{(i)}) \mid t\in[E_h],i\in [N]\}$ \label{alg_step: enhance_stable}
\State Compute a stable matching $\bX_c$ via GS algorithm with inputs ${\enhancedest}$ and $\bPhi^*$
\For{$t \in [T]\setminus [E_h] $}
\State Choose matching $\pi_t = {\bX}_c$
\EndFor
\end{algorithmic}
\end{algorithm}

We state the regret upper bound of our \textsf{CompLRB} algorithm for online stable matching as follows.
\begin{theorem}\label{theorem: stable_bandit} 
Let $\Delta_{\min} = \min_{i\in[N]}\{\min_{j\not =j'}|\bTheta^{*(i,j)}-
\bTheta^{*(i,j')}|\}$.
Then, the regret of \textsf{CompLRB} in Algorithm~\ref{alg: lowrank_competing_bandit} for worker type $i$ has
\begin{equation*}
\mathbb{E}[R_i(T)]=\mathcal{O}\left(\frac{r^3K\max\{\log^2[(N+K)T], r\log[(N+K)T]\}}{N\Delta_{\min}^2}\right),
\end{equation*}
where
$E_h=\mathcal{O}\left(\frac{r^3K\max\{\log^2[(N+K)T], r\log[(N+K)T]\}}{N\Delta_{\min}^2}\right)$ and $\lambda=c_{\lambda} \sigma \log\big[(N+K)(3N+3K+5)T\big]/\sqrt{E_h}$ for some positive constant $c_\lambda$. 
\end{theorem}

We provide a proof in Appendix \ref{app: proof_regret_stable}. In comparison, \cite{liu2020competing} provide an upper bound of $\mathcal{O}(K\log(NT)/\Delta_{\min}^2)$ for any worker type $i\in[N]$. Our regret bound improves upon that in \cite{liu2020competing} in the matrix dimensions $N$ and $K$ by a factor of at least $N/r^4$ up to logarithmic terms (recall that $N \gg r$). Specifically, our algorithm exploits the tightness of the double-enhanced low-rank estimator on entry-wise errors, and achieves an improved performance through reduced explorations. Instead, the algorithm in \cite{liu2020competing} estimates every entry of $\bTheta^*$ using a naive sample average estimate. Our theoretical result shows that \textsf{CompLRB} can be very useful for large markets with many participants, i.e., large $N$ and $K$.

\begin{remark} 
\label{rmk: unknown_phi}
In practice, the job side preferences, i.e., $\bPhi^*$, might be unknown to the matching platform as well. Then, we need to learn both matrices $\bTheta^*$ and $\bPhi^*$ online. Our algorithm can be easily adapted to this setting, and will result in a regret bound of the same scale as in Theorem \ref{theorem: stable_bandit}.
\end{remark}

\section{Experiments}\label{sec: experiments}

We now demonstrate the practical relevance and effectiveness of our proposed approaches in both offline and online settings using synthetic data and real data of labor market.

In the offline setting, we compare the following methods: (i) \textsf{Low Rank}: the nuclear norm regularization approach we propose, and (ii) \textsf{Naive}: an entry-wise sample-average estimator. 

In the online setting, we compare the following online optimal matching algorithms: (i) \textsf{CombLRB}: combinatorial low-rank bandit algorithm we propose, (ii) \textsf{CUCB}: combinatorial upper confidence bound algorithm developed by \citep{chen2013combinatorial},
and (iii) \textsf{CTS}: combinatorial thompson sampling proposed by \citep{wang2018thompson}. For online stable matching, we compare the following algorithms: (i) \textsf{CompLRB}: competing low-rank bandit algorithm we propose, and (ii) \textsf{CompB}: competing bandit algorithm proposed by \cite{liu2020competing}.

\subsection{Synthetic Data}\label{sec: synt}
\begin{figure}[h]
    \centering

        \centering
        \begin{subfigure}{0.48\textwidth}  
            \centering
        \includegraphics[width=\linewidth]{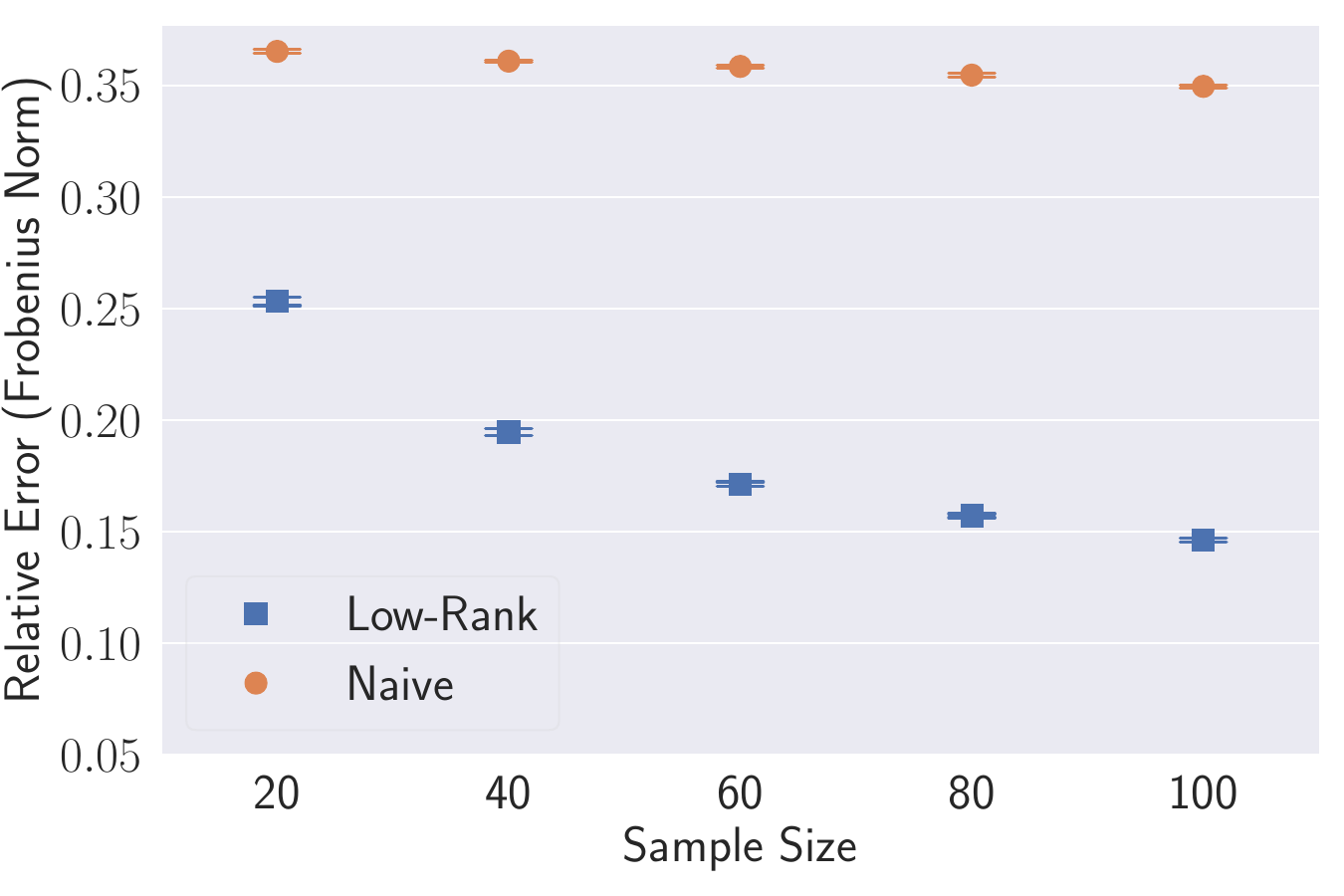}
        \caption{Frobenius Norm Error ($\|\cdot\|_F$)}
        \label{fig: offline_synthetic_Fnorm}
        \end{subfigure}%
        \hfill
        \begin{subfigure}{0.48\textwidth}  
            \centering
        \includegraphics[width=\linewidth]{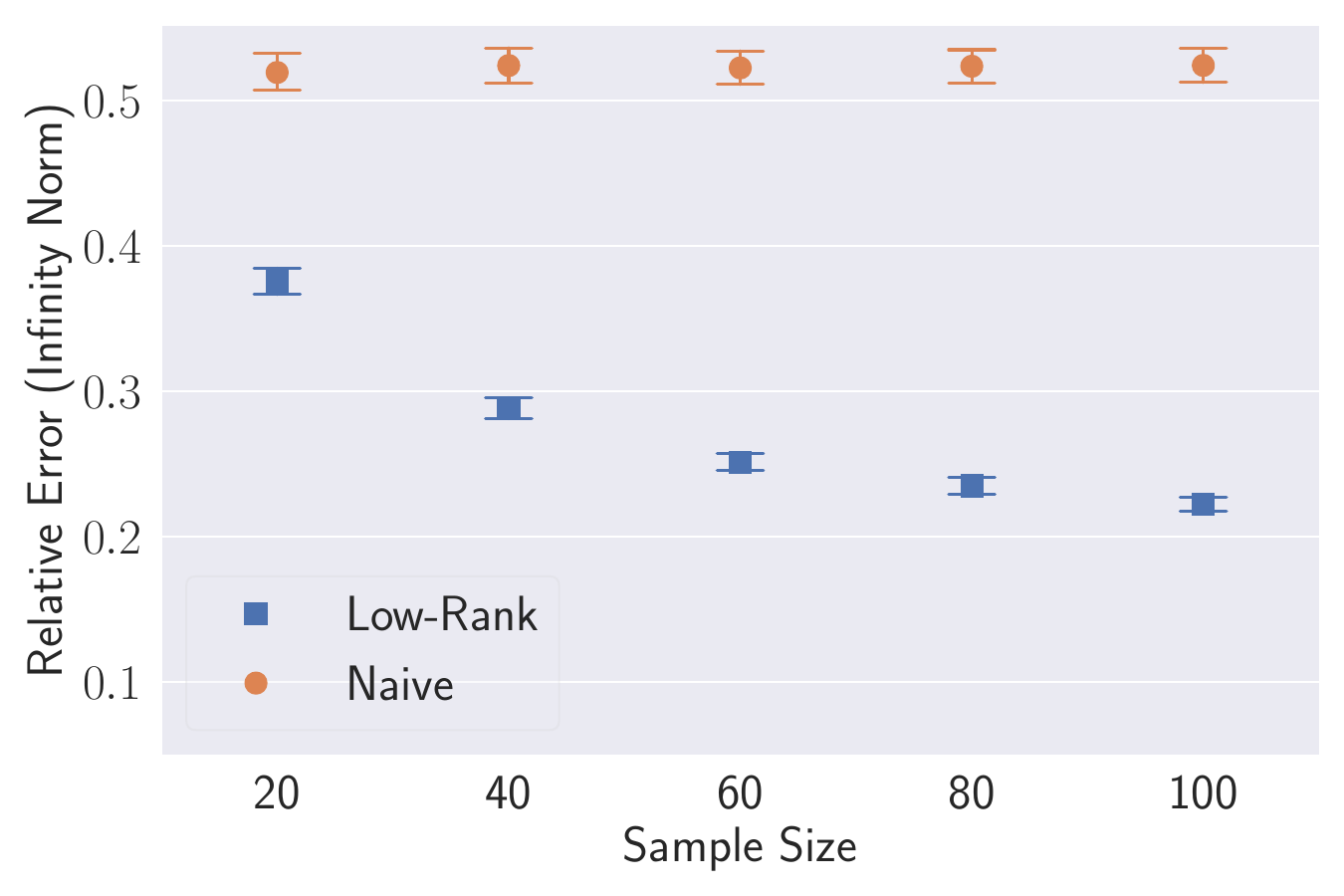}
        \caption{Entry-Wise Error ($\|\cdot\|_\infty$)}
        \label{fig: offline_synthetic_infnorm}
        \end{subfigure}
    \caption{Relative estimation errors of the matching reward matrix $\bTheta^*$ in Frobenius norm (left) and infinity norm (right) averaged over 50 trials. Error bars represent 95\% confidence intervals. We consider $N=100$ worker types, $K=100$ job types, and the matrix rank $r=3$. Sample size on the x-axis refers to the number of matchings $n$. ``Low-Rank" represents the nuclear norm estimator, and ``Naive" represents the entry-wise sample average estimator.}
    \label{fig: offline_synthetic}
\end{figure}

\textbf{Offline Matching Data.} We synthetically generate the matching reward matrix $\bTheta^*$, and the details are provided in Appendix \ref{app: synthetic_exp}. We first compare the the relative estimation errors of learning the matching reward matrix $\bTheta^*$ via \textsf{Low Rank} and \textsf{Naive} estimators in Frobenius norm and entry-wise norm in Figure~\ref{fig: offline_synthetic}.  The relative estimation error of a specified norm is equal to the estimation error of the matrix in that norm divided by the specified norm of the ground-truth matrix $\bTheta^*$. 

We find that our proposed estimator \textsf{Low-Rank} substantially outperforms the benchmarks in both Frobenius norm and infinity norm. The entry-wise sample-average estimator \textsf{Naive} 
takes the entry-wise empirical mean as estimates for observed entries, and the average of observations from the same row for unobserved entries. Thus, \textsf{Naive} does not fully utilize the low-rank structure of the reward matrix, and might introduce additional bias for unobserved entries. Instead, our approach captures the low-rank structure of $\bTheta^*$ efficiently, and thus delivers much smaller estimation errors through shared information across entries. Our results are consistent over varying matching sample size $n$. Matching our theory, the estimation error of \textsf{Low-Rank} decreases with increasing sample size; in contrast, \textsf{Naive} converges slowly due to insufficient samples entry-wise and potential bias for unobserved entries. 

\begin{figure}[htbp]
    \centering

        \centering
        \begin{subfigure}{0.48\textwidth}  
            \centering
            \includegraphics[width=\linewidth]{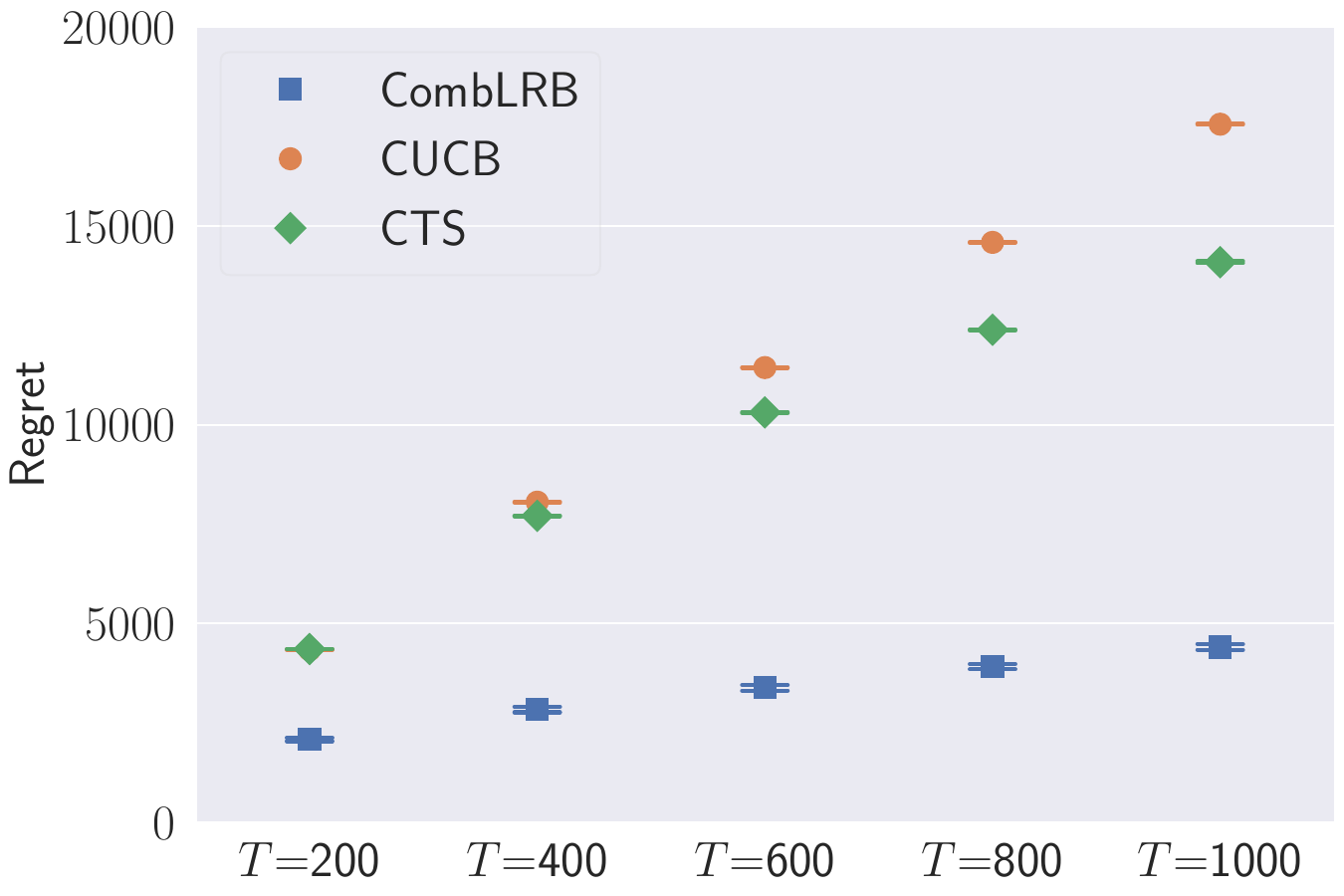}
            \caption{Optimal Matching}
            \label{fig: online_synthetic_data}
        \end{subfigure}%
        \hfill
        \begin{subfigure}{0.48\textwidth}  
            \centering
            \includegraphics[width=\linewidth]{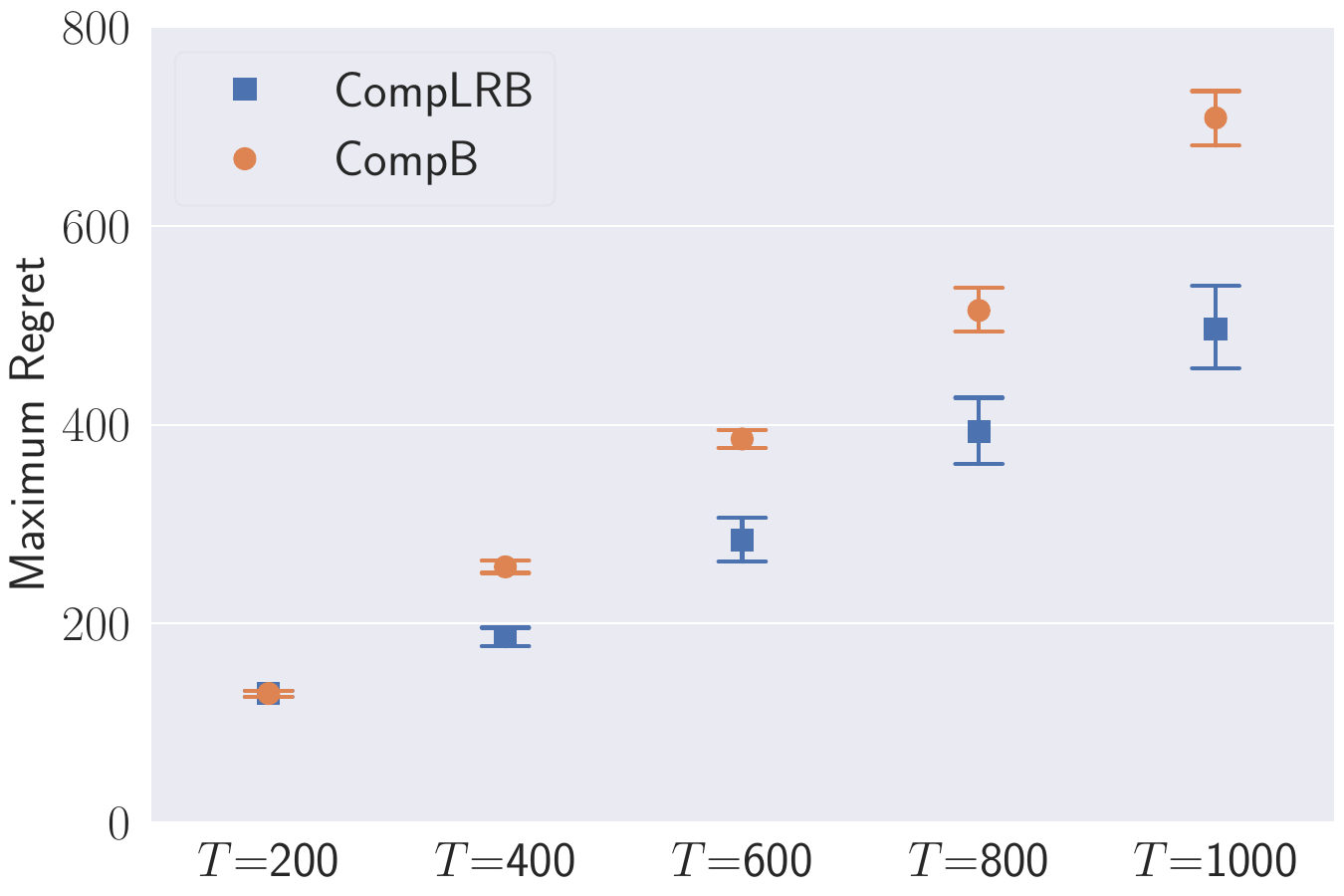}
            \caption{Stable Matching}
            \label{fig: online_stable_synthetic_data}
        \end{subfigure}
    \caption{Regret for optimal matching (left) and maximum per-worker regret for stable matching (right) averaged over 50 trials. Error bars represent 95\% confidence intervals. We consider $N=100$ worker types, $K=100$ job types, and the matrix rank $r=3$. `\textsf{CombLRB}' and `\textsf{CompLRB}' represent our combinatorial low-rank bandit algorithm and competing low-rank bandit algorithm respectively. }
    \label{fig: online_synthetic}
\end{figure}

\textbf{Online Matching Algorithms.} 
Figure \ref{fig: online_synthetic} compares the cumulative regrets over varying time horizon $T$ for optimal matching (Figure \ref{fig: online_synthetic_data}) and stable matching (Figure \ref{fig: online_stable_synthetic_data}) respectively. 
Appendix \ref{app: synthetic_exp} provides more details.

Similar to the offline setting, we find in Figure \ref{fig: online_synthetic_data} that our low-rank approach \textsf{CombLRB} significantly outperforms other benchmarks for optimal matching. \textsf{CUCB} and $\textsf{CTS}$ are based on the ideas of upper confidence bound and Thompson sampling; they do not exploit the low-rank structure but instead learn the true reward of each arm individually. Thus, these algorithms cannot efficiently learn and identify the optimal matching given relatively short time horizons. 

Figure \ref{fig: online_stable_synthetic_data} similarly shows that our algorithm \textsf{CompLRB} achieves much smaller cumulative regret for stable matching over other benchmarks. Note that here we consider the worst-case scenario with ``maximum regret'' --- i.e., we compare the maximum of the $N$ per-worker cumulative regrets for different algorithms. Since our algorithm has better performance in maximum regret, our algorithm also obtains a significant improvement over other benchmarks in total cumulative regrets of all workers in the market. As expected, our \textsf{CompLRB} obtains such an improvement through leveraging the underlying low-rank structure of the worker reward matrix, compared to \textsf{CompB}, which uses sample average to estimate the worker rewards and hence worker preferences.

In summary, the empirical results obtained from synthetic experiments align with our theory, given the low-rank nature of the true reward matrix. 

Next, we further explore the robustness of our algorithms on a real data, where the low-rank assumption might not hold.

\subsection{Real Data of Labor Market}
\label{sec: real_exp}

We further evaluate the real-world performance of all our approaches on one of the largest workforce dataset provided by Revelio Lab\footnote{See \url{https://wrds-www.wharton.upenn.edu/pages/about/data-vendors/revelio-labs/}.}, which collects matching information for a diverse set of job and candidate profiles. We use a subset of the individual-level data of employment duration of mid-level software engineers in the United States from 2010 to 2015. 
In this experiment, we cluster the engineers (i.e., the workers) and companies (i.e., the jobs) into 50 groups respectively, i.e., $N=K=50$. Then, we create the true reward matrix $\bTheta^*$, where the value of each entry takes the average of an indicator of whether an employment exceeds six months over all workers who belong to the corresponding worker and job group. That is, the $(i, j)^\tth$ entry $\bTheta^{*(i, j)}$ represents the probability of a worker employment from group $i$ lasting long than six months in a company from group $j$.
Intuitively, we measure the matching reward by the worker satisfaction; the higher the probability is, the more satisfied the worker is with the corresponding company.
It is worth noting that, in this real-world setting, our reward matrix might not satisfy the low-rank assumption, which is different from our synthetic experiments with low-rankness imposed.

\begin{figure}[htbp]
    \centering
        \centering
        \begin{subfigure}{0.48\textwidth}  
            \centering
            \includegraphics[width=\linewidth]{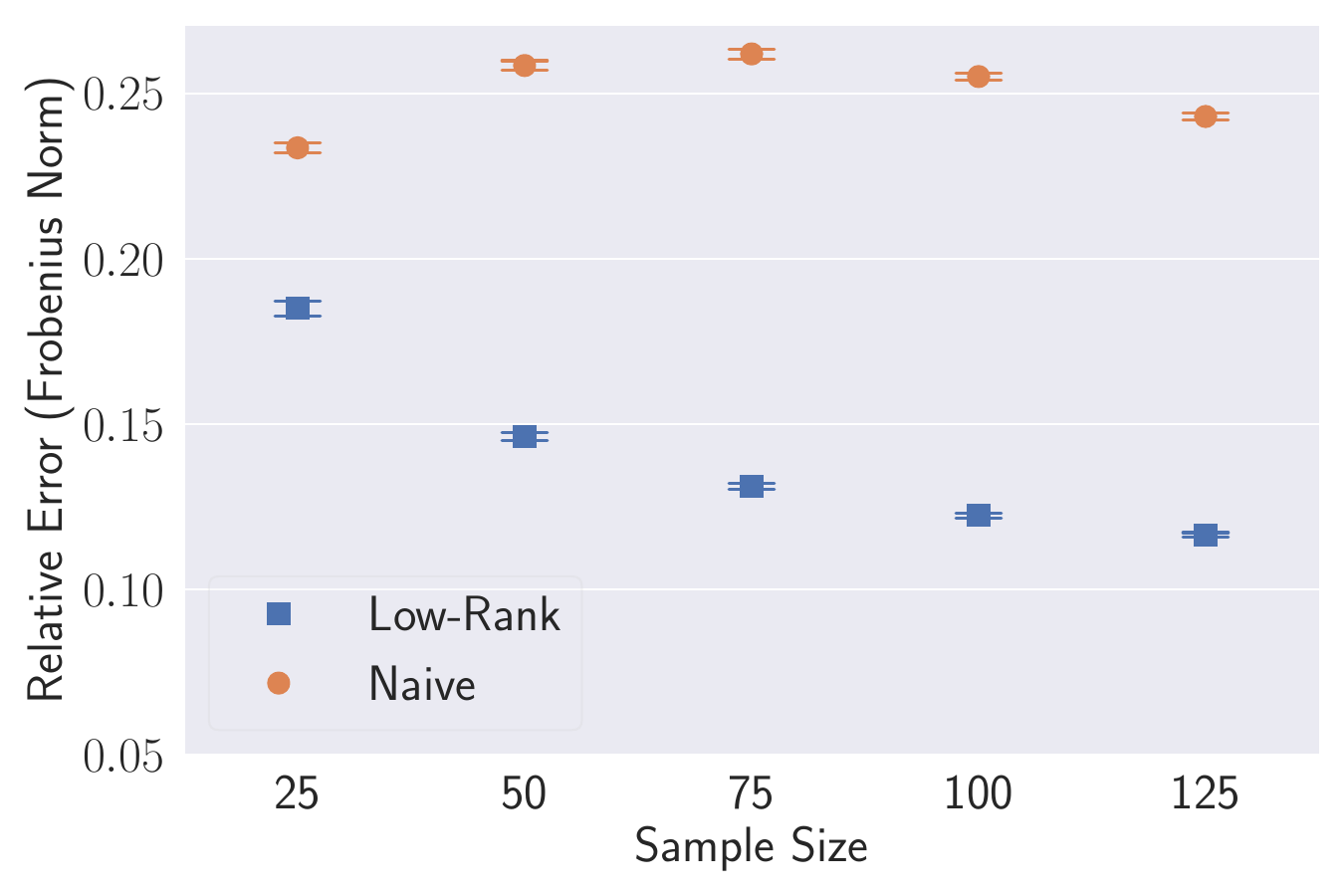}
            \caption{Frobenius Norm Error ($\|\cdot\|_F$)}
            \label{fig: offline_real_Fnorm}
        \end{subfigure}%
        \hfill
        \begin{subfigure}{0.48\textwidth}  
            \centering
            \includegraphics[width=\linewidth]{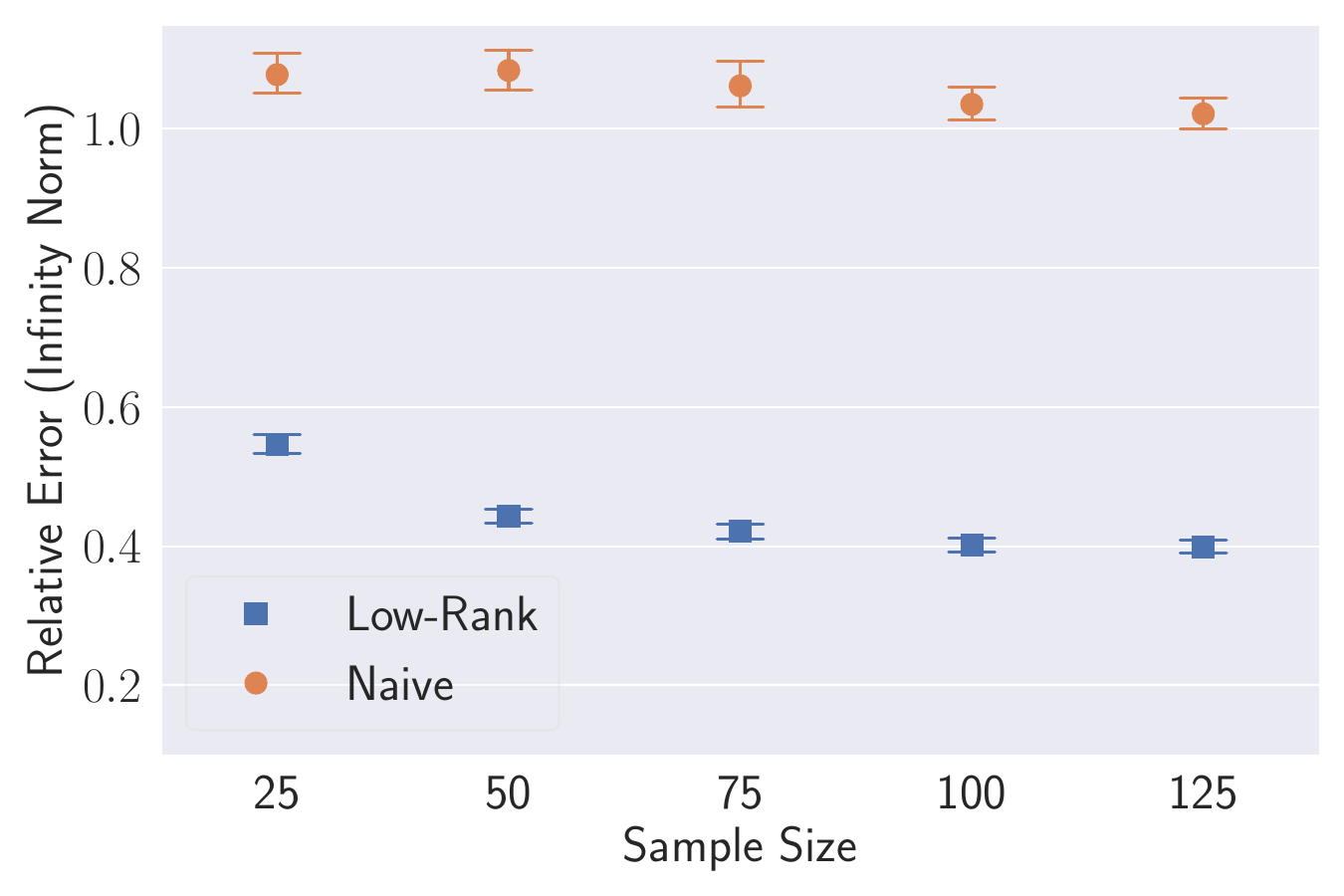}
            \caption{Entry-Wise Error ($\|\cdot\|_\infty$)}
            \label{fig: offline_real_infnorm}
        \end{subfigure}
    \caption{Relative estimation errors of the matching reward matrix $\bTheta^*$ in Frobenius norm (left) and infinity norm (right) averaged over 50 trials. Error bars represent 95\% confidence intervals. We consider $N=50$ worker types, and $K=50$ job types. Sample size on the x-axis refers to the number of matchings $n$. `\textsf{Low-Rank}' represents our nuclear norm regularization approach.}
    \label{fig: offline_real}
\end{figure}

\textbf{Offline Matching Data.}
Figure~\ref{fig: offline_real} presents the results of an offline reward learning experiment on our real data of labor market. The results again exhibit the superb performance of our low-rank matrix completion approach, similar to our synthetic experiments. Notably, our approach improves upon the best benchmark in the Frobenius norm and entry-wise norm by 44\% and 43\% respectively on average across all different sample sizes. Basically, our approach can learn the worker satisfaction with only few samples, and thus gain early insights into a company's employment condition. 

\textbf{Online Matching Algorithms.} We also aim to learn the matching decisions directly in an online manner through our matrix completion approach. Note that the total number of rewards (i.e., $N\times K=2500$) to learn is much larger than the time horizon $T$ considered in our experiments; that is, we have relatively short time horizons and large matching markets. 

\begin{figure}[htbp]
    \centering
        \centering
        \begin{subfigure}{0.48\textwidth}  
            \centering
            \includegraphics[width=\linewidth]{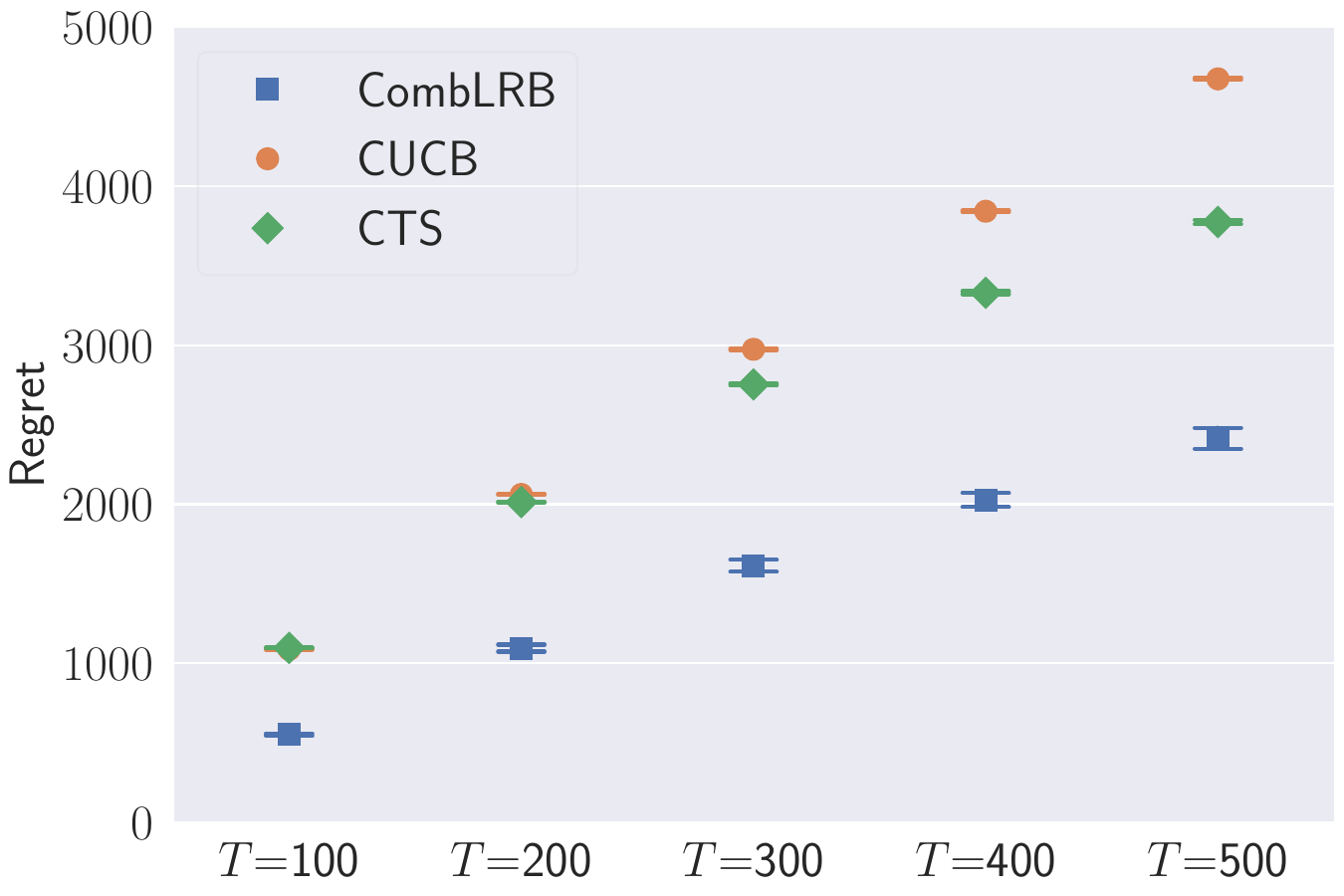}
            \caption{Optimal Matching}
            \label{fig: online_realdata}
        \end{subfigure}%
        \hfill
        \begin{subfigure}{0.48\textwidth}  
            \centering
            \includegraphics[width=\linewidth]{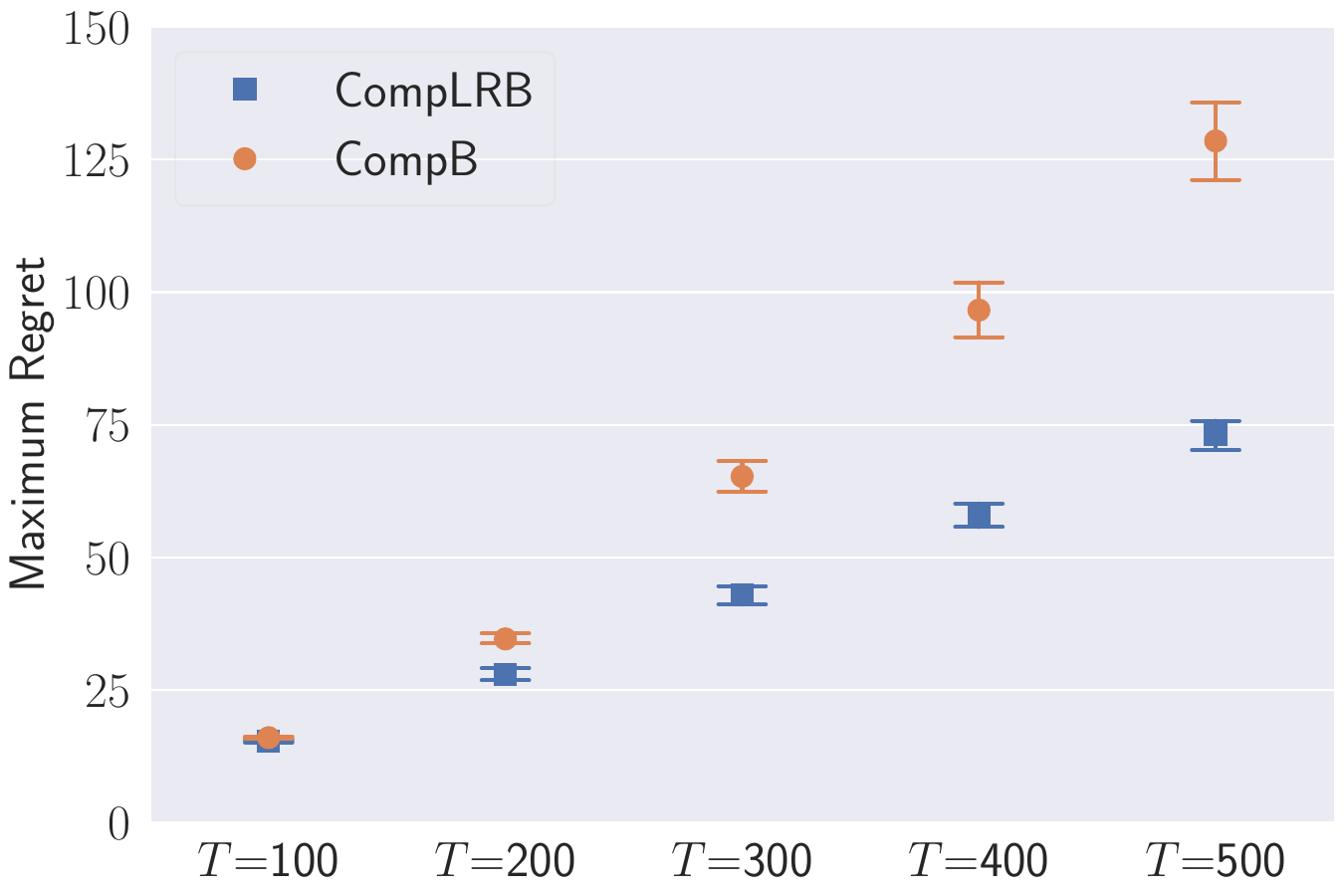}
            \caption{Stable Matching}
            \label{fig: online_stable_realdata}
        \end{subfigure}
    \caption{Regret for optimal matching (left) and maximum per-worker regret for stable matching (right) averaged over 50 trials. Error bars represent 95\% confidence intervals. We consider $N=50$ worker types, and $K=50$ job types. `\textsf{CombLRB}' and `\textsf{CompLRB}' represent our combinatorial low-rank bandit algorithm and competing low-rank bandit algorithm respectively. }
    \label{fig: online_real}
\end{figure}

The results are presented in Figure~\ref{fig: online_real}. Figure~\ref{fig: online_realdata} shows that, even when the reward matrix might not be low-rank, our \textsf{CombLRB} algorithm outperforms other two benchmark algorithms \textsf{CUCB} and \textsf{CTS} over varying time horizons for online optimal matching. Specifically, \textsf{CombLRB} improves on the regret by 41\% compared to the best performing \textsf{CTS} among the benchmarks.  
Figure~\ref{fig: online_stable_realdata} further confirms the efficiency of using low-rank matrix completion in the online stable matching setting. Our algorithm \textsf{CompLRB} beats the benchmark algorithm by 36\% on average across all different time horizons.

\section{Conclusion}
In this paper, we focus on efficiently learning matching qualities from a small amount of offline matching data for large-scale centralized matching platforms. Motivated by the natural low-rank matrix structure of two-sided markets, we utilize a matrix completion approach via nuclear norm regularization to estimate the matching rewards efficiently. To the best of our knowledge, we propose the first matrix completion framework to address a reward learning problem in the matching setting. Our matching problem involves a challenging dependent sampling scheme due to matching interference;
we develop a new proof strategy based on a linearization technique and establish a near-optimal error bound in Frobenius norm. Furthermore, we propose a novel double-enhancement procedure that carefully quantify entry-wise estimation atop the nuclear norm regularized estimates and ensures an entry-wise guarantee. In the online setting, we propose two algorithms \textsf{CombLRB} and \textsf{CompLRB} to efficiently learn optimal matching and stable matching policies respectively, thereby improving regret bounds in the matrix dimensions. Finally, our empirical experiments show that our matrix completion approach can indeed boost both offline reward learning and online decision making in the matching problems. Both our theoretical and empirical findings underscore the importance of adopting matrix completion methods in matching markets.

\bibliographystyle{ormsv080} 
{\SingleSpacedXI
\bibliography{ref}
}

\newpage
\renewcommand{\theHsection}{\Alph{section}}
\begin{APPENDICES}

\section{Proof of Theorem~\ref{theorem:errorbound}}
\subsection{Major Steps of the Proof}
\label{subsec: major_steps}

We summarize the three steps of our proof in Lemma \ref{lem:step1}, Lemma \ref{lem:step2} and Proposition~\ref{lem:step3} (stated in Section \ref{sec:theorems}) respectively. 
We define $\vD = \bTheta^*-\nuclearest$, and let $\mathfrak{C}=\sum_{t=1}^{\sizeone}\sum_{i=1}^N\varepsilon_{t}^{(i)}\bX_t^i$. 
The operator norm of the noise matrix $\mathfrak{C}$ indicates the ``scale" of the noises; the larger the operator norm, the higher the noise level. Lemma \ref{lem:step1} upper bounds the error ``collected" by $\bX_t$ for deterministic $\mathfrak{C}$. As long as the regularized hyperparameter $\lambda$ takes a sufficiently large value, the error collected by $\bX_t$ can be upper bounded with respect to the corresponding error term's Frobenius norm.
\begin{lemma}
\label{lem:step1}
For any $\lambda\geq3\|\mfC\|_{\mathrm{op}}/\sizeone, $ 
we have
\begin{equation*}    
\frac{\sum_{t=1}^{\sizeone}\sum_{i=1}^N\langle \bX_{t}^i,\vD\rangle^2}{\sizeone}\leq 4\lambda \sqrt{r}\|\vD\|_F.
\end{equation*}
\end{lemma}

Next, Lemma~\ref{lem:step2} provides a probabilistic bound on the operator norm of the noise matrix $\mathfrak{C}$. In other words, the condition $\lambda\geq 3\|\mfC\|_{\mathrm{op}}/\sizeone$ in Lemma~\ref{lem:step1} holds with high probability,
given our choice of $\lambda$ in Theorem \ref{theorem:errorbound}. Similar to Proposition~\ref{lem:step3}, the key challenge of Lemma~\ref{lem:step2} lies in the dependent structure of the observational data. 
\begin{lemma}\label{lem:step2}
   Suppose $\sqrt{\sizeone}\geq \max\{\alpha,\log(NK\sizeone^2)\}$. Then, there exists a constant $c_1$ such that
    \begin{equation}
        \|\mfC\|_{\op}\leq c_1\sigma\big(\alpha+\log(N+K)\big)\sqrt{\sizeone}\nonumber
    \end{equation}
    with probability greater than $1-3\exp(-\alpha)$ for any $\alpha>0$.
\end{lemma}

As summarized before, our last step will be the RSC condition previously mentioned in Proposition~\ref{lem:step3}. With these three steps, we are now ready to prove 
Theorem~\ref{theorem:errorbound}.

\subsection{Proof of Theorem~\ref{theorem:errorbound}}\label{sec: proof_of_f_norm}
\proof{Proof of Theorem~\ref{theorem:errorbound}.} By the constraint in \eqref{program:nonconvex}, the rank of ${\nuclearest}$ is less than or equal to $r.$ Since $\rank(\bTheta^*)\leq r$, it is straightforward to see that $\rank(\vD)\leq 2r.$ 
    
First note that if $\|\vD\|_F^2\leq {c_0\pmin NK\alpha/{\sizeone}},$ then our argument goes. Otherwise, since $\|\vD\|_{L^2(\Pi)}^2 \geq  \|\vD\|_F^2 /(\pmin K) > c
_0 N\alpha/\sizeone,$ by the definition of the set $\mathcal{C}_{\alpha}(2r)$ in Proposition~\ref{lem:step3}, we have $\frac{1}{2}\vD\in\mathcal{C}_{\alpha}(2r
).$ Then we can apply Proposition~\ref{lem:step3} and have with probability greater than $1-\exp(-\alpha)$ that,
\begin{align}\label{ineq: lhs}
\frac{1}{\sizeone}\sum_{t=1}^{\sizeone}\sum_{i=1}^N\big\langle \bX_t^i, \frac{1}{2}\vD\big\rangle^2 > c_2\left\|\frac{1}{2}\vD\right\|_{L^2(\Pi)}^2 - c_3\left(\frac{\pmin r^2K\log[(N+K)\sizeone]}{\sizeone}\right).
\end{align}
By Lemma~\ref{lem:step2} and our choice of $\lambda=c_{\lambda} \sigma\frac{\alpha+\log(N+K)}{\sqrt{\sizeone}}$ where $c_{\lambda}$ is large enough, we have $\lambda\geq3\|\mfC\|_{\mathrm{op}}/\sizeone$ holds with probability greater than $1-3\exp(-\alpha)$. Then by Lemma~\ref{lem:step1}, we have
\begin{equation}  \label{ineq:rhs}  \frac{1}{\sizeone}{\sum_{t=1}^{\sizeone}\sum_{i=1}^N\langle \bX_{t}^i,\vD\rangle^2}\leq 4\lambda \sqrt{r}\|\vD\|_F= \frac{4c_\lambda \sigma\sqrt{r}\big(\alpha+\log(N+K)\big)}{\sqrt{\sizeone}}\|\vD\|_F
\end{equation}
 with probability higher than $1-3\exp(-\alpha).$
    By combining \eqref{ineq: lhs} and \eqref{ineq:rhs}, we obtain with probability greater than $1-4\exp(-\alpha)$ that,
    \begin{align}
        c_2\|\vD\|_{L^2(\Pi)}^2-4c_3\left(\frac{\pmin r^2K\log[(N+K)\sizeone]}{\sizeone}\right)\leq \frac{4c_\lambda \sigma\sqrt{r}\left(\alpha+\log(N+K)\right)}{\sqrt{\sizeone}}\|\vD\|_F.\label{ineq:combine}
    \end{align}
On the RHS, the basic inequality yields that
    \begin{align*}
        \frac{4c_\lambda \sigma \sqrt{r}\big(\alpha+\log(N+K)\big)}{\sqrt{\sizeone}}\|\vD\|_F
        &\leq \frac{c_2}{2\pmin K}\|\vD\|_F^2+\frac{8\pmin Kc_\lambda^2\sigma^2 r\big(\alpha+\log(N+K)\big)^2}{c_2\sizeone},
    \end{align*}
    while on the LHS, we have 
\begin{equation*}
\|\bold{\Delta}\|_{L^2(\Pi)}^2\geq \frac{\|\bold{\Delta}\|_F^2}{\pmin K}.
\end{equation*}     
    Plugging the above into \eqref{ineq:combine}, we have
    \begin{align*}
        c_2\frac{\|\vD\|_{F}^2}{\pmin K}-4c_3\left(\frac{\pmin r^2K\log[(N+K)\sizeone]}{\sizeone}\right)
        \leq \frac{c_2}{2\pmin K}\|\vD\|_F^2+\frac{8\pmin Kc_\lambda^2\sigma^2 r\big(\alpha+\log(N+K)\big)^2}{c_2\sizeone}.
    \end{align*}
    Rearranging the above inequality gives us
    \begin{align*}
        \frac{c_2}{2\pmin K}\|\vD\|_{F}^2&\leq 4c_3\left(\frac{\pmin r^2K\log[(N+K)\sizeone]}{\sizeone}\right)+\frac{8\pmin Kc_\lambda^2\sigma^2 r\big(\alpha+\log(N+K)\big)^2}{c_2\sizeone},
    \end{align*}
   which further yields
    \begin{align*}
        \|\vD\|_F^2\leq \frac{8c_3\pmin^2 r^2K^2\log[(N+K)\sizeone]}{c_2\sizeone}+\frac{16\pmin^2 K^2c_\lambda^2\sigma^2 r\big(\alpha+\log(N+K)\big)^2}{c_2^2\sizeone}.
    \end{align*}
   So \eqref{ineq:combine} implies that
    \begin{equation}
       \frac{\|\vD\|_F}{\sqrt{NK}}\leq c_4\pmin\max\left\{c_\lambda\sigma\left(\alpha+\log(N+K)\right)\sqrt{\frac{rK}{N\sizeone}},r \sqrt{\frac{ K\log[(N+K)\sizeone]}{N\sizeone}}\right\},
       \label{ineq: thm1explicit}
    \end{equation}
    where $c_4$ is a universal constant that depends on $c_2$ and $c_3$. Since \eqref{ineq:combine} holds with probability greater than $1-4\exp(-\alpha),$ our argument goes.\Halmos

\subsection{Proof of Lemma~\ref{lem:step1}}
\proof{Proof of Lemma~\ref{lem:step1}.}
By \eqref{program:nonconvex} we have
    \begin{equation*}
        \frac{1}{\sizeone}\sum_{t=1}^{\sizeone}\|Y_t-\mathcal{X}_t\big({{\nuclearest}}\big)\|^2+\lambda\|{{\nuclearest}}\|_*\leq \frac{1}{\sizeone}\sum_{t=1}^{\sizeone}\|Y_t-\mathcal{X}_t({\bTheta^*})\|^2+\lambda\|\bTheta^*\|_*,
    \end{equation*}
    which implies
    \begin{equation}\label{ineq_proof_of_step1_1}
        \frac{1}{\sizeone}\sum_{t=1}^{\sizeone}\|Y_t-\mathcal{X}_t({{\nuclearest}})\|^2-\frac{1}{\sizeone}\sum_{t=1}^{\sizeone}\|Y_t-\mathcal{X}_t({\bTheta^*})\|^2\leq \lambda\|\bTheta^*\|_*-\lambda\|{{\nuclearest}}\|_*.
    \end{equation}
   By the definition of $\mathcal{X}_t$, we have 
   \begin{align*}
       Y_t -\mathcal{X}_t({\nuclearest}) &= Y_t -\mathcal{X}_t(\bTheta^*)+ \mathcal{X}_t(\bTheta^*)-\mathcal{X}_t({\nuclearest})\\
       & = Y_t -\mathcal{X}_t(\bTheta^*)+ \mathcal{X}_t(\vD).
   \end{align*}
   Then
   \begin{align*}
        &\frac{1}{\sizeone}\sum_{t=1}^{\sizeone}\|Y_t-\mathcal{X}_t({{\nuclearest}})\|^2-\frac{1}{\sizeone}\sum_{t=1}^{\sizeone}\|Y_t-\mathcal{X}_t({\bTheta^*})\|^2\\
        & =  \frac{1}{\sizeone}\sum_{t=1}^{\sizeone}\|Y_t-\mathcal{X}_t({\bTheta^*})+\mathcal{X}_t(\vD)\|^2-\frac{1}{\sizeone}\sum_{t=1}^{\sizeone}\|Y_t-\mathcal{X}_t({\bTheta^*})\|^2\\
        & = \frac{2}{\sizeone}\sum_{t=1}^{\sizeone}\langle Y_t-\mathcal{X}_t({\bTheta^*}),\mathcal{X}_t(\vD)  \rangle+\frac{1}{\sizeone} \sum_{t=1}^{\sizeone} \|\mathcal{X}_t(\vD)\|^2.
   \end{align*}
   By \eqref{model: data-generating}, we have 
   \[Y_t-\mathcal{X}_t({\bTheta^*})=\begin{bmatrix}
    \varepsilon_t^{(1)} & \varepsilon_t^{(2)} & \cdots & \varepsilon_t^{(N)}
\end{bmatrix}.\]
   
   Then by the definition of the noise matrix $\mathfrak{C},$
   \begin{align*}
       \sum_{t=1}^{\sizeone}\langle Y_t-\mathcal{X}_t({\bTheta^*}),\mathcal{X}_t(\vD)  \rangle&=\sum_{t=1}^{\sizeone}\sum_{i=1}^N \varepsilon_t^{(i)}\langle\bX_t^i, \vD\rangle\\
       &=\sum_{t=1}^{\sizeone}\sum_{i=1}^N \langle \varepsilon_t^{(i)}\bX_t^i, \vD\rangle\\
       & = \left\langle \sum_{t=1}^{\sizeone}\sum_{i=1}^N \varepsilon_t^{(i)}\bX_t^i, \vD\right\rangle = \langle \mathfrak{C},\vD\rangle.
   \end{align*}
   Plugging the above into \eqref{ineq_proof_of_step1_1} yields that
   \begin{align*}
      \frac{2}{\sizeone}\sum_{t=1}^{\sizeone}\langle Y_t-\mathcal{X}_t({\bTheta^*}),\mathcal{X}_t(\vD)  \rangle+\frac{1}{\sizeone} \sum_{t=1}^{\sizeone} \|\mathcal{X}_t(\vD)\|^2& = \frac{2}{\sizeone}\langle \mathfrak{C},\vD\rangle + \frac{1}{\sizeone}\sum_{t=1}^{\sizeone}\|\mathcal{X}_t(\vD)\|^2\\
      & = \frac{2}{\sizeone}\langle \mathfrak{C},\vD\rangle + \frac{1}{\sizeone}\sum_{t=1}^{\sizeone}\sum_{i=1}^N \langle\bX_t^i, \vD\rangle^2\leq \lambda\|\bTheta^*\|_*-\lambda\|{\nuclearest}\|_*
   \end{align*}
   where the second equality is given by the definition of $\mathcal{X}_t.$
Then by $\lambda\geq 3\|\mfC\|_{\text{op}}/\sizeone$, the above inequality gives
    \begin{align}
        \frac{1}{\sizeone}\sum_{t=1}^{\sizeone}\sum_{i=1}^N \langle \bX_t^i,\bold{\bold{\Delta}}\rangle^2&\leq -\frac{2}{\sizeone}\langle \mfC, \vD\rangle+\lambda\|\bTheta^*\|_*-\lambda \|{\nuclearest}\|_*\nonumber\\
        &\leq \frac{2}{\sizeone}\|\bold{\Delta}\|_*\|\mfC\|_\op+\lambda\|\bold{\Delta}\|_*\nonumber\\
        &\leq \frac{5}{3}\lambda\|\bold{\Delta}\|_*.\label{ineq: proof_of_step1_2}
    \end{align}
In Section~\ref{sec: proof_of_f_norm} we have shown that $\rank(\vD)\leq 2r.$ Then we can derive the following inequality: $\|\bold{\Delta}\|_*\leq \sqrt{2r} \|\bold{\Delta}\|_F$. By substituting this inequality into the previous inequality~\eqref{ineq: proof_of_step1_2}, we obtain:
\begin{equation*}
\frac{1}{\sizeone}\sum_{t=1}^{\sizeone}\sum_{i=1}^N \langle \bX_t^i,\bold{\Delta}\rangle^2\leq 4\lambda \sqrt{r}\|\bold{\Delta}\|_F.
\end{equation*}
Since $\vD = \bTheta^*-{\nuclearest},$ our argument goes.
\Halmos

\subsection{Proof of Lemma~\ref{lem:step2}}
\proof{Proof of Lemma~\ref{lem:step2}.}
For every $t\in[\sizeone],$ define
$$\bold{B}_t=\sum_{i=1}^N \varepsilon_t^{(i)}\bX_t^i.$$ Thus we have $\mfC=\sum_{t=1}^{\sizeone}\bold{B}_t$, $\mathbb{E}[\bold{B}_t]=\boldsymbol{0}$, and each $\bold{B}_t$ is independent. Let $\vartheta= \sigma\sqrt{2\log\big(NK\sizeone\big)+2\alpha}$. For each $t\in[\sizeone]$ and $i\in[N]$ define the truncation of $\varepsilon_t^{(i)}$ as $\bar\varepsilon_t^{(i)}=\varepsilon_t^{(i)}\mathbbm{1}\{|\varepsilon_t^{(i)}|\leq \vartheta\}.$ Also define \[\bar{\bold{B}}_t=\sum_{i=1}^N\bar\varepsilon_t^{(i)}\bX_t^i.\] 
Notice that for any $\gamma\geq 0,$
\begin{align}
    \mathbb{P}[\|\mfC\|_{\op}\geq \gamma+\vartheta]&=\mathbb{P}\bigg[\Big\|\sum_{t=1}^{\sizeone}\bold{B}_t\bigg\|_{\op}\geq \gamma+\vartheta\Big]\nonumber\\
    &\leq \mathbb{P}\bigg[\Big\|\sum_{t=1}^{\sizeone}\bar{\bold{B}}_t\Big\|_{\op}\geq \gamma\bigg]+\mathbb{P}\bigg[\bigcup_{t=1}^{\sizeone}\bigcup_{i=1}^N\{|\varepsilon_t^{(i)}|> \vartheta\}\bigg]\label{ineq: prob_cop}.
\end{align}
Since $\varepsilon_t^{(i)}$ are $\sigma$-subgaussian noises (defined in Definition \ref{def:subgaussian}), we have
\begin{align}
    \mathbb{P}\bigg[\bigcup_{t=1}^{\sizeone}\bigcup_{i=1}^N\{|\varepsilon_t^{(i)}|> \vartheta\}\bigg]&\leq \sum_{t=1}^{\sizeone}\sum_{i=1}^N\mathbb{P}\big[|\varepsilon_t^{(i)}|> \vartheta\big]
    \leq N\sizeone \cdot 2\exp\left(\frac{-\vartheta^2}{2\sigma^2}\right)=2\exp(-\alpha)\cdot K^{-1}\leq 2\exp(-\alpha).\label{ineq: noise}
\end{align}
Next, we provide an upper bound for $\mathbb{P}\bigg[\Big\|\sum_{t=1}^{\sizeone}\bar{\bold{B}}_t\Big\|_{\op}\geq \gamma\bigg].$ For each $t\in[\sizeone],$ define a centered random matrix $$\bold{G}_t= \bar{\bold{B}}_t- \sum_{i=1}^N\mathbb{E}[\bar\varepsilon_t^{(i)}]\bX_t^i.$$ Then
\begin{align}
\Big\|\sum_{t=1}^{\sizeone}\bar{\bold{B}}_t\bigg\|_\op&= \Big\|\sum_{t=1}^{\sizeone}\Big(\bold{G}_t+\sum_{i=1}^N\mathbb{E}[\bar\varepsilon_t^{(i)}]\bX_t^i\Big)\bigg\|_\op
\nonumber\\&\leq \Big\|\sum_{t=1}^{\sizeone}{\bold{G}}_t\Big\|_\op+\Big\|\sum_{t=1}^{\sizeone}\sum_{i=1}^N\mathbb{E}[\bar\varepsilon_t^{(i)}]\bX_t^i\Big\|_{\op}\nonumber\\
&\leq  \Big\|\sum_{t=1}^{\sizeone}{\bold{G}}_t\Big\|_\op+\Big\|\sum_{t=1}^{\sizeone}\sum_{i=1}^N\mathbb{E}[\bar\varepsilon_t^{(i)}]\bX_t^i\Big\|_{F}\nonumber\\
&\leq\underbrace{\Big\|\sum_{t=1}^{\sizeone}{\bold{G}}_t\Big\|_\op}_{:=h_0}+\underbrace{\sqrt{NK}\Big\|\sum_{t=1}^{\sizeone}\sum_{i=1}^N\mathbb{E}[\bar\varepsilon_t^{(i)}]\bX_t^i\Big\|_{\infty}}_{:=h_1} .\label{ineq: barB_t}
\end{align}
For the term $h_1$ on the RHS, we have
\begin{align*}
    h_1=\sqrt{NK}\Big\|\sum_{t=1}^{\sizeone}\sum_{i=1}^N\mathbb{E}[\bar\varepsilon_t^{(i)}]\bX_t^i\Big\|_{\infty}&\leq \sqrt{NK}\sum_{t=1}^{\sizeone}\Big\|\sum_{i=1}^N\mathbb{E}[\bar\varepsilon_t^{(i)}]\bX_t^i\Big\|_{\infty}\\
    &\leq \sqrt{NK}\sizeone\max_{i\in[N], t\in[\sizeone]}|\mathbb{E}[\bar\varepsilon_t^{(i)}]|.
\end{align*}

Since $\mathbb{E}[\varepsilon_t^{(i)}]=0,$ we have for any $t\in[\sizeone], i\in[N]$
\begin{align*}
    \big|\mathbb{E}[\bar{\varepsilon}_{t}^i]\big|=\bigg|\mathbb{E}\Big[{\varepsilon}_{t}^i\mathbbm{1}\{|\varepsilon_t^{(i)}|\leq \vartheta\}\Big]\bigg|&=\bigg|\mathbb{E}\Big[\varepsilon_t^{(i)}\mathbbm{1}\{|\varepsilon_t^{(i)}|> \vartheta\}\Big]\bigg|\\
    &\leq \sqrt{\mathbb{E}\big[(\varepsilon_t^{(i)})^2\big]\mathbb{P}[|\varepsilon_t^{(i)}|> \vartheta]}\\
    &\leq \sqrt{2\sigma^2\exp\big[-\vartheta^2/(2\sigma^2)\big]}\\
    &=\frac{\sqrt{2}\sigma \exp(-\alpha/2)}{\sqrt{NK\sizeone}}\leq \frac{\sqrt{2}\sigma}{\sqrt{NK\sizeone}}
\end{align*}
where the last inequality is given by $\alpha>0.$
Thus 
\begin{align}
    h_1\leq \sqrt{NK}\sizeone\max_{i\in[N], t\in[\sizeone]}|\mathbb{E}[\bar\varepsilon_t^{(i)}]|\leq \sqrt{2}\sigma.
    \label{eq:h1}
\end{align}
Now we will bound the term $h_0$ via
matrix Berstein inequality. First, we provide an upper bound for $\|\bold{G}_t\|_\op$ and
$$\sigma_Z^2=\max\bigg\{\bigg\|\mathbb{E}\Big[\sum_{t=1}^{\sizeone} \bold{G}_t\bold{G}_t^\top\Big]\bigg\|_\op,\bigg\|\mathbb{E}\Big[\sum_{t=1}^{\sizeone} \bold{G}_t^\top\bold{G}_t\Big]\bigg\|_\op\bigg\}.$$ We have the following lemma which is proved at the end of this subsection:
\begin{lemma}\label{claim: G_t}
We have
\begin{equation*}
    \|\bold{G}_t\|_\op\leq 2\vartheta,\forall t\in[n],
    \quad\text{and}\quad
    \sigma_Z^2=\max\bigg\{\bigg\|\mathbb{E}\Big[\sum_{t=1}^{\sizeone} \bold{G}_t\bold{G}_t^\top\Big]\bigg\|_\op,\bigg\|\mathbb{E}\Big[\sum_{t=1}^{\sizeone} \bold{G}_t^\top\bold{G}_t\Big]\bigg\|_\op\bigg\}\leq \sizeone\sigma^2.
\end{equation*}
\end{lemma}

Given Lemma~\ref{claim: G_t}, we can bound $h_0$ via Lemma~\ref{lem: matrix_berstein_inequality}. By Lemma~\ref{lem: matrix_berstein_inequality}, for any $\gamma >0$,
\begin{align*}
    \mathbb{P}\bigg[h_0\geq\gamma\bigg]=\mathbb{P}\bigg[\Big\|\sum_{t=1}^{\sizeone} \bold{G}_t\Big\|_\op\geq\gamma\bigg]&\leq(N+K)\exp\Big\{\frac{-\gamma^2}{2\sigma_Z^2+(4\vartheta\gamma)/3}\Big\}\\
    &\leq (N+K)\exp\Big\{\frac{-\gamma^2}{2\sizeone\sigma^2+(4\vartheta\gamma)/3}\Big\}.
\end{align*}
By choosing
\[\gamma\geq \max\left\{2\sigma \sqrt{\sizeone}\sqrt{\alpha+\log(N+K)}, \frac{8\vartheta}{3}\big(\alpha+\log(N+K)\big)\right\},\]
we have
\begin{align*}
    (N+K)\exp\Big\{\frac{-\gamma^2}{2\sizeone\sigma^2+(4\vartheta\gamma)/3}\Big\}&= (N+K)\exp\Big\{\frac{-\frac{\gamma^2}{2}-\frac{\gamma^2}{2}}{2\sizeone\sigma^2+(4\vartheta\gamma)/3}\Big\}\\
    &\leq (N+K)\exp\Big\{\frac{-2\sizeone\sigma^2 (\alpha+\log(N+K))-\frac{4\vartheta\gamma}{3}\big(\alpha+\log(N+K)\big)}{2\sizeone\sigma^2+(4\vartheta\gamma)/3}\Big\}\\
    & = (N+K)\exp\left(-\alpha-\log(N+K)\right)\\
    & = \exp(-\alpha).
\end{align*}
Since 
\begin{align*}
    \frac{8\vartheta}{3}\big(\alpha+\log(N+K)\big)
    & = \frac{8\sqrt{2}\sigma}{3}\left(\alpha+\log(N+K)\right)\left(\alpha+\log(NK\sizeone)\right)\\
    & \leq {5\sigma}\left(\alpha+\log(N+K)\right)\left(\alpha+\log(NK\sizeone)\right),
\end{align*}
we can choose
\begin{align*}
    \gamma = 5\sigma\big(\alpha+\log(N+K)\big)\left(\sqrt{\sizeone}+\alpha+\log (NK\sizeone^2)\right),
\end{align*}
which yields that
\begin{equation*}
    \mathbb{P}\left[h_0\geq 5\sigma\big(\alpha+\log(N+K)\big)\left(\sqrt{\sizeone}+\alpha+\log (NK\sizeone^2)\right)\right]\leq \exp(-\alpha).
\end{equation*}
Combining with \eqref{ineq: barB_t} and $h_1\leq \sqrt{2}\sigma$ in \eqref{eq:h1} gives us
\begin{align*}
    &\mathbb{P}\left[\Big\|\sum_{t=1}^{\sizeone}\bar{\bold{B}}_t\bigg\|_\op\geq 5\sigma\big(\alpha+\log(N+K)\big)\left(\sqrt{\sizeone}+\alpha+\log (NK\sizeone^2)\right)+\sqrt{2}\sigma\right]\\
    &\leq \mathbb{P}\left[h_0\geq 5\sigma\big(\alpha+\log(N+K)\big)\left(\sqrt{\sizeone}+\alpha+\log (NK\sizeone^2)\right)\right]
    \leq \exp(-\alpha).
\end{align*}
By \eqref{ineq: prob_cop} and \eqref{ineq: noise}, we have that, with probability larger than $1-3\exp(-\alpha)$,
\begin{align*}
   \|\mfC\|_\op&\leq 5\sigma\big(\alpha+\log(N+K)\big)\left(\sqrt{\sizeone}+\alpha+\log (NK\sizeone^2)\right)+\sqrt{2}\sigma+\sigma\sqrt{2\log\big(NK\sizeone\big)+2\alpha}\\
   &\leq 8\sigma\big(\alpha+\log(N+K)\big)\left(\sqrt{\sizeone}+\alpha+\log (NK\sizeone^2)\right)\\
   &\leq 24\sigma \big(\alpha+\log(N+K)\big)\sqrt{n}
\end{align*}
where the last inequality is given by $\sqrt{n}\geq\max\{\alpha,\log(NK\sizeone^2)\},$
which completes the proof.
\Halmos

\proof{Proof of Lemma~\ref{claim: G_t}}
Given any $t\in[n]$, we can write $\bold{G}_t$ as
\begin{align*}
    \bold{G}_t &= \bar{\bold{B}}_t- \sum_{i=1}^N\mathbb{E}[\bar\varepsilon_t^{(i)}]\bX_t^i
    = \sum_{i=1}^N \widetilde\varepsilon_t^{(i)} \bX_t^i,
\end{align*}
where $\widetilde\varepsilon_t^{(i)} = \bar\varepsilon_t^{(i)}-\mathbb{E}[\bar\varepsilon_t^{(i)}].$ We first bound $\|\bold{G}_t\|_\op.$ By the definition of $\widetilde\varepsilon_t^{(i)}$ and that $\widetilde\varepsilon_t^{(i)}$ are independent with $\bX_t^i$, we have
\begin{align*}
    \|\bold{G}_t\|_\op& = \Big\|\sum_{i=1}^N \widetilde\varepsilon_t^{(i)} \bX_t^i\Big\|_\op
    \leq \max_{k\in[n]}|\widetilde{\varepsilon}_t^k|\Big\|\sum_{i=1}^N \bX_t^i\Big\|_\op
    \leq 2\vartheta\Big\|\sum_{i=1}^N \bX_t^i\Big\|_\op.
\end{align*}
We can then treat $\sum_{i=1}^N \bX_t^i$ as a binary matrix obtained by permuting the columns of a binary diagonal matrix in $\mathbb{R}^{N\times K}.$ Since the norm of a matrix does not depend on the column order, we have
$
    \Big\|\sum_{i=1}^N \bX_t^i\Big\|_\op=1
$
and thus $\|\bold{G}_t\|_\op\leq 2\vartheta.$ 
By an analogous argument, we can also bound $\sigma_Z^2.$ We have
\begin{align*}
    \bold{G}_t\bold{G}_t^\top = \sum_{i=1}^N \sum_{k=1}^N \widetilde\varepsilon_t^{(i)}\widetilde{\varepsilon}_t^{(k)} \bX_t^i\bX_t^{k\top}.
\end{align*}
Notice that for any $i\not =k,$ we have $\bX_t^i\bX_t^{k\top}=\bold{0}$. Additionally, for any realization of $\bX_t,$ we have
\begin{align*}
    \sum_{i=1}^N \bX_t^i\bX_t^{i\top}&=\sum_{i=1}^N e_i(N)e_{j_t(i)}^\top(K)e_{j_t(i)}(K)e_i^\top(N)\\
    &= \sum_{i=1}^N e_i(N)e_i^\top (N)\\
    &= \bold{I}_{N\times N}.
\end{align*}
Thus
\begin{equation}\label{eq: X_top1}
    \left\|\sum_{i=1}^N \bX_t^i\bX_t^{i\top}\right\|_\op = 1,
\end{equation}
and
\begin{equation}\label{eq: X_top2}
    \left\|\sum_{i=1}^N \bX_t^{i\top}\bX_t^i\right\|_\op = 1.
\end{equation}
Then
\begin{equation*}
        \bold{G}_t\bold{G}_t^\top = \sum_{i=1}^N \sum_{k=1}^N \widetilde\varepsilon_t^{(i)}\widetilde{\varepsilon}_t^{(k)} \bX_t^i\bX_t^{k\top} = \sum_{i=1}^N (\widetilde\varepsilon_t^{(i)})^2\bX_t^i\bX_t^{i\top}
\end{equation*}
and
\begin{align*}
    \mathbb{E}\left[\| \bold{G}_t\bold{G}_t^\top \|_\op\right]&= \mathbb{E}\left[\Big\|\sum_{i=1}^N (\widetilde\varepsilon_t^{(i)})^2\bX_t^i\bX_t^{i\top}\Big\|_\op\right] \\
   & \leq \max_{i\in[N]}\big\{\mathbb{E}[(\widetilde\varepsilon_t^{(i)})^2]\big\}\mathbb{E}\left[\Big\|\sum_{i=1}^N \bX_t^i\bX_t^{i\top}\Big\|_\op\right]\\
   &\leq \sigma^2\mathbb{E}\left[\Big\|\sum_{i=1}^N \bX_t^i\bX_t^{i\top}\Big\|_\op\right]\\
   &\leq \sigma^2,
\end{align*}
where the first inequality is due to that $\widetilde\varepsilon_t^{(i)}$ and $\bX_t^i$ are independent with each other, the second inequality is given by $\mathbb{E}[(\widetilde\varepsilon_t^{(i)})^2]\leq \mathbb{E}[(\bar\varepsilon_t^{(i)})^2]\leq \mathbb{E}[(\varepsilon_t^{(i)})^2]\leq \sigma^2.$ Similarly we have $\mathbb{E}\left[\| \bold{G}_t^\top\bold{G}_t \|_\op\right]\leq \sigma^2.$ Thus, $$\sigma_Z^2\leq n\max_{t\in[n]}\{\mathbb{E}[\|\bold{G}_t\bold{G}_t^{\top}\|_\op],\mathbb{E}[\|\bold{G}_t^\top\bold{G}_t\|_\op]\}\leq n\sigma^2.\Halmos$$
\endproof

\subsection{Proof of Proposition~\ref{lem:step3}}
\label{sec: proof_of_rsc}

\proof{Proof of Proposition~\ref{lem:step3}.} Recall that, for any $t\in [n]$,
\begin{equation}\label{def: X_t}
    \bX_t=\sum_{i=1}^N \bX_t^i. 
\end{equation}
 We will prove Proposition~\ref{lem:step3} by a standard peeling argument. Let 
$$w_{\zeta}=\sup_{\bold{\Delta}\in\mathcal{C}_{\alpha}(r,\zeta)}\left|\sum_{t=1}^{\sizeone}\sum_{i=1}^N \langle \bX_t^i,\bold{\Delta}\rangle^2-\|\bold{\Delta}\|_{L^2(\Pi)}^2\right|,$$
where
$$\mathcal{C}_{\alpha}(r,\zeta)=\left\{\bold{\Delta}\in\mathcal{C}_{\alpha}(r)\,\middle|\,\frac{\zeta}{2}\leq \|\bold{\Delta}\|_{L^2(\Pi)}^2<\zeta\right\}.$$ 
First notice that 
\begin{equation}\label{eq:hadtrick}
    \sum_{t=1}^{\sizeone}\sum_{i=1}^N \langle \bX_t^i,\bold{\Delta}\rangle^2=\sum_{t=1}^{\sizeone}\langle \bX_t, \bold{\Delta}\circ\bold{\Delta}\rangle.
\end{equation}
Then by Lemma~\ref{lem: massart_inequality}, we have
\begin{equation}\label{ineq: prob_w_zeta}
    \mathbb{P}\left[w_{\zeta}>2\mathbb{E}[w_\zeta]+\frac{7\zeta}{24}\right]\leq \exp\left(-\frac{\zeta}{288N}\right)
\end{equation}
since 
$$\mathcal{C}_{\alpha}(r,\zeta)\subseteq \left\{\bold{\Delta}\in\mathcal{C}_{\alpha}(r)\,\middle|\, \|\bold{\Delta}\|_{L^2(\Pi)}^2<\zeta\right\}.$$
Now we provide an upper bound for $\mathbb{E}[w_\zeta].$ For $t\in[\sizeone]$, let $\xi_t$ be independent Rademacher random variables. 
By the symmetrization inequality,
\begin{align}
\mathbb{E}[w_\zeta]&=\mathbb{E}\left[\sup_{\bold{\Delta}\in\mathcal{C}(r,\zeta)}\left|\sum_{t=1}^{\sizeone}\sum_{i=1}^N \langle \bX_t^i,\bold{\Delta}\rangle^2-\|\bold{\Delta}\|_{L^2(\Pi)}^2\right|\right]\nonumber\\
&\leq 2\mathbb{E}\bigg[\sup_{\bold{\Delta}\in\mathcal{C}(r,\zeta)} \sum_{t=1}^{\sizeone}\xi_t(\sum_{i=1}^N \langle \bX_t^i,\bold{\Delta}\rangle^2) \bigg] \nonumber\\
&= 2\mathbb{E}\bigg[\sup_{\bold{\Delta}\in\mathcal{C}(r,\zeta)} \sum_{t=1}^{\sizeone}\xi_t\langle \bX_t,\bold{\Delta}\circ\bold{\Delta}\rangle\bigg] \nonumber\\
&= 2\mathbb{E}\bigg[\sup_{\bold{\Delta}\in\mathcal{C}(r,\zeta)}\langle\bold{\Sigma}_R,\bold{\Delta}\circ\bold{\Delta}\rangle\bigg]\nonumber,
\end{align}
where $\bold{\Sigma}_R=\sum_{t=1}^n \xi_t\bX_t.$
Notice that Lemma~\ref{lem: hadamard-product} gives $\rank(\bold{\Delta}\circ\bold{\Delta})\leq r^2,$ and thus we have $\|\bold{\Delta}\circ\bold{\Delta}\|_*\leq \sqrt{\rank(\bold{\Delta}\circ\bold{\Delta})}\|\bold{\Delta}\|_F\leq \sqrt{r^2}\|\bold{\Delta}\|_F.$ In addition, since $\|\bold{\Delta}\|_\infty\leq 1,$ we have $\|\bold{\Delta}\circ\bold{\Delta}\|_F\leq \|\bold{\Delta}\|_F.$ Then it holds that
\begin{align}
 \mathbb{E}\big[\sup_{\bold{\Delta}\in\mathcal{C}(r,\zeta)} \langle\bold{\Sigma}_R,\bold{\Delta}\circ\bold{\Delta}\rangle\big]&\leq \sup_{\bold{\Delta}\in\mathcal{C}(r,\zeta)} \|\bold{\Delta}\circ\bold{\Delta}\|_*\mathbb{E}\left[\|\bold{\Sigma}_R\|_\op\right]\nonumber\\
 &\leq \sup_{\bold{\Delta}\in\mathcal{C}(r,\zeta)}\sqrt{r^2}\|\bold{\Delta}\circ\bold{\Delta}\|_F\mathbb{E}\left[\|\bold{\Sigma}_R\|_\op\right]\nonumber\\
 &\leq \sup_{\bold{\Delta}\in\mathcal{C}(r,\zeta)}\sqrt{r^2}\|\bold{\Delta}\|_F\mathbb{E}\left[\|\bold{\Sigma}_R\|_\op\right]\nonumber\\
 &\leq r\sqrt{{K}{\pmin}\|\bold{\Delta}\|_{L^2(\Pi)}^2}\mathbb{E}\left[\|\bold{\Sigma}_R\|_\op\right]\nonumber\\
 &\leq r\sqrt{K\pmin\zeta}\mathbb{E}\left[\|\bold{\Sigma}_R\|_\op\right]\nonumber\\
 &\leq 4\pmin r^2 K \left(\mathbb{E}\left[\|\bold{\Sigma}_R\|_\op\right]\right)^2+\frac{\zeta}{16}\nonumber,
\end{align}
where the first inequality is given by the duality between nuclear norm and operator norm (i.e., $|\langle\bold{P},\bold{Q}\rangle|\leq \|\bold{P}\|_*\cdot\|\bold{Q}\|_\op,\forall \,\bold{P},\bold{Q}\in\mathbb{R}^{N\times K}$).
Plugging the above into \eqref{ineq: prob_w_zeta} gives us
\begin{equation}
\label{ineq: peeling}
\mathbb{P}\left[w_\zeta > 4\pmin r^2K \left(\mathbb{E}\left[\|\bold{\Sigma}_R\|_\op\right]\right)^2+\frac{11\zeta}{24}\right]\leq \exp\left(\frac{-\zeta}{288N}\right).
\end{equation}
Lemma~\ref{lem: bound_of_rademacher} shows that
\begin{equation*}
    \mathbb{E}[\|\bold{\Sigma}_R\|_\op]\leq {3\sqrt{\sizeone{\log[(N+K)\sizeone]}}},
\end{equation*}
and thus
\begin{align}\label{ineq: varphi}
    4\pmin r^2K\big[\mathbb{E}[\|\bold{\Sigma}_R\|_\op]\big]^2\leq {36\pmin r^2K\sizeone\log[(N+K)\sizeone]}.
\end{align}
Now set $$\varphi={36\pmin r^2K\sizeone\log[(N+K)\sizeone]}$$
and define the bad event
\begin{equation*}
    \mathcal{B}=\left\{\exists \bold{\Delta}\in\mathcal{C}_{\alpha}(r)\quad \text{s.t.} \quad \left|\sum_{t=1}^{\sizeone}\sum_{i=1}^N\langle \bX_t^i, \bold{\Delta}\rangle^2-\|\bold{\Delta}\|_{L^2(\Pi)}^2\right|\geq \frac{11}{24}\|\bold{\Delta}\|_{L^2(\Pi)}^2+\varphi\right\}.
\end{equation*}
Notice that if event $\mathcal{B}$ holds with small probability, then our argument goes. For any $l\in\mathbb{N}^+,$ define 
\begin{equation*}
    \mathcal{B}_l=\left\{\exists \bold{\Delta}\in\mathcal{C}\Big(r,{{576N\alpha}}\cdot 2^l\Big) \quad \text{s.t.} \quad \left|\sum_{t=1}^{\sizeone}\sum_{i=1}^N\langle \bX_t^i, \bold{\Delta}\rangle^2-\|\bold{\Delta}\|_{L^2(\Pi)}^2\right|\geq \frac{11}{24}\|\bold{\Delta}\|_{L^2(\Pi)}^2+\varphi \right\}.
\end{equation*}
According to \eqref{ineq: peeling} and \eqref{ineq: varphi}, we have
$$\mathbb{P}[\mathcal{B}_l]\leq \exp\big(-2\alpha\cdot 2^l\big).$$
Note that $\mathcal{B}\subseteq\bigcup_{l=1}^\infty \mathcal{B}_l,$ which implies
\begin{align*}
    \mathbb{P}[\mathcal{B}]&\leq \sum_{l=1}^{\infty}\mathbb{P}[\mathcal{B}_l]
    \leq\sum_{l=1}^{\infty} \exp\big(-2\alpha\cdot 2^l\big)
    \leq \sum_{l=1}^\infty \exp\big(-2\alpha\cdot l\big)
    \leq \frac{\exp(-2\alpha)}{1-\exp(-2\alpha)}\leq \exp(-\alpha). \Halmos
\end{align*}
\endproof

\begin{lemma}\label{lem: bound_of_rademacher}
    For any $t\in[\sizeone],$ let $\xi_t$ be i.i.d. Rademacher random variables and let $$\bold{\Sigma}_R=\sum_{t=1}^{\sizeone} \xi_t\bX_t$$ where $\bX_t$ satisfies \eqref{def: X_t}.  Then we have
    \begin{align*}
        \mathbb{E}\left[\|\bold{\Sigma}_R\|_\op\right]&\leq {3\sqrt{{\sizeone\log[(N+K)\sizeone]}}}{}.
    \end{align*}
\end{lemma}
\proof{Proof of Lemma~\ref{lem: bound_of_rademacher}}
Our proof techniques rely on the concentration inequality for matrix Rademacher series in Lemma~\ref{lem: matrix_rademacher}. Thus we need to first calculate the value of $\sigma_Z^2.$ Our approach is similar to the argument in the proof of Lemma~\ref{claim: G_t}. First note that 
\begin{align*}
   \left\|\bX_t\bX_t^\top\right\|_\op=\left\|\sum_{i=1}^N\sum_{j=1}^N \bX_t^i\bX_t^{j\top}\right\|_\op&=\left\|\sum_{i=1}^N \bX_t^i\bX_t^{i\top}\right\|_\op
\end{align*}
since $\bX_t^i\bX_t^{j\top}=\bold{0},\forall i\not =j.$ By \eqref{eq: X_top1} we have
    \begin{equation*}
        \left\|\sum_{t=1}^{\sizeone} \bX_t\bX_t^\top\right\|_\op\leq \sizeone\cdot \max_{t\in[\sizeone]}\|\bX_t\bX_t^\top\|_\op =\sizeone.
    \end{equation*}
    Similarly by \eqref{eq: X_top2} we have
    \begin{equation*}
        \left\|\sum_{t=1}^{\sizeone} \bX_t^\top\bX_t\right\|_\op\leq \sizeone\cdot \max_{t\in[\sizeone]}\|\bX_t^\top\bX_t\|_\op =\sizeone.
    \end{equation*}
    By applying Lemma~\ref{lem: matrix_rademacher} with $\sigma_Z^2=\sizeone$, we have
    \begin{align*}
        \mathbb{P}\left[\Big\|\sum_{t=1}^{\sizeone}\xi_t\bX_t\Big\|_\op\geq \rho\right]\leq (N+K)\exp\left(\frac{-\rho^2}{2\sizeone}\right),\forall \rho>0.
    \end{align*}
    Set $\rho=\sqrt{2\sizeone\log[(N+K)\sizeone^{1/2}]}$ and we obtain
    \begin{align*}      \mathbb{P}\left[\bigg\|\sum_{t=1}^{\sizeone}\xi_t\bX_t\bigg\|_\op\geq \sqrt{2\sizeone\log[(N+K)\sizeone^{1/2}]}\right]\leq\frac{1}{\sizeone^{1/2}}.
    \end{align*}
    We also have that $\big\|\sum_{t=1}^{\sizeone}\xi_t\bX_t\big\|_\op \leq \sizeone$ almost surely. Therefore, we have
    \begin{align*}
        \mathbb{E}[\|\bold{\Sigma}_R\|_\op]&=\mathbb{E}\bigg[\|\bold{\Sigma}_R\|_\op\mathbbm{1}\left\{\|\bold{\Sigma}_R\|_\op\geq \sqrt{{2\sizeone\log[(N+K)\sizeone^{1/2}]}}\right\}\bigg]
        \\&\quad+\mathbb{E}\bigg[\|\bold{\Sigma}_R\|_\op\mathbbm{1}\left\{\|\bold{\Sigma}_R\|_\op<\sqrt{{2\sizeone\log[(N+K)\sizeone^{1/2}]}}\right\}\bigg]\\
        &\leq \mathbb{P}\Big[\|\bold{\Sigma}_R\|_\op\geq \sqrt{{2\sizeone\log[(N+K)\sizeone^{3/2}]}}\Big]\times \sizeone+\sqrt{{2\sizeone\log[(N+K)\sizeone^{3/2}]}}\\
        &\leq \sqrt{\sizeone}+{\sqrt{{2\sizeone\log[(N+K)\sizeone^{1/2}]}}}\leq{3\sqrt{\sizeone{\log[(N+K)\sizeone]}}}.    \Halmos
    \end{align*}

\section{Proof of Theorem~\ref{theorem: lower_bound}}\label{sec: prrof_of_lower_bound}
\proof{Proof of Theorem~\ref{theorem: lower_bound}.}
Suppose that $\Pi_t$ is the uniform distribution on $\mathcal{M}$ for all $t\in[n].$ Consider a set $\{\bTheta_1,\bTheta_2,\cdots, \bTheta_{n(\varsigma)}\}$ where $\bTheta_i\in\mathcal{C},\|\bTheta_i\|_F\leq \varsigma ,\forall i\in[n(\varsigma)]$ and $\|\bTheta_i-\bTheta_j\|_F\geq \varsigma, \forall i\not=j.$ We first choose index $m\in[n(\varsigma)]$ uniformly at random, and we are given the observations $\mathcal{S}=\{(\bX_t^i, Y_t^{(i)}),t\in[\samplesize],i\in[N]\}$ sampled according to \eqref{model: data-generating} with $\bTheta^*=\bTheta_m.$ Suppose we obtain an estimation of $m$ by samples in $\mathcal{S}$, denoted by $\widehat{m}.$ Then we have
\[\mathbb{P}\left[\|{\enhancedest}-\bTheta^*\|_F\geq\frac{\varsigma}{2}\right]\geq \mathbb{P}[\widehat{m}\not=m]\]
by the triangle inequality. The Fano's inequality (see, e.g., Theorem~9 of \citet{scarlett2019introductory}) yields that
\begin{equation}\label{eq: hatm_notequal_m}
    \mathbb{P}[\widehat{m}\not=m\mid \mathcal{X}_1,\cdots,\mathcal{X}_{\samplesize}]\geq 1-\frac{\max_{i,j\in[n(\varsigma)],i\not =j} \operatorname{D}(\bTheta_i \parallel \bTheta_j)+\log 2}{\log \big(n(\varsigma)\big)},
\end{equation}
where $
    \operatorname{D}(\bTheta_i \parallel \bTheta_j)$
is the KL divergence between the distributions of $$(Y_1,Y_2,\cdots, Y_{\samplesize}\mid \mathcal{X}_1,\cdots,\mathcal{X}_{\samplesize},\bTheta_i)\quad\text{and}\quad(Y_1,Y_2,\cdots, Y_{\samplesize}\mid \mathcal{X}_1,\cdots,\mathcal{X}_{\samplesize},\bTheta_j).$$
For Gaussian noises with variance $\sigma^2$, we have, for any $i\not =j,$
\begin{equation*}
   \operatorname{D}(\bTheta_i \parallel \bTheta_j)=\frac{1}{2\sigma^2}\sum_{t=1}^{\samplesize} \|\mathcal{X}_t(\bTheta_i)-\mathcal{X}_t(\bTheta_j)\|^2
\end{equation*}
and thus
\begin{align*}
     \mathbb{E}[\operatorname{D}(\bTheta_i \parallel \bTheta_j)]&=\frac{n}{2K\sigma^2}\|\bTheta_i-\bTheta_j\|_F^2
\end{align*}
where the expectation is taken over $(\mathcal{X}_1,\cdots,\mathcal{X}_n).$
Then by \eqref{eq: hatm_notequal_m}, we have
\begin{align*}
    \mathbb{P}[\widehat{m}\not=m]&\geq 1-\frac{ \frac{n}{2K\sigma^2}\|\bTheta_i-\bTheta_j\|_F^2+\log 2}{\log \big(n(\varsigma)\big)}\\
    & \geq 1-\frac{\frac{2n\varsigma^2}{K\sigma^2}+\log 2}{\log \big(n(\varsigma)\big)}
\end{align*}
where the last inequality is given by $\|\bTheta_i-\bTheta_j\|_F\leq \|\bTheta_i\|_F+\|\bTheta_j\|_F\leq2\varsigma .$
By the proof of Theorem~5 in \citet{koltchinskii2011nuclear}, for any constant $0<\upsilon\leq 1,$ when 
\[\varsigma = \upsilon (\sigma \wedge 1)\sqrt{\frac{rK^2}{n}},\] 
there exists $\{\bTheta_1,\bTheta_2,\cdots, \bTheta_{n(\varsigma)}\}$
with $n(\varsigma)\geq 2^{rK/8}+1$ that satisfies $\|\bTheta_i\|_F=\varsigma,\bTheta_i\in\mathcal{C}, \forall i\in[n(\varsigma)]$ and $\|\bTheta_i-\bTheta_j\|_F\geq \varsigma, \forall i\not=j.$ 
Then choosing $\upsilon=\sqrt{\log 2}/8$ yields
\begin{align*}
    \mathbb{P}[\widehat{m}\not=m] \geq 1- \frac{\frac{2\upsilon^2(\sigma^2\wedge 1)rK^2}{K\sigma^2}+\log 2}{\frac{rK\log 2}{8}}
   \geq 1-\frac{2\upsilon^2rK+\log 2}{\frac{rK\log 2}{8}}
    =1-\frac{rK/4+8}{rK}
    \geq \frac{1}{2},
\end{align*}
when $rK\geq 32.$
Thus, we have
\begin{align*}
    \ell(\bTheta^*, \|\cdot\|_F) &= \inf_{\widetilde{\bTheta}}\sup_{\bTheta^*\in\mathcal{C}}\mathbb{E}\left[\|\widetilde{\bTheta}-\bTheta^*\|_F \right]\\
    & \geq \inf_{\widetilde{\bTheta}}\sup_{\bTheta^*\in\mathcal{C}}\frac{\varsigma}{2}\mathbb{P}\left[\|{\enhancedest}-\bTheta^*\|_F\geq\frac{\varsigma}{2}\right]\\
    &\geq \inf_{\widetilde{\bTheta}}\frac{\varsigma}{2}\mathbb{P}\left[\hat{m}\not =m\right]\\
    &\geq \frac{\varsigma}{4}=\frac{\sqrt{\log 2}(\sigma \wedge 1)}{8}\sqrt{\frac{rK^2}{n}},
\end{align*}
which completes our proof. 
\section{Proof of Theorem~\ref{theorem: second_stage_error_bound}}\label{sec: proof_of_second_stage}
\subsection{Major Steps of the Proof}
We first define the conditional number for any matrix $\bTheta,$ as $\kappa = \sigma_{\max}(\bTheta)/\sigma_{\min}(\bTheta)$.
The very first step of our proof will be showing that the number of samples collected for each row and column in the enhancement procedure is proportional to the total number of enhancement samples $\sizetwo=|\mathcal{J}_2|$. To establish this, we let 
\[
\minsample =\min\left\{\min_{i\in[N]}|\mathcal{R}_i|,\min_{j\in[K]}|\mathcal{C}_j|\right\}
\]
denote the minimum sample size of each row $i\in[N]$ and each column $j\in[K]$ in the enhancement procedure, and we have the following lemma which is proved in Appendix~\ref{app:proof_stage_two_sample}:

\begin{lemma}\label{lem: stage_two_sample}
     If 
    \begin{equation*}
    \sizetwo\geq \secondlowerbound,
\end{equation*}
    then
    \begin{equation*}
        \minsample\geq \frac{\sizetwo N}{2\pmin K}
    \end{equation*}
    with probability greater than $1-\exp(-\alpha)$ for any $\alpha>0$.
\end{lemma}

Given the minimum sample size requirement, we are now able to derive Theorem \ref{theorem: second_stage_error_bound}. We break our proof into two steps.

The first step (Lemma~\ref{lem: row_enhancement}) shows that with sufficient samples at each row and column, the $\widetilde{\bU}$ and $\widetilde{\bV}$ returned by the row enhancement procedure in Algorithm~\ref{alg: db_enhancement} will better approximate $\bU^*\bold{D}^*$ and $\bV^*\bold{D}^*$ (recall the SVD $\bTheta^* = \bU^*\bD^*\bV^{*\top}$); specifically, the row-wise distance, i.e., $\ell_{2, \infty}$ norm error, is improved. 
This improvement is enabled by two least square regressions in the double-enhancement procedure Algorithm~\ref{alg: db_enhancement}. The proof technique for this lemma is similar to that in \cite{hamidi2019personalizing}. However, as aforementioned, they are only interested in the row-wise error $\|{\enhancedest}-\bTheta^*\|_{2,\infty}$, and hence only discuss how to enhance $\widehat \bV$ to better approximate the row space of $\bTheta^*$ but not the column space (i.e., $\widehat\bU$). In contrast, our analysis requires a simultaneous enhancement of both $\widetilde{\bV}$ and $\widetilde{\bU},$ which is more involved. The proof of Lemma~\ref{lem: row_enhancement} is in Appendix~\ref{app:proof_row_enhance}.

\begin{lemma}\label{lem: row_enhancement}
 Suppose there exists $\epsilon_f>0$ such that the first stage estimator ${\nuclearest}$ has
\begin{equation}\label{eq: cond_row1}
    \frac{\big\|\bTheta^*-\nuclearest\big\|_F}{\sqrt{NK}}\leq \epsilon_f\leq \frac{\|\bTheta^*\|_{\infty}}{2\eta\kappa\sqrt{r}};
\end{equation}
besides, assume that the minimum row and column sample size has
    \begin{equation}\label{cond: row_2}
         \minsample\geq \frac{64\kappa^2 \eta^2 r\log r\log (N+K)}{\|\bTheta^*\|_{\infty}^2}.
    \end{equation}
   Then, the $\widetilde{\bU},\widetilde{\bV}$ returned by Algorithm~\ref{alg: db_enhancement} satisfy
    \begin{equation*}
        \|\widetilde{\bU}-\bU^*\bold{D}^*\|_{2,\infty}\leq \left(\frac{15\epsilon_f\kappa\eta\sqrt{r}}{2\|\bTheta^*\|_\infty}+\frac{4\sqrt{2\alpha}\eta^2\kappa r\sigma }{\sqrt{\minsample}{\|\bTheta^*\|_\infty^2}}\right)\|\bU^*\bold{D}^*\|_{2,\infty}
    \end{equation*}
    and
    \begin{equation*}
        \|\widetilde{\bV}-\bV^*\bold{D}^*\|_{2,\infty}\leq \left(\frac{15\epsilon_f\kappa\eta\sqrt{r}}{2\|\bTheta^*\|_\infty}+\frac{4\sqrt{2\alpha}\eta^2\kappa r\sigma }{\sqrt{\minsample}{\|\bTheta^*\|_\infty^2}}\right)\|\bV^*\bold{D}^*\|_{2,\infty}
    \end{equation*}
    with probability greater than $1-3(N+K)\exp(-\alpha)$ for any $\alpha>0$.
\end{lemma}

Under the sample size requirement in Theorem~\ref{theorem: second_stage_error_bound}, $\epsilon_f$ in the above lemma is of order $\widetilde{\mathcal{O}}(r\sqrt{K/N n})$ according to Theorem~\ref{theorem:errorbound} given our choice of $\lambda$, which will be helpful in establishing the condition in our second key step.

The second step (Lemma~\ref{lem: infinity_norm}) shows that, by setting ${\enhancedest} = \bU_1\bold{Q}_1\widetilde{\bV}^\top,$ the entry-wise error $\|{\enhancedest}-\bTheta^*\|_{\infty}$ is controlled by the row-wise bounds $\|\widetilde{\bU}-\bU^*\bold{D}^*\|_{2,\infty}$ and $\|\widetilde{\bV}-\bV^*\bold{D}^*\|_{2,\infty}$ derived in Lemma \ref{lem: row_enhancement}. This bound is deterministic, leading towards our final probabilistic bound in Theorem \ref{theorem: second_stage_error_bound}. The proof of Lemma~\ref{lem: infinity_norm} is provided in Appendix~\ref{sec: proof_of_entry_second}.

\begin{lemma} \label{lem: infinity_norm}
Suppose there exists $\epsilon<1/(2\sqrt{r}\eta\kappa)$ such that 
\begin{equation}\label{cond: deltaV}
\|\widetilde{\bU}-\bU^*\bold{D}^*\|_{2,\infty}\leq\epsilon \|\bU^*\bold{D}^*\|_{2,\infty} \quad\text{and}\quad
    \|\widetilde{\bV}-\bV^*\bold{D}^*\|_{2,\infty}\leq\epsilon \|\bV^*\bold{D}^*\|_{2,\infty}.
\end{equation} 
Then, ${\enhancedest} = \bU_1\bold{Q}_1\widetilde{\bV}^\top$ satisfies
\begin{equation*}
    \|{\enhancedest}-\bTheta^*\|_\infty \leq\frac{33\epsilon\sqrt{r}\eta\kappa\|\bTheta^*\|_\infty}{4},
\end{equation*}
where $\bU_1,\bold{Q}_1$ are from the SVD $\widetilde{\bU}=\bU_1\bold{D}_1\bold{Q}_1.$
\end{lemma}
The $\epsilon$ in Lemma~\ref{lem: infinity_norm} exists with probability higher than $1-3(N+K)\exp(-\alpha)$ for any $\alpha>0$ if we set
\[
\epsilon = \frac{15\epsilon_f\kappa\eta\sqrt{r}}{2\|\bTheta^*\|_\infty}+\frac{4\sqrt{2\alpha}\eta^2\kappa r\sigma }{\sqrt{\minsample}{\|\bTheta^*\|_\infty^2}},
\]
 according to our results in Lemma~\ref{lem: row_enhancement}.
As aforementioned, $\epsilon_f$ is of order $\widetilde{\mathcal{O}}(r\sqrt{K/N n})$, thus $\epsilon$ is of order $\widetilde{\mathcal{O}}(r^{3/2}\sqrt{K/N n}).$ 
By Lemma~\ref{lem: infinity_norm}, this yields our final result order of $\widetilde{\mathcal{O}}(r^2\sqrt{K/N n}).$

\subsection{Proof of Theorem~\ref{theorem: second_stage_error_bound}}
\proof{Proof of Theorem~\ref{theorem: second_stage_error_bound}.}
First we provide the explicit form of our condition $n=\widetilde{\Omega}\left(\max\{(r^4K)/N, (K/N)^2\}\right);$ that is,
\begin{multline*}
    n\geq\max\Bigg\{\secondlowerbound+1, \frac{256\pmin\kappa^2\eta^2rK\log r\log(N+K)}{\|\bTheta^*\|_{\infty}^2N}+1, \frac{1800c_4^2c_\lambda^2\sigma^2\pmin^2 \eta^4\kappa^4r^3K\big(\alpha+\log(N+K)\big)^2}{N\|\bTheta^*\|_{\infty}^2},\\\frac{1800c_4^2 \eta^4\pmin^2\kappa^4r^4K \log[(N+K)n]}{N\|\bTheta^*\|_{\infty}^2},\frac{2048\alpha\sigma^2\pmin\eta^6\kappa^4r^3K}{N\|\bTheta^*\|_{\infty}^2}+1\Bigg\},
\end{multline*}
where $c_4$ is the constant from \eqref{ineq: thm1explicit}. This requirement for sample size is to ensure that the conditions in Lemma~\ref{lem: row_enhancement} and Lemma~\ref{lem: infinity_norm} hold with high probability. Specifically, we will show that if the sample size requirement is met, then \begin{equation}\label{lower_bound_on_nmin}
    \minsample\geq \frac{\sizetwo N}{2\pmin K}\geq \frac{(\samplesize-1) N}{4\pmin K}\geq \frac{64\kappa^2 \eta^2 r\log r\log (N+K)}{\|\bTheta^*\|_{\infty}^2}
\end{equation}
with probability higher than $1-\exp(-\alpha)$ and 
\begin{equation}\label{upper_bound_of_epsilonf}
    \frac{\|\bTheta^*-\widehat{\bTheta}\|_F}{\sqrt{NK}}\leq\epsilon_f\leq\frac{\|\bTheta^*\|_{\infty}}{30r\eta^2\kappa^2}\leq \frac{\|\bTheta^*\|_{\infty}}{2\eta\kappa\sqrt{r}} 
\end{equation} with probability higher than $1-4\exp(-\alpha)$ by choosing \[\epsilon_f=c_4\pmin \max\left\{c_\lambda\sigma\left(\alpha+\log(N+K)\right)\sqrt{\frac{rK}{N\sizeone}},r\sqrt{\frac{K\log[(N+K)\sizeone]}{N\sizeone}}\right\}.\] 
We first show \eqref{lower_bound_on_nmin} holds with high probability. By Lemma~\ref{lem: stage_two_sample} and our assumption that \[n\geq \frac{256\kappa^2\eta^2\pmin rK\log r\log(N+K)}{N\|\bTheta^*\|_{\infty}^2}+1,\] we immediately have \eqref{lower_bound_on_nmin} holds with probability higher than $1-\exp(-\alpha).$ Then we will show that \eqref{upper_bound_of_epsilonf} holds with high probability via Theorem~\ref{theorem:errorbound}. The explicit form of Theorem~\ref{theorem:errorbound} in \eqref{ineq: thm1explicit} tells us with probability higher than $1-4\exp(-\alpha),$
\begin{equation*}
    \frac{\big\|\bTheta^*-\nuclearest\big\|_F}{\sqrt{NK}}\leq c_4\pmin\max\left\{c_\lambda\sigma\left(\alpha+\log(N+K)\right)\sqrt{\frac{rK}{Nn_1}}, r\sqrt{\frac{K\log[(N+K)n_1]}{Nn_1}}\right\}.
\end{equation*}
Since \[n_1\geq \frac{n}{2}\geq \max\left\{\frac{900c_4^2c_\lambda^2\sigma^2\pmin^2 \eta^4\kappa^4r^3K\big(\alpha+\log(N+K)\big)^2}{N\|\bTheta^*\|_{\infty}^2},\frac{900c_4^2 \pmin^2 \eta^4\kappa^4r^4K \log[(N+K)n]}{N\|\bTheta^*\|_{\infty}^2}\right\},\]
by choosing 
\begin{equation*}
    \epsilon_f=c_4\pmin\max\left\{c_\lambda\sigma\left(\alpha+\log(N+K)\right)\sqrt{\frac{rK}{Nn_1}},r\sqrt{\frac{K\log[(N+K)n_1]}{Nn_1}}\right\},
\end{equation*}
we have
\begin{equation*}
    \frac{\big\|\bTheta^*-\nuclearest\big\|_F}{\sqrt{NK}}\leq\epsilon_f\leq\frac{\|\bTheta^*\|_{\infty}}{30r\eta^2\kappa^2}\leq \frac{\|\bTheta^*\|_{\infty}}{2\eta\kappa\sqrt{r}} 
\end{equation*} with probability higher than $1-4\exp(-\alpha),$
where the last inequality is given by $\eta, \kappa, r\geq 1.$ 

By our choice of $\epsilon_f$ and sufficient sample size established in \eqref{lower_bound_on_nmin} and \eqref{upper_bound_of_epsilonf}, we  have the conditions in Lemma~\ref{lem: row_enhancement} hold. Therefore,
\begin{equation*}
        \|\widetilde{\bU}-\bU^*\bold{D}^*\|_{2,\infty}\leq \left(\frac{15\epsilon_f\kappa\eta\sqrt{r}}{2\|\bTheta^*\|_\infty}+\frac{4\sqrt{2\alpha}\eta^2\kappa r\sigma }{\sqrt{\minsample}{\|\bTheta^*\|_\infty^2}}\right)\|\bU^*\bold{D}^*\|_{2,\infty}
    \end{equation*}
    
    \begin{equation*}
        \|\widetilde{\bV}-\bV^*\bold{D}^*\|_{2,\infty}\leq \left(\frac{15\epsilon_f\kappa\eta\sqrt{r}}{2\|\bTheta^*\|_\infty}+\frac{4\sqrt{2\alpha}\eta^2\kappa r\sigma }{\sqrt{\minsample}{\|\bTheta^*\|_\infty^2}}\right)\|\bV^*\bold{D}^*\|_{2,\infty}
    \end{equation*}
    with probability higher than $1-(3N+3K)\exp(-\alpha).$  
This further implies that, as long as 
    \begin{equation}\label{cond: condonepsilon}
        \frac{15\epsilon_f\kappa\eta\sqrt{r}}{2\|\bTheta^*\|_\infty}+\frac{4\sqrt{2\alpha}\eta^2\kappa r\sigma }{\sqrt{\minsample}{\|\bTheta^*\|_\infty^2}}\leq \frac{1}{2\sqrt{r}\eta\kappa},
    \end{equation}
then the condition in Lemma~\ref{lem: infinity_norm} is satisfied with probability higher than $1-(3N+3K)\exp(-\alpha)$ by choosing 
\[\epsilon=\frac{15\epsilon_f\kappa\eta\sqrt{r}}{2\|\bTheta^*\|_\infty}+\frac{4\sqrt{2\alpha}\eta^2\kappa r\sigma }{\sqrt{\minsample}{\|\bTheta^*\|_\infty^2}}.\] 
Now we show that \eqref{cond: condonepsilon} holds. By \eqref{lower_bound_on_nmin} and
\[n\geq\frac{2048\alpha\sigma^2\pmin \eta^6\kappa^4r^3K}{N\|\bTheta^*\|_{\infty}^2}+1, \]
we have
\begin{equation*}
    n_{\min}\geq \frac{(n-1)N}{4\pmin K}\geq \frac{512\alpha\sigma^2\eta^6\kappa^4r^3}{\|\bTheta^*\|_{\infty}^2};
\end{equation*}
Thus, it gives
\begin{equation*}
    \frac{4\sqrt{2\alpha}\eta^2\kappa r\sigma }{\sqrt{\minsample}{\|\bTheta^*\|_\infty^2}}\leq \frac{1}{4\sqrt{r}\eta\kappa}.
\end{equation*}
By combining this argument with \eqref{upper_bound_of_epsilonf}, we have
\begin{equation*}
    \frac{15\epsilon_f\kappa\eta\sqrt{r}}{2\|\bTheta^*\|_\infty}+\frac{4\sqrt{2\alpha}\eta^2\kappa r\sigma }{\sqrt{\minsample}{\|\bTheta^*\|_\infty^2}}\leq \frac{15\frac{\|\bTheta^*\|_{\infty}}{30r\eta^2\kappa^2}\kappa\eta\sqrt{r}}{2\|\bTheta^*\|_\infty}+\frac{1}{4\sqrt{r}\eta\kappa}\leq\frac{1}{2\sqrt{r}\eta\kappa},
\end{equation*}
which implies that \eqref{cond: condonepsilon} holds. 

Now by our choice of $\epsilon,$ the condition in Lemma~\ref{lem: infinity_norm} holds with probability higher than $1-(3N+3K)\exp(-\alpha).$ We have
\begin{align*}
\epsilon&=\frac{15\epsilon_f\kappa\eta\sqrt{r}}{2\|\bTheta^*\|_\infty}+\frac{4\sqrt{2\alpha}\eta^2\kappa r\sigma }{\sqrt{\minsample}{\|\bTheta^*\|_\infty^2}}\\
&=\frac{15\kappa\eta\sqrt{r}}{2\|\bTheta^*\|_\infty}\left(c_4\max\left\{c_\lambda\sigma\left(\alpha+\log(N+K)\right)\sqrt{\frac{rK}{Nn_1}},r\sqrt{\frac{K\log[(N+K)n_1]}{Nn_1}}\right\}\right)+\frac{4\sqrt{2\alpha}\eta^2\kappa r\sigma }{\sqrt{\minsample}{\|\bTheta^*\|_\infty^2}}\\
&\leq \frac{15\kappa\eta\sqrt{r}}{2\|\bTheta^*\|_\infty}\left(c_4\pmin\max\left\{c_\lambda\sigma\left(\alpha+\log(N+K)\right)\sqrt{\frac{2 rK}{Nn}},r\sqrt{\frac{2K\log[(N+K)\sizeone]}{Nn}}\right\}\right)+\frac{4\sqrt{2\alpha}\eta^2\kappa r\sigma }{{\|\bTheta^*\|_\infty^2}}\sqrt{\frac{4\pmin K}{N(n-1)}}\\
&\leq c_5\frac{\kappa\eta r}{\|\bTheta^*\|_{\infty}}\left(\pmin\max\left\{c_\lambda\sigma\left(\alpha+\log(N+K)\right)\sqrt{\frac{K}{Nn}},\sqrt{\frac{rK\log[(N+K)n]}{Nn}}\right\}+\frac{\eta\sigma\sqrt{\alpha}}{\|\bTheta^*\|_{\infty}}\sqrt{\frac{\pmin K}{Nn}}\right),
\end{align*}
where $c_5$ is an absolute constant.
Plugging the above into Lemma~\ref{lem: infinity_norm} yields 
\begin{align}
&\|{\enhancedest}-\bTheta^*\|_\infty  \nonumber \\
&\leq \frac{33\epsilon\sqrt{r}\eta\kappa\|\bTheta^*\|_{\infty}}{4}\nonumber\\
&\leq \frac{33\sqrt{r}\eta\kappa\|\bTheta^*\|_{\infty}}{4}\cdot c_5\frac{\kappa\eta r}{\|\bTheta^*\|_{\infty}}\left(\pmin\max\left\{c_\lambda\sigma\left(\alpha+\log(N+K)\right)\sqrt{\frac{K}{Nn}},\sqrt{\frac{rK\log[(N+K)n]}{Nn}}\right\}+\frac{\eta\sigma\sqrt{\alpha}}{\|\bTheta^*\|_{\infty}}\sqrt{\frac{\pmin K}{Nn}}\right)\nonumber\\
&\leq c_6{\eta^2\kappa^2r^{3/2}}\max\Bigg\{c_\lambda\sigma\pmin\left(\alpha+\log(N+K)\right)\sqrt{\frac{K}{Nn}},\pmin\sqrt{\frac{rK\log[(N+K)n]}{Nn}},\frac{\eta\sigma\sqrt{\alpha}}{\|\bTheta^*\|_{\infty}}\sqrt{\frac{\pmin K}{Nn}}\Bigg\},
\label{thm: infinity_explicit}
\end{align}
where $c_6$ is an absolute constant.

Since \eqref{lower_bound_on_nmin} and \eqref{upper_bound_of_epsilonf} hold with probability higher than $1-5\exp(-\alpha),$ we have \eqref{thm: infinity_explicit} holds with probability higher than $1-(3N+3K+5)\exp(-\alpha),$ which completes the proof.
\Halmos

\endproof

\subsection{Proof of Lemma~\ref{lem: stage_two_sample}}
\label{app:proof_stage_two_sample}
\proof{Proof of Lemma~\ref{lem: stage_two_sample}.}
Let $r_i =|\mathcal{R}_i|$ and $c_i =|\mathcal{C}_j|$ where $\mathcal{R}_i$ and $\mathcal{C}_j$ are set in Algorithm~\ref{alg: db_enhancement} and $i\in [N], j\in[K]$. By the definition of matching, we have $r_i = \sizetwo,\forall i\in [N].$ Then, it suffices to show that $$\min_{j\in[K]}c_j\geq \frac{N}{2\pmin K\sizetwo}$$ with high probability. We have for any $j\in[K],$
\begin{align*}
    c_j = \left|\{t\in\mathcal{J}_2, j_t(i)=j \}\right|
    =  \sum_{t\in\mathcal{J}_2}\sum_{i\in[N]}\indicator\{j_t(i)=j\}.
\end{align*}
By our matching structure, for any $t\in\mathcal{J}_2$, $\sum_{i\in[N]}\indicator\{j_t(i)=j\}$ are independent Bernoulli random variables.
Then, by Hoeffding inequality, we have
\begin{align*}
    \mathbb{P}\left[\left|c_j-\mathbb E\left[\sum_{t\in\mathcal{J}_2}\sum_{i\in[N]}\indicator\{j_t(i)=j\}\right]\right|\geq \frac{\sizetwo N}{2\pmin K}\right]
    \leq 2\exp\left(\frac{-2\left(\frac{\sizetwo N}{2\pmin K}\right)^2}{\sizetwo}\right)
    \leq 2\exp\left(-\frac{\sizetwo N^2}{2\pmin^2K^2}\right)
    \leq 2\exp(-\alpha),
\end{align*}
where the last inequality is given by the condition $\sizetwo\geq 2K^2\alpha/(\pmin^2N^2)$ of this lemma. Further note that
\[\mathbb E\left[\sum_{t\in\mathcal{J}_2}\sum_{i\in[N]}\indicator\{j_t(i)=j\}\right]\geq |\mathcal{J}_2|\frac{N}{\pmin K}=\frac{N\sizetwo}{\pmin K},\]
which completes the proof.

\subsection{Proof of Lemma~\ref{lem: row_enhancement}}
\label{app:proof_row_enhance}
\proof{Proof of Lemma~\ref{lem: row_enhancement}.}
We will prove the result for $\widetilde{\bU},$ and the result for $\widetilde{\bV}$ can be obtained via an analogous argument. 
    We first recall some notations for this proof. We let $e_i(N)$ for $i\in[N]$ denote the canonical basis vector in \( \mathbb{R}^N \); that is, \( e_i(N) \) is a column vector with 1 in the \( i \)-th entry and 0 in the other entries. Similarly, $e_j(K)$ for $j\in [K]$ denotes the basis vector in \( \mathbb{R}^K \). 

    Let ${\beta^\top= \bU^{*(i,\cdot)}\bD^* =e_i(N)^\top\bU^*\bD^* \in \mathbb{R}^r}$ for abbreviation. Recall that in Algorithm~\ref{alg: db_enhancement}, we use $\mathcal{R}_i$ to denote $\{\big(\bX_{t}^i,Y_{t}^{(i)}\big) \mid t\in\mathcal{J}_2\}$, which is a set of enhancement samples from row $i$. Define $r_i=|\mathcal{R}_i|$ and index the elements in $\mathcal{R}_i$ as
    $$(\bold{Z}_k, y_k), k\in[r_i].$$
    Then, define
    \begin{equation*}
    {\bold{Z}}=\left[\begin{array}{c}
           e_i(N)^{\top}\bold{Z}_1\\
           \vdots \\
           e_i(N)^{\top}\bold{Z}_{r_i}
    \end{array}\right] \in \mathbb{R}^{r_i\times K},
\end{equation*}
and the response vector
\begin{equation*}
    Y = \begin{bmatrix}
      y_1 & \cdots & y_{r_i}  
    \end{bmatrix}^{\top}.
\end{equation*}
By \eqref{model: data-generating}, we have
\begin{equation*}
Y=\bold{Z}\bV^*\beta+\varepsilon,
\end{equation*}
where $\varepsilon$ is a vector in $\mathbb{R}^{r_i}$ consisting of independent $\sigma$-subgaussian noises.
Then the solution $\widetilde{\beta}_i$ in Algorithm~\ref{alg: db_enhancement} is equivalent to
\begin{equation}\label{def:estimator}
    \widetilde{\beta}_i\in\argmin_{\gamma\in\mathbb{R}^{r}}\|Y-\bold{Z}{\nuclearestV}\gamma\|^2.
\end{equation}
Define
\begin{equation*}
    \bold{H}=\bold{Z}{\nuclearestV}.
\end{equation*}
Assume that $\bH^\top\bH$ is invertible, which implies that the unique solution for \eqref{def:estimator} is
\begin{equation*}
\argmin_{\gamma\in\mathbb{R}^{r}}\|Y-\bold{Z}{\nuclearestV}\gamma\|^2 = \big(\bold{H}^\top\bold{H}\big)^{-1}\bold{H}^\top {Y}.
\end{equation*}
We use 
$\widetilde{\beta}=\big(\bold{H}^\top\bold{H}\big)^{-1}\bold{H}^\top {Y}$ to denote this solution for ease of notation.
We will show later in \eqref{ineq: prob_of_invertible} that $\bH^\top\bH$ is invertible with high probability. Now we proceed to analyze $\|\widetilde{\beta}-\beta\|.$ We have
\begin{align*}
    \widetilde{\beta}&=(\bH^\top\bH)^{-1}\bH^\top {Y}\\
    &=   (\bH^\top\bH)^{-1}\bH^\top (\bold{Z}\bV^*\beta+\varepsilon)\\
    &= (\bH^\top\bH)^{-1}\bH^\top (\bold{Z}{\nuclearestV}\beta+\bold{Z}\bV^*\beta-\bold{Z}{\nuclearestV}\beta+\varepsilon)\\
    &= (\bH^\top\bH)^{-1}\bH^\top \bH\beta+(\bH^\top\bH)^{-1}\bH^\top \bold{Z}(\bV^*-{\nuclearestV})\beta+(\bH^\top\bH)^{-1}\bH^\top\varepsilon\\
    & = \beta +(\bH^\top\bH)^{-1}\bH^\top \bold{Z}(\bV^*-{\nuclearestV})\beta+(\bH^\top\bH)^{-1}\bH^\top\varepsilon.
\end{align*}
So 
\begin{align}
    \|\widetilde{\beta}-\beta\| &= \left\|(\bH^\top\bH)^{-1}\bH^\top \bold{Z}(\bV^*-{\nuclearestV})\beta+(\bH^\top\bH)^{-1}\bH^\top\varepsilon\right\| \nonumber\\
    &\leq \left\|(\bH^\top\bH)^{-1}\bH^\top \bold{Z}(\bV^*-{\nuclearestV})\beta\right\|+\left\|(\bH^\top\bH)^{-1}\bH^\top\varepsilon\right\| \nonumber\\
    &\leq \left\|(\bH^\top\bH)^{-1}\bH^\top \bold{Z}\right\|_\op\left\|\bV^*-{\nuclearestV}\right\|_\op \left\|\beta\right\|+\left\|(\bH^\top\bH)^{-1}\bH^\top\varepsilon\right\| \nonumber\\
    &\leq \underbrace{\big\|(\bH^\top\bH)^{-1}\bH^\top \bold{Z}\big\|_{\op}}_{=h_1}\big\|\bV^*-{\nuclearestV}\big\|_F\big\|\beta\big\|+\underbrace{\left\|(\bH^\top\bH)^{-1}\bH^\top\varepsilon\right\|}_{=h_2}.\label{ineq: beta_gap}
\end{align}
We first bound $h_2.$ We have
\begin{align}
    h_2&=\Big\|\big(\bold{H}^\top\bold{H}\big)^{-1}\bold{H}^\top\varepsilon\Big\|\nonumber\\
    &\leq \Big\|\big(\bold{H}^\top\bold{H}\big)^{-1}\Big\|_\op\|\bold{H}^\top\varepsilon\|\nonumber\\
    &=\frac{\big\|\bold{H}^\top\varepsilon\big\|}{\lambda_{\min}(\bold{H}^\top\bold{H})}.\label{eq:h_2}
\end{align}
So we will upper bound $h_2$ by upper bounding $\bH^\top\varepsilon$ and lower bounding $\lambda_{\min}(\bold{H}^\top\bold{H}).$ 
We will use Lemma~\ref{lem: subgaussian_vector_sequence} on the norm of the weighted sum of subgaussian vectors to upper bound $\|\bH^\top\varepsilon\|.$ Notice that
\begin{align*}
   \|\bH^\top\varepsilon\| = \|\varepsilon^\top\bH\|=\left\|\sum_{k=1}^{r_i} \varepsilon^{(k)}   e_i(N)^{\top}\bold{Z}_k{\nuclearestV}\right\|
\end{align*}
and each random vector $e_i(N)^{\top}\bold{Z}_k{\nuclearestV}\in\mathbb{R}^r$ follows an independent and uniform distribution over the set $\{e_j(K)^\top{\nuclearestV},j\in[K]\},$ thus
$\|e_i(N)^{\top}\bold{Z}_k{\nuclearestV}\|\leq \|{\nuclearestV}\|_{2,\infty},\forall k\in[r_i].$ Then by Lemma~\ref{lem: subgaussian_vector_sequence}, we have for any $\rho>0,$
\begin{align*}
    \mathbb{P}\left[\left\|\sum_{k=1}^{r_i} \varepsilon^{(k)}  e_i(N)^{\top}\bold{Z}_k{\nuclearestV}\right\|\geq \rho\right]\leq2\exp\left(\frac{-\rho^2}{2r_ir\|{\nuclearestV}\|_{2,\infty}^2\sigma^2}\right) \leq 2\exp\left(\frac{-\rho^2}{2\minsample r\|{\nuclearestV}\|_{2,\infty}^2\sigma^2}\right)
\end{align*}
where the second inequality is given by the definition $\minsample=\min\{\min_{i\in[N]}|\mathcal{R}_i|,\min_{j\in[K]}|\mathcal{C}_j|\}.$
Taking $\rho=\sigma\sqrt{2r\alpha \minsample}\|{\nuclearestV}\|_{2,\infty}$ yields
\begin{align*}
    & \mathbb{P}\left[\big\|\bH^\top \varepsilon\big\|\geq \sigma\sqrt{2r\alpha \minsample}\|{\nuclearestV}\|_{2,\infty}\right]\leq 2\exp(-\alpha).
\end{align*}
We will then use the matrix Chernoff bound in Lemma~\ref{lem: matrix_chernoff} to lower bound $\lambda_{\min}(\bold{H}^\top\bold{H}).$ Note that by Weyl's inequality (Lemma~\ref{lem:Weryl's theorem}),
\begin{align*}
    \lambda_{\min}(\bold{H}^\top\bold{H})&=\lambda_{\min}\left(\sum_{k=1}^{r_i}{\nuclearestV}^\top \bold{Z}_k^\top e_i(N)e_i(N)^{\top}\bold{Z}_k{\nuclearestV}\right)\\
    &\geq \lambda_{\min}\left(\sum_{k=1}^{\minsample}{\nuclearestV}^\top \bold{Z}_k^\top e_i(N)e_i(N)^{\top}\bold{Z}_k{\nuclearestV}\right).
\end{align*}
 By our previous argument that each random vector $e_i(N)^{\top}\bold{Z}_k{\nuclearestV}\in\mathbb{R}^r$ follows an independent and uniform distribution over the set $\{e_j(K)^\top{\nuclearestV},j\in[K]\}$, we have
 \[\lambda_{\max}({\nuclearestV}^\top \bold{Z}_k^\top e_i(N)e_i(N)^{\top}\bold{Z}_k{\nuclearestV}) = \|e_i(N)^{\top}\bold{Z}_k{\nuclearestV}\|^2\leq \|{\nuclearestV}\|_{2,\infty}^2 ,\forall k\in[r_i]\]
and
\begin{align*}
    \lambda_{\min}\left(\mathbb{E}\left[{\nuclearestV}^\top \bold{Z}_k^\top e_i(N)e_i(N)^{\top}\bold{Z}_k{\nuclearestV}\right]\right)&=\lambda_{\min}\left(\sum_{j=1}^K \frac{1}{K} {\nuclearestV}^\top e_j(K)e_j(K)^\top{\nuclearestV}\right)\\
    &=\frac{1}{K}\lambda_{\min}\left[{\nuclearestV}^\top \left(\sum_{j=1}^K e_j(K)e_j(K)^\top\right){\nuclearestV}\right]\\
    &=\frac{1}{K}\lambda_{\min}\left({\nuclearestV}^{\top}\bold{I}_{K\times K}{\nuclearestV}\right)\\
    &=\frac{1}{K}\lambda_{\min}(\bold{I}_{r\times r})=\frac{1}{K}.
\end{align*}
 Thus, plugging $\mu_{\min}=\minsample/K$ and  $\omega=\|{\nuclearestV}\|_{2,\infty}^2$ into Lemma~\ref{lem: matrix_chernoff}, we obtain
\begin{align*}
    \mathbb{P}\left[\lambda_{\min}\Big(\sum_{k=1}^{\minsample}{\nuclearestV}^\top \bold{Z}_k^\top e_i(N)e_i(N)^{\top}\bold{Z}_k{\nuclearestV}\Big)\leq \frac{\rho \minsample}{K}\right]\leq r\exp\left(\frac{-(1-\rho)^2\minsample}{2K\|{\nuclearestV}\|_{2,\infty}^2}\right),\forall \rho\in(0,1).
\end{align*}
Then, set $\rho=1/2.$ If $\minsample\geq 8\alpha K\|{\nuclearestV}\|_{2,\infty}^2\log r$, we have 
\begin{align}
    \mathbb{P}\bigg[\lambda_{\min}(\bH^\top\bH)\leq  \frac{\minsample}{2K}\bigg]
    &\leq \mathbb{P}\left[\lambda_{\min}\left(\sum_{k=1}^{\minsample}{\nuclearestV}^\top \bold{Z}_k^\top e_i(N)e_i(N)^{\top}\bold{Z}_k{\nuclearestV}\right)\leq \frac{\minsample}{2K}\right]\nonumber\\
    &\leq r\exp\left(\frac{-\minsample}{8K\|{\nuclearestV}\|_{2,\infty}^2}\right)\leq \exp(-\alpha)\label{ineq: prob_of_invertible},
\end{align}
which implies that $\bH^\top\bH$ is invertible with probability higher than $1-\exp(-\alpha),$
and furthermore,
\begin{align}
    h_2\leq \frac{\|\bH^\top\varepsilon\|}{\lambda_{\min}(\bH^\top \bH)}&\leq \frac{\sigma\sqrt{2r\alpha \minsample}\|{\nuclearestV}\|_{2,\infty}}{\frac{\minsample}{2K}}\nonumber\\ &=\frac{2K\sigma\sqrt{2r\alpha}\|{\nuclearestV}\|_{2,\infty}}{\sqrt{\minsample}\|\bU^*\bD^*\|_{2,\infty}}\|\bU^*\bD^*\|_{2,\infty}\nonumber\\
    &=\frac{2K\sigma\sqrt{2r\alpha}\|{\nuclearestV}\|_{2,\infty}}{\sqrt{\minsample}\|\bTheta^*\|_{2,\infty}}\|\bU^*\bD^*\|_{2,\infty}\nonumber\\
    &\leq \frac{2K\sigma\sqrt{2r\alpha}\|{\nuclearestV}\|_{2,\infty}}{\sqrt{\minsample}\frac{\|\bTheta^*\|_F}{\sqrt{N}}}\|\bU^*\bD^*\|_{2,\infty}\nonumber\\
    &\leq  \frac{2K\sigma\sqrt{2r\alpha}\|{\nuclearestV}\|_{2,\infty}}{\sqrt{\minsample}\frac{\sqrt{NK}\|\bTheta^*\|_\infty}{\eta\sqrt{N}}}\|\bU^*\bD^*\|_{2,\infty}\nonumber\\
    &=\frac{2\eta\sigma\sqrt{2Kr\alpha}\|{\nuclearestV}\|_{2,\infty}}{\sqrt{\minsample}{\|\bTheta^*\|_\infty}}\|\bU^*\bD^*\|_{2,\infty}\label{ineq:h_2}
\end{align}
with probability higher than $1-3\exp(-\alpha),$
where the second equation is given by \eqref{eq: UDtwoinfty} in Lemma~\ref{lem: algebra_inequalities}, and the last inequality is given by Assumption~\ref{assump: spikiness}. In the following lemma, we provide an upper bound on $\|{\nuclearestV}\|_{2,\infty}$ and thus $\minsample\geq  8\alpha K\|{\nuclearestV}\|_{2,\infty}^2\log r$ must hold as long as
\begin{equation*}
        \minsample\geq \frac{64\kappa^2 \eta^2 r\log r\log (N+K)}{\|\bTheta^*\|_{\infty}^2}\quad\text{and}\quad\epsilon_f\leq \frac{\|\bTheta^*\|_{\infty}}{2\eta\kappa\sqrt{r}}.
    \end{equation*}
\begin{lemma}\label{claim: scale_of_V'}
If
    \begin{equation}\label{cond: epsilon_f}
    \epsilon_f\leq\frac{\|\bTheta^*\|_{\infty}}{2\eta\kappa\sqrt{r}},
\end{equation}
then
\begin{equation*}
    \sigma_{\min}({\nuclearest})\geq \frac{1}{2\sigma_{\min}(\bTheta^*)}, \quad\text{and}\quad\|{\nuclearestV}\|_{2,\infty}\leq \frac{2\kappa\eta}{\|\bTheta^*\|_\infty}\sqrt{\frac{r}{K}}\,.
\end{equation*}
\end{lemma}

We leave the proof of this lemma at the end of this subsection.
Combining Lemma~\ref{claim: scale_of_V'} and \eqref{ineq:h_2}, when \eqref{cond: epsilon_f} holds, we have
\begin{align*}
    h_2&\leq \frac{2\eta\sigma\sqrt{2Kr\alpha}\|{\nuclearestV}\|_{2,\infty}}{\sqrt{\minsample}{\|\bTheta^*\|_\infty}}\|\bU^*\bD^*\|_{2,\infty}\\
    &\leq \frac{2\eta\sigma\sqrt{2Kr\alpha}\frac{2\kappa\eta}{\|\bTheta^*\|_\infty}\sqrt{\frac{r}{K}}}{\sqrt{\minsample}{\|\bTheta^*\|_\infty}}\|\bU^*\bD^*\|_{2,\infty}\\
&=\frac{4\sqrt{2\alpha}\eta^2\kappa r\sigma }{\sqrt{\minsample}{\|\bTheta^*\|_\infty^2}}\|\bU^*\bD^*\|_{2,\infty}
\end{align*}
with probability higher than $1-3\exp(-\alpha).$ 

The term $h_1$ can be bounded via basic algebra. Notice that
\begin{align*}
    h_1=\big\|(\bold{H}^\top\bold{H}\big)^{-1}\bold{H}^\top \bold{Z}\big\|_\op
    &= \big\|(\bold{H}^\top\bold{H}\big)^{-1}\bold{H}^\top \bold{Z}{\nuclearestV}\big\|_\op\\
    &=\big\|(\bold{H}^\top\bold{H}\big)^{-1}\bold{H}^\top \bold{H}\big\|_\op\\
    &= 1,
\end{align*}
where the first equality is due to Lemma~\ref{lem: semi-ortho matrix}.
Plugging the above bounds of $h_1$ and $h_2$ into \eqref{ineq: beta_gap} gives us
\begin{align}
    \|\widetilde{\beta}-\beta\|&\leq {h_1}\big\|\bV^*-{\nuclearestV}\big\|_F\big\|\beta\big\|+{h_2}\nonumber\\
    &\leq \big\|\bV^*-{\nuclearestV}\big\|_F\big\|\beta\big\|+h_2\nonumber\\
    &\stackrel{(a)}{\leq }5\|\bTheta^*-{\nuclearest}\|_F\left(\frac{1}{\sigma_{\min}(\bTheta^*)}+\frac{1}{\sigma_{\min}({\nuclearest})}\right)\|\beta\|+h_2\nonumber\\
    &\stackrel{(b)}{\leq }\frac{15\epsilon_f\sqrt{NK}}{2\sigma_{\min}(\bTheta^*)}\|\beta\|+h_2\nonumber\\
    &\stackrel{(c)}{\leq} \frac{15\epsilon_f\kappa\eta\sqrt{r}}{2\|\bTheta^*\|_\infty}\|\beta\|+h_2\nonumber\\
    &\leq \left(\frac{15\epsilon_f\kappa\eta\sqrt{r}}{2\|\bTheta^*\|_\infty}+\frac{4\sqrt{2\alpha}\eta^2\kappa r\sigma }{\sqrt{\minsample}{\|\bTheta^*\|_\infty^2}}\right)\|\bU^*\bD^*\|_{2,\infty}\label{ineq: rowibound}
\end{align}
with probability higher than $1-3\exp(-\alpha)$, if \eqref{eq: cond_row1} and \eqref{cond: row_2} hold. In detail, (a) is given by Lemma~\ref{lem:optimal_rotation}, (b) is given by \eqref{eq: cond_row1} and Lemma~\ref{claim: scale_of_V'},
 and (c) is given by \eqref{ineq: sigmamin_lower} in Lemma~\ref{lem: algebra_inequalities}. 
By our definition, $\widetilde{\beta}-\beta=\left(\widetilde{\bU}-\bU^*\bD^*\right)^{(i,\cdot)}.$ Extending \eqref{ineq: rowibound} to every $i\in[N]$, we have
\begin{equation*}
   \left\| \left(\widetilde{\bU}-\bU^*\bD^*\right)^{(i,\cdot)}\right\|\leq \left(\frac{15\epsilon_f\kappa\eta\sqrt{r}}{2\|\bTheta^*\|_\infty}+\frac{4\sqrt{2\alpha}\eta^2\kappa r\sigma }{\sqrt{\minsample}{\|\bTheta^*\|_\infty^2}}\right)\|\bU^*\bD^*\|_{2,\infty} 
\end{equation*}
 with probability higher than $1-3\exp(-\alpha), \forall i\in[N].$
Since $\|\widetilde{\bU}-\bU^*\bold{D}^*\|_{2,\infty}=\max_{i\in[N]} \left\| \left(\widetilde{\bU}-\bU^*\bD^*\right)^{(i,\cdot)}\right\|,$
by union probability, we have
\begin{align*}
      \mathbb{P}\left[  \|\widetilde{\bU}-\bU^*\bold{D}^*\|_{2,\infty}\leq \left(\frac{15\epsilon_f\kappa\eta\sqrt{r}}{2\|\bTheta^*\|_\infty}+\frac{4\sqrt{2\alpha}\eta^2\kappa r\sigma }{\sqrt{\minsample}{\|\bTheta^*\|_\infty^2}}\right)\|\bU^*\bold{D}^*\|_{2,\infty}\right]\geq 1-3N\exp(-\alpha).
\end{align*}
Similarly, we can also show that 
\begin{equation*}
       \mathbb{P}\left[ \|\widetilde{\bV}-\bV^*\bold{D}^*\|_{2,\infty}\leq \left(\frac{15\epsilon_f\kappa\eta\sqrt{r}}{2\|\bTheta^*\|_\infty}+\frac{4\sqrt{2\alpha}\eta^2\kappa r\sigma }{\sqrt{\minsample}{\|\bTheta^*\|_\infty^2}}\right)\|\bV^*\bold{D}^*\|_{2,\infty}\right]\geq 1-3K\exp(-\alpha).
    \end{equation*}
Therefore our argument goes.
\Halmos

\proof{Proof of Lemma~\ref{claim: scale_of_V'}.}
We first prove the bound for $\sigma_{\min}({\nuclearest}).$ By the condition in the claim statement and \eqref{ineq: sigmamin_lower}, we have
\begin{align*}
    \epsilon_f\leq \frac{\|\bTheta^*\|_{\infty}}{2\eta\kappa\sqrt{r}}\leq\frac{\sqrt{KN}\sigma_{\min}(\bTheta^*)}{2}.
\end{align*}
By Lemma~\ref{lem: matrix_pertubation}, we have
\begin{align*}
    \sigma_{\min}({\nuclearest})\geq \sigma_{\min}(\bTheta^*)-\|\bTheta^*-{\nuclearest}\|_F
    &\geq \sigma_{\min}(\bTheta^*)-\epsilon_f\sqrt{KN}\\
    &\geq \sigma_{\min}(\bTheta^*)-\frac{\sigma_{\min}(\bTheta^*)}{2}=\frac{\sigma_{\min}(\bTheta^*)}{2}.
\end{align*}
Hence by Lemma~\ref{lem: row-bound} and \eqref{ineq: sigmamin_lower}, we have
\begin{align*}
    \|{\nuclearestV}\|_{2,\infty}\leq\frac{\sqrt{N}\|{\nuclearest}\|_{\infty}}{\sigma_{\min}({\nuclearest})}\leq \frac{2\sqrt{N}\|{\nuclearest}\|_{\infty}}{\sigma_{\min}(\bTheta^*)}\leq \frac{2\sqrt{N}\|{\nuclearest}\|_{\infty}}{\frac{\sqrt{NK}\|\bTheta^*\|_{\infty}}{\sqrt{r}\eta\kappa}}
    &\leq \frac{2\kappa\eta}{\|\bTheta^*\|_{\infty}}\sqrt{\frac{r}{K}},
\end{align*}
where the last inequality is given by the constraint in \eqref{program:nonconvex}.\Halmos

\subsection{Proof of Lemma~\ref{lem: infinity_norm}}\label{sec: proof_of_entry_second}

\proof{Proof of Lemma~\ref{lem: infinity_norm}.}
First we show that as long as $ \epsilon\leq\frac{1}{2\sqrt{r}\eta\kappa},$ we have $\rank(\widetilde{\bU})=r.$ Then it suffices to show that 
$\sigma_r(\widetilde{\bU})>0.$ By Weyl's inequality (Lemma~\ref{lem:Weryl's theorem}) and the definition of $\epsilon$ in \eqref{cond: deltaV}, we have $\forall i\in[r]$
\begin{align*}
    |\sigma_i(\bD_1)-\sigma_i(\bTheta^*)|=|\sigma_i(\widetilde{\bU})-\sigma_i(\bU^*\bD^*)|&\leq \|\widetilde{\bU}-\bU^*\bD^*\|_F\\
    &\leq \sqrt{N}\|\widetilde{\bU}-\bU^*\bD^*\|_{2,\infty}\\
    &\leq \epsilon\sqrt{NK}\|\bTheta^*\|_\infty.
\end{align*}
Then by \eqref{ineq: sigmamin_lower}, we have
\begin{align}
    \sigma_r(\widetilde{\bU})=\sigma_{r}(\bD_1)&\geq \sigma_{\min}(\bTheta^*)-|\sigma_{r}(\bold{D}_1)-\sigma_{\min}(\bTheta^*)|\nonumber\\
    &\geq \frac{\sqrt{NK}\|\bTheta^*\|_{\infty}}{\sqrt{r}\eta\kappa}- \epsilon\sqrt{NK}\|\bTheta^*\|_\infty\nonumber\\
    &=\left(\frac{1}{\sqrt{r}\eta\kappa}-\epsilon\right)\sqrt{NK}\|\bTheta^*\|_\infty\geq \max\left\{\frac{\epsilon}{2},\frac{1}{2\sqrt{r}\eta\kappa}\right\}\sqrt{NK}\|\bTheta^*\|_\infty,
    \label{ineq: sigma_r_D1}
\end{align}
where the last inequality is given by the condition 
$\epsilon\leq\frac{1}{2\sqrt{r}\eta\kappa}.$
\eqref{ineq: sigma_r_D1} shows that $\bD_1$ is a full rank diagonal matrix in $\mathbb{R}^{r\times r}$, and thus $\bD_1^{-1}$ is well-defined. Then we can write
\begin{align*}
    {\enhancedest} &= \bU_1\bold{Q}_1\widetilde{\bV}^\top\\
    &= ({\bU_1}\bold{D}_1\bold{Q}_1)\bold{Q}_1^\top\bold{D_1}^{-1}\bold{Q}_1\widetilde{\bV}^\top\\
    &= ({\bU^*}\bold{D}^*+\underbrace{\widetilde{\bU}-{\bU}^*\bold{D}^*}_{=\bold{\Delta}_U})(\bold{D}^{*-1}+\underbrace{\bold{Q}_1^\top\bold{D}_1^{-1}\bold{Q}_1-\bold{D}^{*-1}}_{=\bold{\Delta}_D})(\bold{D}^*\bV^{*\top}+\underbrace{\widetilde{\bV}^\top-\bold{D}^*\bV^{*\top}}_{=\bold{\Delta}_V^\top}).
\end{align*}
We first bound the three terms $\bold{\Delta}_U$, $\bold{\Delta}_D$ and $\bold{\Delta}_V$ respectively.
Lemma~\ref{lem: matrix_inverse} and \eqref{ineq: sigma_r_D1} yield
\begin{align}
    \|\bold{\Delta}_D\|_\op&\leq \frac{\max_{i\in[r]} |\sigma_i(\bold{D}_1)-\sigma_i(\bold{D}^*)|}{\sigma_{\min}(\bold{D}_1)\sigma_{\min}(\bold{D}^*)}\nonumber\\
    &\leq \frac{\epsilon\sqrt{NK}\|\bTheta^*\|_\infty}{\max\left\{\frac{\epsilon}{2},\frac{1}{2\sqrt{r}\eta\kappa}\right\}\sqrt{NK}\|\bTheta^*\|_\infty\sigma_{\min}(\bTheta^*)}\nonumber\\
    &\leq \min\left\{\frac{2}{\sigma_{\min}(\bTheta^*)},\frac{2\sqrt{r}\eta\kappa\epsilon}{\sigma_{\min}(\bTheta^*)}\right\}\label{ineq: deltaDop}.
\end{align}
Additionally, given our conditions in the lemma statement, we have
$$\|\bold{\Delta}_U\|_{2,\infty}=\|\widetilde{\bU}-\bU^*\bold{D}^*\|_{2,\infty}\leq\epsilon \|\bU^*\bold{D}^*\|_{2,\infty}$$
and
\begin{equation}\label{ineq: deltaV}
    \|\bold{\Delta}_V\|_{2,\infty}=\|\widetilde{\bV}-\bV^*\bold{D}^*\|_{2,\infty}\leq\epsilon \|\bV^*\bold{D}^*\|_{2,\infty}.
\end{equation}
Now we decompose ${\enhancedest}-\bTheta^*$ as
\begin{align}
    {\enhancedest}-\bTheta^*&=(\bU^*\bold{D}^*+\bold{\Delta}_U)(\bold{D}^{*-1}+\bold{\Delta}_D)(\bold{D}^*\bV^{*\top}+\bold{\Delta}_V^\top)-\bTheta^* \nonumber\\ 
    &=\bU^*\bold{D}^*(\bold{D}^{*-1}+\bold{\Delta}_D)(\bold{D}^*\bV^{*\top}+\bold{\Delta}_V^\top)+\bold{\Delta}_U(\bold{D}^{*-1}+\bold{\Delta}_D)(\bold{D}^*\bV^{*\top}+\bold{\Delta}_V^\top)-\bTheta^* \nonumber \\ 
    &=\bU^*\bold{D}^*\bold{D}^{*-1}\bold{D}^*\bV^{*\top}-\bTheta^*+\bU^*\bold{D}^*(\bold{D}^{*-1}+\bold{\Delta}_D)\bold{\Delta}_V^\top+\bU^*\bold{D}^*\bold{\Delta}_D(\bold{D}^*\bV^{*\top}+\bold{\Delta}_V^\top)\nonumber \\ 
    &\quad +\bold{\Delta}_U(\bold{D}^{*-1}+\bold{\Delta}_D)(\bold{D}^*\bV^{*\top}+\bold{\Delta}_V^\top)\nonumber \\ 
    &=\underbrace{\bU^*\bold{D}^*(\bold{D}^{*-1}+\bold{\Delta}_D)\bold{\Delta}_V^\top}_{=\vD_1}+\underbrace{\bU^*\bold{D}^*\bold{\Delta}_D(\bold{D}^*\bV^{*\top}+\bold{\Delta}_V^\top)}_{=\vD_2}+\underbrace{\bold{\Delta}_U(\bold{D}^{*-1}+\bold{\Delta}_D)(\bold{D}^*\bV^{*\top}+\bold{\Delta}_V^\top)}_{=\vD_3}.
    \label{eq: decomp}
\end{align}
In what follows, we bound the infinity norm of the three terms $\vD_1,$ $\vD_2$ and $\vD_3$ respectively.

For the term $\vD_1,$ we have the following inequalities:
\begin{align}\label{ineq: two_infinityforboth}
    &\|\bU^*\bD^*\|_{2,\infty}=\|\bTheta^*\|_{2,\infty}\leq \|\bTheta^*\|_{\infty}\sqrt{K}, \quad \text{and} \quad  \|\bV^*\bD^*\|_{2,\infty}=\|\bTheta^{*\top}\|_{2,\infty}\leq \|\bTheta^*\|_{\infty}\sqrt{N},
\end{align}
which are given by \eqref{eq: UDtwoinfty}, and
\begin{align}
    \|\bD^{*-1}+\vD_{D}\|_{\op} \leq \|\bD^{*-1}\|_\op + \|\vD_D\|_\op &= \frac{1}{\sigma_{\min}(\bD^*)} +\|\vD_D\|_\op \nonumber\\
    &\leq \frac{1}{\sigma_{\min}(\bD^*)}+\frac{2}{\sigma_{\min}(\bD^*)}\nonumber\\
    &= {\frac{3}{\sigma_{\min}(\bD^*)}}\leq \frac{3\sqrt{r}\eta\kappa}{\sqrt{NK}\|\bTheta^*\|_{\infty}},\label{ineq: operator_norm}
\end{align}
where the first inequality is given by \eqref{ineq: deltaDop} and the last inequality is given by \eqref{ineq: sigmamin_lower}.
   Then by Lemma~\ref{lem: 2inftynorm} we have
\begin{align*}
    \|\vD_1\|_{\infty}&\leq \|\bU^*\bold{D}^*\|_{2,\infty}\|\bold{D}^{*-1}+\bold{\Delta}_D\|_\op\|\bold{\Delta}_V\|_{2,\infty}\\
    &\leq \|\bU^*\bold{D}^*\|_{2,\infty}\|\bold{D}^{*-1}+\bold{\Delta}_D\|_\op\epsilon\|\bD^*\bV^{*\top}\|_{2,\infty}\\
    &\leq \|\bTheta^*\|_{\infty}\sqrt{K}\cdot\frac{3\sqrt{r}\eta\kappa}{\sqrt{NK}\|\bTheta^*\|_{\infty}}\cdot\epsilon\|\bTheta^*\|_{\infty}\sqrt{N}\\
    &=3\epsilon\sqrt{r}\kappa\eta\|\bTheta^*\|_\infty.
\end{align*}
where the second inequality is given by \eqref{ineq: deltaV}, \eqref{ineq: two_infinityforboth} and \eqref{ineq: operator_norm}. Similarly, we have
\begin{align*}
    \|\vD_3\|_\infty&=\left\|\bold{\Delta}_U(\bold{D}^{*-1}+\bold{\Delta}_D)(\bold{D}^*\bV^{*\top}+\bold{\Delta}_V^\top)\right\|_\infty\\
    & \leq \|\bold{\Delta}_U\|_{2,\infty}\|\bD^{*-1}+\bold{\Delta}_D\|_\op \|\bD^*\bV^{*\top}+\bold{\Delta}_V^\top\|_{2,\infty}\\
    &\leq \epsilon\|\bU^*\bD^*\|_{2,\infty}\|\bD^{*-1}+\bold{\Delta}_D\|_\op (1+\epsilon)\|\bD^*\bV^{*\top}\|_{2,\infty}\\
    &\leq (1+\epsilon)\frac{3\epsilon\sqrt{r}\eta\kappa\|\bTheta^*\|_\infty}{2}\\
    &\leq \left(1+\frac{1}{2\sqrt{r}\eta\kappa}\right)\frac{3\epsilon\sqrt{r}\eta\kappa\|\bTheta^*\|_\infty}{2}\\
    &\leq \frac{9\epsilon\sqrt{r}\eta\kappa\|\bTheta^*\|_\infty}{4},
\end{align*}
where the last inequality is given by $\kappa,r,\eta>1.$
For the term $\vD_2$, we have an analogous argument
\begin{align*}
    \|\vD_2\|_\infty &\leq \|\bU^*\bold{D}^*\|_{2,\infty}\|\bold{\Delta}_D\|_\op\|\bold{D}^*\bV^{*\top}+\bold{\Delta}_V^\top\|_{2,\infty}\\
    &\leq \|\bU^*\bold{D}^*\|_{2,\infty}\|\bold{\Delta}_D\|_\op(1+\epsilon)\|\bold{D}^*\bV^{*\top}\|_{2,\infty}\\
    &\leq (1+\epsilon)\|\bTheta^*\|_{\infty}\sqrt{K}\cdot\min\left\{\frac{2}{\sigma_{\min}(\bTheta^*)},\frac{2\sqrt{r}\eta\kappa\epsilon}{\sigma_{\min}(\bTheta^*)}\right\}\cdot\|\bTheta^*\|_{\infty}\sqrt{N}\\
    &\leq (1+\epsilon)\|\bTheta^*\|_{\infty}\sqrt{K}\cdot\frac{2\sqrt{r}\eta\kappa}{\sqrt{NK}\|\bTheta^*\|_{\infty}}\cdot\|\bTheta^*\|_{\infty}\sqrt{N}\\
    &\leq 3\epsilon\sqrt{r}\eta\kappa\|\bTheta^*\|_{\infty}
\end{align*}
where the third inequality is given by \eqref{ineq: deltaDop}.
Thus we conclude
\begin{align*}
    \|\widetilde{\bTheta}-\bTheta^*\|_\infty &\leq \|\vD_1\|_\infty+\|\vD_2\|_\infty+\|\vD_3\|_\infty  \\
    &\leq\frac{33\epsilon\sqrt{r}\eta\kappa\|\bTheta^*\|_\infty}{4}.\Halmos 
\end{align*}
   
\section{Proofs of Theorem~\ref{theorem: optimal_matching_bandit}}
\label{app: proof_regret_optimal}
\subsection{Major Steps of the Proof}
To prove Theorem~\ref{theorem: optimal_matching_bandit}, we will decompose the regret of Algorithm~\ref{alg: lowrank_bandit} into the regret from the exploration phase and the regret from the exploitation phase. The regret in the exploration phase can be easily upper bounded by $NE_h$ since the per-period regret is less than $N$ for any period $t\in[T],$ regardless of which $\bX_t$ we choose.

To show the regret in the exploitation phase, we first prove Proposition~\ref{col: decision_loss}, which says that the per-period regret for the exploitation phase $\langle \bX^*,\bTheta^*\rangle-\langle \bX_c,\bTheta^*\rangle=\widetilde{O}(r^{3/2}KE_h^{-1/2})$, with probability higher than $1-4(NT)^{-1}.$  We provide the proof in Appendix~\ref{app:proof_col_decision_loss}.

\begin{proposition}\label{col: decision_loss}
Given $\lambda=c_{\lambda} \sigma (\log(NT)+\log(N+K))/\sqrt{E_h}$ in \eqref{program:nonconvex} for a constant $c_{\lambda}$, $\bX_c$ obtained in Algorithm~\ref{alg: lowrank_bandit} after $E_h$ exploration steps satisfies
    \[\langle \bX^*,\bTheta^*\rangle-\langle \bX_c,\bTheta^*\rangle\leq2\sqrt{2r}\delta_F(E_h)\]with probability higher than $1-4(NT)^{-1},$
    where 
    \begin{equation*}\label{def: deltaF}
        \delta_F(E_h)=c_4\pmin\max\left\{c_\lambda\sigma K\left(\alpha+\log(N+K)\right)\sqrt{\frac{r}{E_h}},rK\sqrt{\frac{\log[(N+K)E_h]}{NE_h}}\right\}
    \end{equation*}
    for the constant $c_4$ from \eqref{ineq: thm1explicit}.
\end{proposition}

Equipped with Proposition~\ref{col: decision_loss}, we can show the following lemma that combines the regret from the two phases, which is sufficient for proving Theorem \ref{theorem: optimal_matching_bandit}. The proof for this lemma is in Appendix~\ref{app:proof-mid-step-optimal-regret}.

\begin{lemma}
     Set \[\lambda=c_{\lambda} \sigma \frac{\log(NT)+\log(N+K)}{\sqrt{{E_h}}}\] in  \eqref{program:nonconvex} where $c_{\lambda}$ is a constant.  Then the regret of Algorithm~\ref{alg: lowrank_bandit} satisfies
\begin{equation*}
    \mathbb{E}[R(T)]\leq NE_h+2\sqrt{2r}(T-E_h)\delta_F(E_h)+4.
\end{equation*}\label{lemma: mid-step-optimal-regret}
\end{lemma}

\subsection{Proof of Theorem~\ref{theorem: optimal_matching_bandit}}
\proof{Proof of Theorem~\ref{theorem: optimal_matching_bandit}.} 
By Lemma~\ref{lemma: mid-step-optimal-regret}, choosing $E_h=c_h rT^{2/3}$ where $c_h$ is a suitable constant yields
\begin{align*}
    \mathbb{E}[R(T)]&\leq NE_h+2\sqrt{2r}(T-E_h)\delta_F(E_h)+4\\
    &=c_h rNT^{2/3}+2\sqrt{2r}(T-c_h rT^{2/3})\delta_F(c_h rT^{2/3})+4\\
    &\leq c_h rNT^{2/3}+2\sqrt{2r}T\delta_F(c_h rT^{2/3})+4\\
    & = \widetilde{\mathcal{O}}\left(NrT^{3/2}+r^{1/2}TrK (rT^{2/3})^{-1/2}\right)\\
    & =\widetilde{\mathcal{O}}\left(r(N+K)T^{3/2}\right),
\end{align*}
which completes our proof.
\Halmos
\endproof

\subsection{Proof of Proposition~\ref{col: decision_loss}}
\label{app:proof_col_decision_loss}
\proof{Proof of Proposition~\ref{col: decision_loss}.}We first invoke the explicit form of Theorem~\ref{theorem:errorbound} in \eqref{ineq: thm1explicit} with $\samplesize=E_h$ to get
\begin{equation*}
    \mathbb{P}[\|\bTheta^*-\widehat{\bTheta}\|_F\geq \delta_F(E_h)]\leq 4(NT)^{-1},
\end{equation*}
by our choice of $\lambda.$
Thus
\begin{align*}
   \langle\bX ^*,\bTheta^*\rangle-\langle\bX_c,\bTheta^*\rangle&= \langle\bX ^*,\bTheta^*-\widehat{\bTheta}\rangle+\langle \bX_c,\bTheta^*-\widehat{\bTheta}\rangle\\
    &\leq \big|\langle\bX ^*,\bTheta^*-\widehat
    {\bTheta}\rangle\big|+\big|\langle \bX_c,\bTheta^*-\widehat{\bTheta}\rangle\big|\\
    &\leq  \|\bX ^*\|_{\op}\|\bTheta^*-\widehat{\bTheta}\|_*+ \|\bX_c\|_{\op}\|\bTheta^*-\widehat{\bTheta}\|_*\\
    &=  2\|\bTheta^*-\widehat{\bTheta}\|_*\\
    &\leq 2\sqrt{2r}\|\bTheta^*-\widehat
    {\bTheta}\|_F\leq 2\sqrt{2r}\delta_F(E_h)
\end{align*}
with probability higher than $1-4(NT)^{-1}$ where the second equality is given by the matching structure of $\bX^*$ and $\bX_c$, and the last inequality is given by $\rank (\bTheta^*-\widehat{\bTheta})\leq \rank(\bTheta^*)+\rank(\widehat{\bTheta})\leq 2r.$
\Halmos

\subsection{Proof of Lemma \ref{lemma: mid-step-optimal-regret}}
\label{app:proof-mid-step-optimal-regret}
\proof{Proof of Lemma \ref{lemma: mid-step-optimal-regret}.}
For the first $E_h$ rounds, we have
\begin{equation*}
    \sum_{t=1}^{E_h} \langle\bX ^*,\bTheta^*\rangle-\langle\bX_t,\bTheta^*\rangle\leq \sum_{t=1}^{E_h} \langle\bX ^*,\bTheta^*\rangle\leq \sum_{t=1}^{E_h} N\|\bTheta^*\|_{\infty}\leq NE_h.
\end{equation*}
By Proposition~\ref{col: decision_loss}, we have $\langle \bX^*,\bTheta^*\rangle-\langle \bX_c,\bTheta^*\rangle\leq2\sqrt{2r}\delta_F(E_h)$ with probability higher than $1-4(NT)^{-1}.$ This implies that
\begin{equation*}
    \sum_{t=E_h}^{T} \langle\bX ^*,\bTheta^*\rangle-\langle\bX_t,\bTheta^*\rangle = \sum_{t=E_h}^{T} \langle\bX ^*,\bTheta^*\rangle-\langle\bX_c,\bTheta^*\rangle\leq (T-E_h)2\sqrt{2r}\delta_F(E_h)
\end{equation*}
with probability higher than $1-4(NT)^{-1}.$ Therefore, the total regret
\begin{equation*}
    R(T) = \sum_{t=1}^T\langle\bX ^*,\bTheta^*\rangle-\langle\bX_t,\bTheta^*\rangle\leq NE_h+(T-E_h)2\sqrt{2r}\delta_F(E_h)
\end{equation*}
with probability higher than $1-4(NT)^{-1}.$ Now notice that for any $t\in[T]$ and any $\bX_t\in\mathcal{M},$ we have 
\begin{equation*}
    \langle\bX ^*,\bTheta^*\rangle-\langle\bX_t,\bTheta^*\rangle\leq N,
\end{equation*}
which implies that 
\begin{equation*}
    R(T) = \sum_{t=1}^T\left(\langle\bX ^*,\bTheta^*\rangle-\langle\bX_t,\bTheta^*\rangle\right)\leq NT
\end{equation*}
almost surely. Then we can decompose the regret expectation
\begin{align*}
    \mathbb{E}\left[R(T)\right]& = \mathbb{E}\left[\sum_{t=1}^T\left(\langle\bX ^*,\bTheta^*\rangle-\langle\bX_t,\bTheta^*\rangle\right)\right]\\
    &=\mathbb{E}\left[R(T)\cdot\indicator\left\{R(T)\leq NE_h+(T-E_h)2\sqrt{2r}\delta_F(E_h)\right\}\right]\\
    &\quad +\mathbb{E}\left[R(T)\cdot\indicator\left\{R(T)>NE_h+(T-E_h)2\sqrt{2r}\delta_F(E_h)\right\}\right]\\
    &\leq NE_h+(T-E_h)2\sqrt{2r}\delta_F(E_h) +NT\cdot\mathbb{P}\left[R(T)>NE_h+(T-E_h)2\sqrt{2r}\delta_F(E_h)\right]\\
    &\leq NE_h+(T-E_h)2\sqrt{2r}\delta_F(E_h)+NT\cdot 4(NT)^{-1}\\
    & = NE_h+(T-E_h)2\sqrt{2r}\delta_F(E_h)+4,
\end{align*}
which completes our proof. 
\Halmos
\endproof

\section{Proof of Theorem~\ref{theorem: stable_bandit}}
\label{app: proof_regret_stable}
\subsection{Major Steps of the Proof}
Following the literature \citep{gale1962college}, we assume there is no tie in preferences without loss of generality. A key ingredient for the proof of Theorem~\ref{theorem: stable_bandit} is that a stable matching is determined solely based on the preference rankings of both workers and jobs; in other words, accurate reward estimation is not critical as long as the rankings derived from it are precise. When the minimum reward gap between pairs is sufficiently large, accurate rankings and hence a worker-optimal stable matching can still be achieved even under imprecise reward estimations. \cite{liu2020competing} provide a similar observation in their Lemma~3. To this end, we let $$\Delta_{\min} = \min_{i\in[N]}\left\{\min_{j\not =j'}|\bTheta^{*(i,j)}-
\bTheta^{*(i,j')}|\right\}$$ denote the minimum gap of rewards over all workers $i\in[N].$ Then, provided that the infinity-norm estimation error $\|\widetilde{\bTheta}-\bTheta^*\|_{\infty}\leq \Delta_{\min}/2$ where $\widetilde{\bTheta}$ is obtained from Algorithm \ref{alg: lowrank_competing_bandit}, we can ensure that the stable matching we commit to, i.e., $\bX_c$, coincides with the worker-optimal stable matching $\bX^*.$ This main idea gives rise to the bound on the stable regret for every worker (as previously defined in \eqref{def: stable_regret}) in Theorem~\ref{theorem: stable_bandit}.

We introduce the following lemma before we prove Theorem~\ref{theorem: stable_bandit}. Its proof is provided in Appendix~\ref{app:proof-mid-step-stable}.
\begin{lemma}\label{lem: stable_Step}
   Given our choice of $\lambda =c_{\lambda} \sigma \log\big((N+K)(3N+3K+5)T\big)/\sqrt{E_h},$ the stable regret of Algorithm~\ref{alg: lowrank_competing_bandit} for worker $i$ satisfies
\begin{equation}
\mathbb{E}[R_i(T)]\leq 2E_h+2(T-E_h)\cdot\indicator\left\{\delta_{\infty}(E_h)>\Delta_{\min}/2\right\}+2,\quad \forall i\in[N],
\label{eq: expected_regret_stable}
\end{equation}
where 
\begin{multline*}
    \delta_{\infty}(E_h)=c_6{\eta^2\kappa^2r^{3/2}}\max\Bigg\{c_{\lambda} \sigma \pmin\log\big[(N+K)(3N+3K+5)T\big]\sqrt{\frac{K}{NE_h}},\\\pmin\sqrt{\frac{rK\log[(N+K)E_h]}{NE_h}},\frac{\eta\sigma\sqrt{\log\big[(3N+3K+5)T\big]}}{\|\bTheta^*\|_{\infty}}\sqrt{\frac{\pmin K}{NE_h}}\Bigg\}
\end{multline*}
for the constant $c_6$ from \eqref{thm: infinity_explicit}.
\label{lemma: mid-step-stable-regret}
\end{lemma}

\subsection{Proof of Theorem~\ref{theorem: stable_bandit}}
\proof{Proof of Theorem~\ref{theorem: stable_bandit}.}
Following Lemma~\ref{lemma: mid-step-stable-regret}, choose
\begin{align*}
    E_h&=c_7\frac{r^3K}{N\Delta_{\min}^2}\max\left\{c_\lambda^2\sigma^2 \log^2\big[(N+K)(3N+3K+5)T\big],r\log\big[(N+K)T\big],\frac{\eta^2\sigma^2\log\big[(3N+3K+5)T\big]}{\|\bTheta
    ^*\|_{\infty}^2}\right\}
\end{align*}
where $c_7\geq 4c_6^2\eta^4\kappa^4$ is a constant and suppose that $T$ is large enough such that $E_h\leq T.$
Then we have
\begin{align*}
    E_h&=c_7\frac{\pmin r^3K}{N\Delta_{\min}^2}\max\left\{c_\lambda^2\sigma^2\pmin \log^2\big[(N+K)(3N+3K+5)T\big],\pmin r\log\big[(N+K)T\big],\frac{\eta^2\sigma^2\log\big[(3N+3K+5)T\big]}{\|\bTheta
    ^*\|_{\infty}^2}\right\}\\
    &\geq \frac{4c_6^2\pmin\eta^4\kappa^4r^3K}{N\Delta_{\min}^2}\max\left\{c_\lambda^2\sigma^2 \pmin\log^2\big[(N+K)(3N+3K+5)T\big],\pmin r\log\big[(N+K)T\big],\frac{\eta^2\sigma^2\log\big[(3N+3K+5)T\big]}{\|\bTheta
    ^*\|_{\infty}^2}\right\}\\
    &\geq \frac{4c_6^2\pmin \eta^4\kappa^4r^3K}{N\Delta_{\min}^2}\max\left\{c_\lambda^2\sigma^2\pmin \log^2\big[(N+K)(3N+3K+5)T\big],\pmin r\log\big[(N+K)E_h\big],\frac{\eta^2\sigma^2\log\big[(3N+3K+5)T\big]}{\|\bTheta
    ^*\|_{\infty}^2}\right\},
\end{align*}
which implies
\begin{equation*}
    \delta_{\infty}(E_h)\leq \frac{\Delta_{\min}}{2}.
\end{equation*}
 By Lemma~\ref{lem: stable_Step}, we have 
 \begin{align*}
     \mathbb{E}[R_i(T)]&\leq 2E_h+2(T-E_h)\cdot\indicator\left\{\delta_{\infty}(E_h)>\Delta_{\min}/2\right\}+1\\
     &=\mathcal{O}\left(\frac{r^3K\max\{\log^2[(N+K)T], r\log[(N+K)T]\}}{N\Delta_{\min}^2}\right),
 \end{align*}
 which completes our proof. \Halmos

\subsection{Proof of Lemma~\ref{lemma: mid-step-stable-regret}}\label{app:proof-mid-step-stable}
\proof{Proof of Lemma~\ref{lemma: mid-step-stable-regret}.}
First notice that, as long as the estimation error $\|\bTheta^*-\widetilde{\bTheta}\|_{\infty}\leq \Delta_{\min}/2,$ we will have $\bX_c = \bX^*.$ This is because, as aforementioned, the Gale-Shapley algorithm only considers the preference rankings from both sides when deriving the worker-optimal stable matching. In other words, as long as the preference rankings derived from $\widetilde{\bTheta}$ are the same as those derived from $\bTheta^*,$ the result returned by Gale-Shapley algorithm will be the same. The condition $\|\bTheta^*-\widetilde{\bTheta}\|_{\infty}\leq \Delta_{\min}/2$ will ensure that each worker's preferences will be the same under $\bTheta^*$ and $\widetilde{\bTheta}.$ Now with this claim, we can bound the regret in the rounds from $E_h+1$ to $T.$  For any $i\in[N]$, we have 
\begin{align*}    &\mathbb{E}\left[\sum_{t=E_h+1}^{T}\Big(\langle \bX^{*i},\bTheta^*\rangle-\langle \bX_t^i,\bTheta^*\rangle\Big)\right]\\
&=\mathbb{E}\left[\sum_{t=E_h+1}^{T}\Big(\langle \bX_c^{i},\bTheta^*\rangle-\langle \bX_t^i,\bTheta^*\rangle\Big)\right]\\
& = \mathbb{E}\left[\sum_{t=E_h+1}^{T}\Big(\langle \bX_c^{i},\bTheta^*\rangle-\langle \bX_t^i,\bTheta^*\rangle\Big)\cdot\indicator\{\bX_c \not =\bX^*\}\right]\\
&= \mathbb{E}\left[\sum_{t=E_h+1}^{T}\Big(\langle \bX_c^{i},\bTheta^*\rangle-\langle \bX_t^i,\bTheta^*\rangle\Big)\cdot\indicator\{\bX_c \not =\bX^*\}\cdot\indicator\{\|\widetilde{\bTheta}-\bTheta^*\|_{\infty}\leq \delta_{\infty}(E_h)\}\cdot\indicator\{\delta_{\infty}(E_h)\leq \Delta_{\min}/2\}\right]\\
&\quad+\mathbb{E}\left[\sum_{t=E_h+1}^{T}\Big(\langle \bX_c^{i},\bTheta^*\rangle-\langle \bX_t^i,\bTheta^*\rangle\Big)\cdot\indicator\{\bX_c \not =\bX^*\}\cdot\indicator\{\|\widetilde{\bTheta}-\bTheta^*\|_{\infty}\leq \delta_{\infty}(E_h)\}\cdot\indicator\{\delta_{\infty}(E_h)> \Delta_{\min}/2\}\right]\\
&\quad+\mathbb{E}\left[\sum_{t=E_h+1}^{T}\Big(\langle \bX_c^{i},\bTheta^*\rangle-\langle \bX_t^i,\bTheta^*\rangle\Big)\cdot\indicator\{\bX_c \not =\bX^*\}\cdot\indicator\{\|\widetilde{\bTheta}-\bTheta^*\|_{\infty}>\delta_{\infty}(E_h)\}\right]\\
&=\mathbb{E}\left[\sum_{t=E_h+1}^{T}\Big(\langle \bX_c^{i},\bTheta^*\rangle-\langle \bX_t^i,\bTheta^*\rangle\Big)\cdot\indicator\{\bX_c \not =\bX^*\}\cdot\indicator\{\|\widetilde{\bTheta}-\bTheta^*\|_{\infty}\leq \delta_{\infty}(E_h)\}\cdot\indicator\{\delta_{\infty}(E_h)> \Delta_{\min}/2\}\right]\\
&\quad+\mathbb{E}\left[\sum_{t=E_h+1}^{T}\Big(\langle \bX_c^{i},\bTheta^*\rangle-\langle \bX_t^i,\bTheta^*\rangle\Big)\cdot\indicator\{\bX_c \not =\bX^*\}\cdot\indicator\{\|\widetilde{\bTheta}-\bTheta^*\|_{\infty}>\delta_{\infty}(E_h)\}\right]\\
&\leq 2(T-E_h)\cdot\indicator\{\delta_{\infty}(E_h)> \Delta_{\min}/2\}+2(T-E_h)\mathbb{P}\left[\|\widetilde{\bTheta}-\bTheta^*\|_{\infty}>\delta_{\infty}(E_h)\right]\\
&\leq 2(T-E_h)\cdot\indicator\{\delta_{\infty}(E_h)> \Delta_{\min}/2\}+2(T-E_h)T^{-1}\leq 2(T-E_h)\cdot\indicator\{\delta_{\infty}(E_h)> \Delta_{\min}/2\}+2
\end{align*}
where
the first inequality is given by $\langle \bX_c^{i},\bTheta^*\rangle-\langle \bX_t^i,\bTheta^*\rangle\leq |\langle \bX_c^{i},\bTheta^*\rangle|+|\langle \bX_t^i,\bTheta^*\rangle|\leq 2$ and the second inequality comes from $\|\widetilde{\bTheta}-\bTheta^*\|_{\infty}\leq \delta_{\infty}(E_h)$ with probability higher than $1-(T)^{-1}$ given Theorem~\ref{theorem: second_stage_error_bound}.
Then we have 
\begin{align*}
    \mathbb{E}\left[R_i(T)\right]&=\mathbb{E}\left[\sum_{t=1}^{E_h}\Big(\langle \bX^{*i},\bTheta^*\rangle-\langle \bX_t^i,\bTheta^*\rangle\Big)\right]+\mathbb{E}\left[\sum_{t=E_h+1}^T \left(\langle \bX^{*i},\bTheta^*\rangle-\langle \bTheta
    \bX_t^i,\bTheta^*\rangle\right)\right]\\
    &\leq \mathbb{E}\left[\sum_{t=1}^{E_h}\Big(|\langle \bX^{*i},\bTheta^*\rangle|+|\langle \bX_t^i,\bTheta^*\rangle|\Big)\right]+\mathbb{E}\left[\sum_{t=E_h+1}^T \left(\langle \bX^{*i},\bTheta^*\rangle-\langle \bX_t^i,\bTheta^*\rangle\right)\right]\\ 
    &\leq 2E_h+2(T-E_h)\cdot\indicator\{\delta_{\infty}(E_h)> \Delta_{\min}/2\}+2,
\end{align*}
which completes the proof.
\Halmos
\endproof

\section{Technical Lemmas}
\label{app:technical}
\begin{lemma}\label{lem: matrix_berstein_inequality}
    Let $\bold{Z}_1,\cdots,\bold{Z}_n$ be i.i.d. matrices in $\mathbb{R}^{N\times K}$ with $\mathbb{E}[\bold{Z}_i]=\bold{0}$ and $\|\bold{Z}_i\|_\op\leq \omega$ almost surely for all $i\in[n]$; let $\sigma_Z$ be a parameter such that
    \begin{equation*}
        \sigma_Z^2\geq \max\left\{\left\|\sum_{i=1}^n\mathbb{E}\left[\bold{Z}_i^\top\bold{Z}_i\right]\right\|_\op,\left\|\sum_{i=1}^n\mathbb{E}\left[\bold{Z}_i\bold{Z}_i^\top\right]\right\|_\op\right\}.
    \end{equation*}
    Then for any $\rho>0,$
    \begin{equation*}
        \mathbb{P}\left[\left\|\sum_{i=1}^n \bold{Z}_i\right\|_\op \geq \rho \right]\leq (N+K)\exp\left[\frac{-\rho^2}{2\sigma_Z^2+(2\omega\rho)/3}\right].
    \end{equation*}
\end{lemma}
\proof{Proof of Lemma~\ref{lem: matrix_berstein_inequality}.}
    See Proposition 1 in \cite{athey2021matrix}.
\Halmos
\begin{lemma}\label{lem: massart_inequality}
    For any $\zeta>0,$ let $$\mathcal{C}(\zeta)=\left\{\bold{A}\in\mathbb{R}^{N\times K} \mid \|\bold{A}\|_\infty\leq 1, \mathbb{E}_\Pi\left[\sum_{t=1}^n\|\bold{A}\circ \bold{X}_t\|_F^2\right]\leq \zeta\right\},$$ where $\bX_t$ is a random matrix following distribution $\Pi_t$ (defined in Section~\ref{sec: problem_formulation}). Let $z_\zeta = \sup_{\bold{A}\in \mathcal{C}(\zeta)}\left|\sum_{t=1}^n\|\bold{A}\circ\bX_t\|_F^2-\mathbb{E}_\Pi[\|\bold{A}\circ \bold{X}\|_F^2]\right|.$ Then we have
    \begin{equation*}
        \mathbb{P}\left[z_\zeta\geq 2\mathbb{E}[z_\zeta]+\frac{7\zeta}{24}\right]\leq \exp\left(\frac{-\zeta}{288N}\right).
    \end{equation*}
\end{lemma}
\proof{Proof of Lemma~\ref{lem: massart_inequality}.}
Our result relies on the Massart's concentration inequality (see Theorem~3 in \cite{massart2000constants}). First note that 
\begin{equation*}
    \|\bold{A}\circ \bold{X}_t\|_F^2\leq N,\forall t\in[n], \forall \bold{A}\in\mathcal{C}(\zeta).
\end{equation*}
Then we need to provide an upper bound for 
$$\sigma_\zeta^2 = \sup_{\bold{A}\in\mathcal{C}(\zeta)}\sum_{t=1}^n\operatorname{Var}(\|\bold{A}\circ \bold{X}_t\|_F^2).$$
We have
\begin{align*}
    \sigma_\zeta^2 &= \sup_{\bold{A}\in\mathcal{C}(\zeta)}\sum_{t=1}^n\operatorname{Var}(\|\bold{A}\circ \bold{X}_t\|_F^2)\\
    &\leq \sup_{\bold{A}\in\mathcal{C}(\zeta)}\sum_{t=1}^n\mathbb{E}[\|\bold{A}\circ \bold{X}_t\|_F^4]\\
    &\leq \sup_{\bold{A}\in\mathcal{C}(\zeta)}\sum_{t=1}^n\mathbb{E}[\|\bold{A}\circ \bold{X}_t\|_F^2]\sup_{\bold{A}\in\mathcal{C}(\zeta)}\sup_{t\in[n]}\mathbb{E}[\|\bold{A}\circ \bold{X}_t\|_F^2]\\
    &\leq N\zeta,
\end{align*}
where the second inequality is given by the definition of variance and the third inequality is given by the definition of $\mathcal{C}(\zeta).$ 
Thus we choose $\varepsilon'=1, \sigma_\zeta = \sqrt{N\zeta},b = N$ and $x=\zeta/(288N)$ in (11) of \cite{massart2000constants}, and we obtain
\begin{align*}
    \mathbb{P}\left[z_\zeta\geq 2\mathbb{E}[z_\zeta]+\frac{7\zeta}{24}\right]\leq \exp\left(\frac{-\zeta}{288N}\right).\Halmos
\end{align*}
\begin{lemma}\label{lem: hadamard-product}
    For any two matrices $\bold{A},\bold{B}$ we have
    \begin{equation*}
        \rank(\bold{A}\circ\bold{B})\leq \rank(\bold{A}) \cdot \rank(\bold{B})
    \end{equation*}
    where $\bold{A}\circ\bold{B}$ denotes the Hadamard product of matrix $\bold{A}$ and $\bold{B}.$
\end{lemma}
\proof{Proof of Lemma~\ref{lem: hadamard-product}.}See Theorem 4.5 of \cite{million2007hadamard}.\Halmos
\endproof
\begin{lemma}
    \label{lem: matrix_rademacher}
    Let $\bold{Z}_1,\cdots,\bold{Z}_n$ be fixed matrices in $\mathbb{R}^{N\times K}$ and $\xi_1,\xi_2,\cdots, \xi_n$ be independent Rademacher random variables; let $\sigma_Z$ be a parameter such that
    \begin{equation*}
        \sigma_Z^2\geq \max\left\{\left\|\sum_{i=1}^n\bold{Z}_i^\top\bold{Z}_i\right\|_\op,\left\|\sum_{i=1}^n\bold{Z}_i\bold{Z}_i^\top\right\|_\op\right\}.
    \end{equation*}
    Then for any $\rho>0,$
    \begin{equation*}
        \mathbb{P}\left[\left\|\sum_{i=1}^n \xi_i\bold{Z}_i\right\|_\op \geq \rho \right]\leq (N+K)\exp\left[\frac{-\rho^2}{2\sigma_Z^2}\right].
    \end{equation*}
\end{lemma}
\proof{Proof of Lemma~\ref{lem: matrix_rademacher}.}
See Theorem 4.1.1 of \cite{tropp2015introduction}. \Halmos
\endproof
\begin{lemma}\label{lem: subgaussian_vector_sequence}
    Consider $n$ fixed vectors $X_1,X_2,\cdots,X_n\in\mathbb{R}^d$ where $\|X_i\|\leq S,\forall i\in[n].$ Then for n i.i.d. $\sigma$-subgaussian random variables $\varepsilon_i,$ we have
    \begin{equation*}
        \Prob{\Big\|\sum_{i=1}^n\varepsilon_iX_i\Big\|\geq  \rho}\leq 2\exp\left(\frac{-\rho^2}{2nd S^2\sigma^2}\right),\forall \rho>0.
    \end{equation*}
\end{lemma}
\proof{Proof of Lemma~\ref{lem: subgaussian_vector_sequence}.}
It is straightforward to see that $\varepsilon_i X_i$ is a $\sigma S$-subgaussian random vector for each $i\in[n].$ Thus $\sum\nolimits_i \varepsilon_i X_i$ is a $\sqrt{n}\sigma S$-subgaussian random vector. By Lemma~\ref{lem: sub_gaussian_concentration}, we immediately have
\begin{equation*}
\Prob{\Big\|\sum_{i=1}^n\varepsilon_iX_i\Big\|\geq  \rho}\leq 2\exp\left(\frac{-\rho^2}{2nd S^2\sigma^2}\right).\Halmos
\end{equation*}

\begin{lemma}[Matrix Chernoff]
\label{lem: matrix_chernoff}
Consider independent positive-semidefinite matrices $\bold{Z}_1, \bold{Z}_2, \cdots,\bold{Z}_k\in\mathbb{R}^{d\times d}$ that satisfy
$$
\|\bold{Z}_i\|_{\op} \leq \omega, \forall i\in[k] \quad \text { almost surely. } 
$$
Then we have
\begin{align*} 
\mathbb{P}\left[\lambda_{\min }\left(\sum_{i=1}^k \bold{Z}_i\right) \leq \rho \mu_{\min }\right] \leq d \cdot \mathrm{e}^{-(1-\rho)^2 \mu_{\min } / 2 \omega}, \forall\rho \in[0,1] 
\end{align*}
and
\begin{align*} 
\mathbb{P}\left[\lambda_{\max }\left(\sum_{i=1}^k \bold{Z}_i\right) \geq \rho \mu_{\max }\right] \leq d \cdot\left[\frac{\mathrm{e}}{\rho}\right]^{\rho \mu_{\max } / \omega}, \forall \rho \geq \mathrm{e},
\end{align*}
where
$$
\mu_{\min }=\lambda_{\min }\left(\sum_{i=1}^k \mathbb{E} \bold{Z}_i\right) \quad \text{and} \quad \mu_{\max }=\lambda_{\max }\left(\sum_{i=1}^k \mathbb{E} \bold{Z}_i\right) .
$$
\end{lemma}
\proof{Proof of Lemma~\ref{lem: matrix_chernoff}.}
    See Remark~5.3 in \cite{tropp2012user}.
\Halmos

\begin{lemma}[Weyl's Inequality]\label{lem:Weryl's theorem}
    For any matrix $\bold{A},\bold{A}' \in \mathbb{R}^{N\times K}$, it holds for all $i\in[\min\{N,K\}]$ that
    \begin{equation*}
        |\sigma_i(\bold{A})-\sigma_i(\bold{A}')|\leq \|\bold{A}-\bold{A}'\|_{\op}.
    \end{equation*}
\end{lemma}
\proof{Proof of Lemma~\ref{lem:Weryl's theorem}.}
See \cite{weyl1912asymptotische}.
\Halmos
\endproof

\begin{lemma}\label{lem: algebra_inequalities} 
For any $\bTheta^*\in\mathbb{R}^{N\times K}$ with SVD $\bTheta^*=\bU^*\bD^*\bV^{*\top},$ we have
    \begin{equation}\label{eq: UDtwoinfty}
\|\bU^*\bD^*\|_{2,\infty}=\|\bTheta^*\|_{2,\infty}.
    \end{equation}
Further, if $\bTheta^*$ satisfies Assumption~\ref{assump: spikiness}, we have
\begin{equation}\label{ineq: sigmamin_lower}
\sigma_{\min}(\bTheta^*)=\frac{\|\bTheta^*\|_\op}{\kappa}\geq \frac{\|\bTheta^*\|_F}{\sqrt{r}\kappa}\geq \frac{\sqrt{NK}\|\bTheta^*\|_\infty}{\sqrt{r}\eta\kappa},
    \end{equation}
where $\kappa$ is the conditional number of $\bTheta^*$.
\end{lemma}
\proof{Proof of Lemma~\ref{lem: algebra_inequalities}.}
    For \eqref{eq: UDtwoinfty}, notice that
\begin{align*}
    \|\bU^*\bD^*\|_{2,\infty}  = \max_{i\in[N]}\|e_i(N)^\top \bU^*\bD^*\|
    =\max_{i\in[N]}\|e_i(N)^\top \bU^*\bD^*\bV^{*\top}\|
    = \max_{i\in[N]}\|e_i(N)^\top \bTheta^*\|
    = \|\bTheta^*\|_{2,\infty},
\end{align*}
where the second equation is given by that $\bV^*$ is orthogonal and multiplying a vector by an orthogonal matrix does not change its length.

For \eqref{ineq: sigmamin_lower}, the first equation and the first inequality are straightfoward. The last inequality directly follows Assumption~\ref{assump: spikiness}, i.e., 
\begin{align*}
    \|\bTheta^*\|_F\geq \frac{\sqrt{NK}\|\bTheta^*\|_{\infty}}{\eta}.\Halmos
\end{align*}
\endproof

\begin{lemma}\label{lem: semi-ortho matrix}
Let $\bV\in\mathbb{R}^{K\times r}$ be an orthogonal matrix (i.e., $\bV^\top\bV=\bold{I}_{r \times r}$). We have for any matrix $\bold{A}\in\mathbb{R}^{r\times K},$ $\|\bV\bold{A}\|_\op\leq\|\bold{A}\|_\op=\|\bold{A}\bV\|_\op.$ 
\end{lemma}
\proof{Proof of Lemma~\ref{lem: semi-ortho matrix}.}
By the property of operator norm, we have
\begin{align*}
    \|\bV\bold{A}\|_\op&\leq \|\bV\|_\op\|\bold{A}\|_\op
    =\|\bold{A}\|_\op.
\end{align*}

    Let the SVD of $\bold{A}$ be $\bold{A}=\bU_A\bold{D}\bV_A,$ where $\bV_A$ is an $r\times r$ orthogonal matrix. Thus we have $\bold{A}\bV=\bU_A\bold{D}(\bV_A\bV),$ which is a SVD since both $\bU_A$ and $\bV_A\bV$ are orthogonal matrices. Hence the singular values of $\bold{A}$ and $\bold{A}\bV$ are the same, which yields 
    $\|\bold{A}\|_\op=\|\bold{A}\bV\|_\op.$
\Halmos

\begin{lemma}\label{lem:optimal_rotation}
    For any rank $r$ matrices $\bTheta,{\nuclearest}\in\mathbb{R}^{N\times K},$ let ${\nuclearest}=\nuclearestU\widehat{\bD}{\nuclearestV}^\top$  be an arbitrary SVD of ${\nuclearest},$ then there exists SVD  $\bTheta=\bU\bold{D}\bV^\top,$ where
    \begin{equation*}
        \|\bU-\nuclearestU\|_F\leq 5\left(\frac{1}{\sigma_{r}(\bTheta)}+\frac{1}{\sigma_r({\nuclearest})}\right)\|\bTheta-{\nuclearest}\|_F
    \end{equation*}
    and
    \begin{equation*}
        \|\bV-{\nuclearestV}\|_F\leq 5\left(\frac{1}{\sigma_{r}(\bTheta)}+\frac{1}{\sigma_r({\nuclearest})}\right)\|\bTheta-{\nuclearest}\|_F.
    \end{equation*}
\end{lemma}
    \proof{Proof of Lemma~\ref{lem:optimal_rotation}.} In this proof, let $\bTheta=\bU\bold{D}\bV^\top$ be an arbitrary SVD of $\bTheta.$ Moreover, let 
    \begin{equation*}
        \bold{F'}=\begin{bmatrix}
            \bold{U'}\\
            \bold{V'}
        \end{bmatrix}\quad\text{and}\quad\bold{F}=\begin{bmatrix}
            \bU\\
            \bV
        \end{bmatrix}.
    \end{equation*}
    Then it remains to bound
    \begin{equation*}
d(\bold{F},\bold{F}')=\min_{\bold{Q}\in\mathcal{O}_r}\|\bold{F}-\bold{F}'\bold{Q}\|_F= \min_{\bold{Q}\in\mathcal{O}_r}\|\bold{F}\bQ^{\top}-\bold{F}'\|_F.
    \end{equation*}
    By Remark 6.1 in \cite{keshavan2010matrix}, we have \begin{equation*}
        d(\bold{F},\bold{F}')\leq \sqrt{2}\|\bold{F}'\bold{F}'^\top-\bold{F}\bold{F}^\top\|_F=\sqrt{2}(\|\nuclearestU\nuclearestU^\top-\bU\bU^\top\|_F+\|{\nuclearestV}{\nuclearestV}^\top-\bV\bV^\top\|_F+2\|\nuclearestU{\nuclearestV}^\top-\bU\bV^\top\|_F).
    \end{equation*}

    By Lemma~\ref{lem: hamidi}, we have that
    \begin{equation*}
        \|\nuclearestU\nuclearestU^\top-\bU\bU^\top\|_F\leq\frac{\sqrt{2}\|\bTheta-{\nuclearest}\|_F}{\sigma_{r}(\bTheta)}\quad\text{and}\quad\|{\nuclearestV}{\nuclearestV}^\top-\bV\bV^\top\|_F\leq \frac{\sqrt{2}\|\bTheta-{\nuclearest}\|_F}{\sigma_{r}({\nuclearest})}.
    \end{equation*}
    Furthermore, notice that $\nuclearestU{\nuclearestV}^\top=\mathrm{sgn}({\nuclearest})$ and $\bU\bV^\top=\mathrm{sgn}(\bTheta)$, which are defined in Lemma~\ref{lem: sign_matrix}. Then by Lemma~\ref{lem: sign_matrix}, we have
    \begin{equation*}
        \|\nuclearestU{\nuclearestV}^\top-\bU\bV^\top\|_F\leq \frac{3}{2}\left(\frac{1}{\sigma_{r}(\bTheta)}+\frac{1}{\sigma_r({\nuclearest})}\right)\|\bTheta-{\nuclearest}\|_F.
    \end{equation*}
    Thus
    \begin{align*}
        d(\bold{F},\bold{F}')&\leq \left(\frac{3\sqrt{2}}{2}+2\right)\left(\frac{1}{\sigma_{r}(\bTheta)}+\frac{1}{\sigma_r({\nuclearest})}\right)\|\bTheta-{\nuclearest}\|_F\\
        &\leq 5\left(\frac{1}{\sigma_{r}(\bTheta)}+\frac{1}{\sigma_r({\nuclearest})}\right)\|\bTheta-{\nuclearest}\|_F,
    \end{align*}
    which completes the proof.
\Halmos

\begin{lemma}\label{lem: matrix_pertubation}
For any rank $r$ matrices $\bTheta,{\nuclearest}\in\mathbb{R}^{N\times K},$ we have
   \begin{equation*}
      \big| \sigma_{\min}(\bTheta)-\sigma_{\min}({\nuclearest})\big|\leq \|\bTheta-{\nuclearest}\|_F.
   \end{equation*} 
\end{lemma}
\proof{Proof of Lemma~\ref{lem: matrix_pertubation}.}
    By Lemma~\ref{lem:Weryl's theorem}, we have
    \begin{equation*}
        \big| \sigma_{r}(\bTheta)-\sigma_{r}({\nuclearest})\big|\leq \|\bTheta-{\nuclearest}\|_\op \leq \|\bTheta-{\nuclearest}\|_F.\Halmos
    \end{equation*}

\begin{lemma}\label{lem: row-bound}
For any matrix $\bTheta\in\mathbb{R}^{N\times K},$ let the SVD of $\bTheta=\bU\bold{D}\bV^\top,$ we have
\begin{align*}
    \|\bU\|_{2,\infty}\leq \frac{\sqrt{K}\|\bTheta\|_{\infty}}{\sigma_{\min}(\bTheta)}
\quad\text{and}\quad
    \|\bV\|_{2,\infty}\leq \frac{\sqrt{N}\|\bTheta\|_{\infty}}{\sigma_{\min}(\bTheta)}.
\end{align*}
\end{lemma}
\proof{Proof of Lemma~\ref{lem: row-bound}.}
We have 
\begin{equation*}
    \|\bU\|_{2,\infty}\sigma_{\min}(\bTheta)\leq \|\bTheta\|_{2,\infty}\leq\sqrt{K} \|\bTheta\|_{\infty},
\end{equation*}
which gives us the first inequality. The second inequality can be obtained via an analogous argument.
\Halmos
\endproof
\begin{lemma}\label{lem: matrix_inverse}
    For any positive definite matrices $\bold{A},\bold{D}\in\mathbb{R}^{r\times r}$ where $\bold{D}$ is a diagonal matrix, we have
    \begin{equation*}
        \|\bold{A}^{-1}-\bold{D}^{-1}\|_{\op} \leq \frac{\max_{i\in[r]}|\sigma_i(\bold{A})-\bold{D}_{ii}|} {\sigma_{\min}(\bold{A})\sigma_{\min}(\bold{D})}
    \end{equation*}
    \end{lemma}
\proof{Proof of Lemma~\ref{lem: matrix_inverse}.}
        For any positive definite matrix $\bold{A},$ there exists an orthogonal matrix $\bold{Q}$ and a diagonal matrix $\bold{D}_A$ such that
        \begin{equation*}
            \bold{A}=\bold{Q}\bold{D}_A\bold{Q}^\top,
        \end{equation*}
        where the $i^{\tth}$ element on the diagonal of $\bold{D}_A$ is $\sigma_i(\bold{A}).$ Thus we have 
        $\bold{A}^{-1}=\bold{Q}\bold{D}_A^{-1}\bold{Q}^\top,$ and
        \begin{align*}
            \bold{A}^{-1}-\bold{D}^{-1}&=\bold{Q}\bold{D}_A^{-1}\bold{Q}^\top-\bold{D}^{-1}\\
            &=\bold{Q}\bold{D}_A^{-1}\bold{Q}^\top-\bold{Q}\bold{D}^{-1}\bold{Q}^\top\\
            &= \bold{Q}(\bold{D}_A^{-1}-\bold{D}^{-1})\bold{Q}^\top.
        \end{align*}
        By the definition of operator norm, we know that
        \begin{align*}
            \|\bold{A}^{-1}-\bold{D}^{-1}\|_\op &= \max_{i\in[r]}|\bold{D}_{A}^{-1}(i,i)-\bold{D}^{-1}(i,i)|\\
            &= \max_{i\in[r]}|\sigma_i(\bold{A})^{-1}-\bold{D}_{ii}^{-1}|\\
            &= \max_{i\in[r]}\left|\frac{\bold{D}_{ii}-\sigma_i(\bold{A})}{\sigma_i(\bold{A})\bold{D}_{ii}}\right|\\
            &\leq \frac{\max_{i\in[r]}|\sigma_i(\bold{A})-\bold{D}_{ii}|} {\sigma_{\min}(\bold{A})\sigma_{\min}(\bold{D})}.\Halmos
        \end{align*}
        
\begin{lemma}\label{lem: 2inftynorm}
    For any matrices $\bold{A},\bold{B},\bold{C}$, we have
    \begin{equation*}
        \|\bold{A}\bold{B}\bold{C}^\top\|_{\infty}\leq\| \bold{AB}\|_{2,\infty}\|\bold{C}\|_{2,\infty}\leq \|\bold{A}\|_{2,\infty}\|\bold{B}\|_\op\|\bold{C}\|_{2,\infty}.
    \end{equation*}
\end{lemma}
\proof{Proof of Lemma~\ref{lem: 2inftynorm}.}
The first inequality is straightforward. Here we show that 
\begin{equation*}
    \|\bold{A}\bold{B}\|_{2,\infty}\leq \|\bold{A}\|_{2,\infty}\|\bold{B}\|_\op
\end{equation*}
for any matrices $\bold{A}$ and $\bold{B}.$ Let $d_1$ denote the row-dimension of matrix $\bold{A}$ and we have
\begin{align*}
    \|\bold{A}\bold{B}\|_{2,\infty} = \max_{i\in[d_1]}\|e_i(d_1)^\top \bold{A}\bold{B}\|
    \leq \max_{i\in[d_1]}\|e_i(d_1)^\top \bold{A}\|\|\bold{B}\|_\op
    =\|\bold{A}\|_{2,\infty}\|\bold{B}\|_\op.\Halmos
\end{align*}
\endproof

\begin{lemma}\label{lem: sub_gaussian_concentration}
For any $\sigma$-subgaussian random vector $X\in\mathbb{R}^d$, we have
\begin{equation*}
    \Prob{\|X\|\geq \rho}\leq 2\exp\left(\frac{-\rho^2}{2d\sigma^2}\right),\forall \rho>0.
\end{equation*}
    
\end{lemma}
\proof{Proof of Lemma~\ref{lem: sub_gaussian_concentration}.}
    See Lemma~1 in \cite{jin2019short}.
\Halmos
\endproof

\begin{lemma}\label{lem: hamidi}
    For any rank $r$ matrices $\bTheta,{\nuclearest}\in\mathbb{R}^{N\times K},$ let ${\nuclearest}=\nuclearestU\widehat{\bold{D}}{\nuclearestV}^\top$ and $\bTheta=\bU\bD\bV^{\top}$ be their SVDs. Then we have
    \begin{equation*}
        \|\nuclearestU\nuclearestU^\top-\bU\bU^\top\|_F\leq\frac{\sqrt{2}\|\bTheta-{\nuclearest}\|_F}{\sigma_{r}(\bTheta)}.
    \end{equation*}
\end{lemma}
\proof{Proof of Lemma~\ref{lem: hamidi}.}
First notice that
\[\|\nuclearestU\nuclearestU^\top-\bU\bU^\top\|_F=\sqrt{2}\cdot \inf_{\bQ\in\mathbb{R}^{r\times r}}\|\bU-\nuclearestU\bQ\|_F.\]
To show this equality, we first note that $\|\bU-\nuclearestU\bQ\|_F$ is minimized when $\bQ=\nuclearestU^\top \bU.$ Thus we have
\begin{align*}
    \inf_{\bQ\in\mathbb{R}^{r\times r}}\|\bU-\nuclearestU\bQ\|_F&=\|\bU-\nuclearestU\nuclearestU^\top \bU\|_F\\
    &= \sqrt{\operatorname{tr}\left(\left(\bU-\nuclearestU\nuclearestU^\top \bU\right)\left(\bU-\nuclearestU\nuclearestU^\top \bU\right)^\top\right)}\\
    &=\sqrt{\operatorname{tr}\left(\bU\bU^\top\right)-2\operatorname{tr}\left(\nuclearestU\nuclearestU^\top\bU\bU^\top\right)+\operatorname{tr}\left(\nuclearestU\nuclearestU^\top\bU\bU^\top\nuclearestU\nuclearestU^\top\right)}\\
    &=\sqrt{\operatorname{tr}\left(\bU\bU^\top\right)-\operatorname{tr}\left(\nuclearestU\nuclearestU^\top\bU\bU^\top\right)}\\
    &=\sqrt{r-\operatorname{tr}\left(\nuclearestU\nuclearestU^\top\bU\bU^\top\right)}.
\end{align*}
On the other hand, we have
\begin{align*}
    \|\nuclearestU\nuclearestU^\top-\bU\bU^\top\|_F&=\sqrt{\operatorname{tr}\left(\left(\nuclearestU\nuclearestU^\top-\bU\bU^\top\right)\left(\nuclearestU\nuclearestU^\top-\bU\bU^\top\right)^\top\right)}\\
    &=\sqrt{2r-2\operatorname{tr}\left(\nuclearestU\nuclearestU^\top\bU\bU^\top\right)}.
\end{align*}
With this inequality, now we can derive
\begin{align*}
    \|\nuclearestU\nuclearestU^\top-\bU\bU^\top\|_F&=\sqrt{2}\cdot \inf_{\bQ\in\mathbb{R}^{r\times r}}\|\bU-\nuclearestU\bQ\|_F\\
    &=\sqrt{2}\cdot \inf_{\bQ\in\mathbb{R}^{r\times r}}\|\bU^\top-\bQ^\top\nuclearestU^\top\|_F\\
    &= \sqrt{2}\cdot \inf_{\bQ\in\mathbb{R}^{r\times r}}\|\bD^{-1}(\bD\bU^\top-\bD\bQ^\top\nuclearestU^\top)\|_F\\
    &\leq  \sqrt{2}\cdot \|\bD^{-1}\left(\bD\bU^\top-\bD(\bD^{-1}\bV^\top\nuclearestV\widehat{\bold{D}})\nuclearestU^\top\right)\|_F\\
    &=\sqrt{2}\cdot \|\bD^{-1}\left(\bV^\top\bV\bD\bU^\top-\bV^\top\nuclearestV\widehat{\bold{D}}\nuclearestU^\top\right)\|_F\\
    &=\sqrt{2}\cdot \left\|\bD^{-1}\bV^\top\left(\bV\bD\bU^\top-\nuclearestV\widehat{\bold{D}}\nuclearestU^\top\right)\right\|_F\\
    &\leq \sqrt{2}\left\|\bV\bD\bU^\top-\nuclearestV\widehat{\bold{D}}\nuclearestU^\top\|_F\right\|\bD^{-1}\bV^\top\|_\op\\
    &\leq \frac{\sqrt{2}\|\bTheta-\nuclearest\|_F}{\sigma_{\min}(\bTheta)},
\end{align*}
which completes our proof. \Halmos
\endproof

\begin{lemma}\label{lem: sign_matrix}
    For any rank $r$ matrices $\bTheta,{\nuclearest}\in\mathbb{R}^{N\times K}, $ we have
    \begin{equation*}
        \|\mathrm{sgn}(\bTheta)-\mathrm{sgn}({\nuclearest})\|_F\leq\frac{3}{2}\left(\frac{1}{\sigma_{r}(\bTheta)}+\frac{1}{\sigma_r({\nuclearest})}\right)\|\bTheta-{\nuclearest}\|_F
    \end{equation*}
    where $\mathrm{sgn}(\cdot)$ denotes the matrix sign function, i.e., $\mathrm{sgn}(\bTheta)=\bU\bV^\top$ for a matrix $\bTheta$ with SVD $\bU\bold{D}\bV^\top.$
\end{lemma}
    \proof{Proof of Lemma~\ref{lem: sign_matrix}.}
    See Theorem~2.1 of \cite{li2006some}.
    \Halmos
\endproof

\section{Experiment Details}
\label{app: experiment_details}

\subsection{Synthetic Data} 
\label{app: synthetic_exp}

We create our synthetic data through the following steps. In particular, we consider $N = 100$ worker types and $K = 100$ job types --- i.e., there are $N\times K = 10,000$ number of worker-job pairs. For each of the 50 trials, we generate the ground-truth matching reward matrix\footnote{For the online stable matching experiment, $\bTheta^*$ represents the worker reward matrix that indicates worker preference rankings.} $\bTheta^*\in \mathbb{R}^{N\times K}$ with rank $r=3$ as follows. First, we create two matrices $\bU^*\in \mathbb{R}^{N\times r}$ and $\bV^*\in \mathbb{R}^{K\times r}$, of which the entries are independently drawn from the uniform distribution on $[0,1]$. Then, we multiply the two matrices together and obtain $\bTheta^* = \bU^*\bV^{*\top}$.

\paragraph{Offline.} We simulate $n$ number of offline matching samples by drawing the matchings $\bX_t$'s uniformly at random from the set of all matchings $\mathcal{M}$. $n$ takes the values $20, 40, 60, 80, 100$ respectively. Each matching sample $t\in [n]$ contains $N$ noisy rewards from its $N$ matched pairs following \eqref{model: data-generating}. We draw the noises $\varepsilon_t^{(i)}$'s i.i.d. from a Gaussian distribution $\mathcal{N}(0,\sigma^2)$ with $\sigma^2 = 0.1$. For our algorithm, we tune $c_\lambda$ (i.e., a hyperparameter of $\lambda$ stated in Theorem \ref{theorem:errorbound}) over a predefined grid $\{3 \times 10^{-5}, 1 \times 10^{-4}, 3 \times 10^{-4}, 1 \times 10^{-3}, 3 \times 10^{-3}, 1 \times 10^{-2}, 3 \times 10^{-2}, 1 \times 10^{-1}, 3 \times 10^{-1}, 1\}$, and select \( c_\lambda = 1\times 10^{-3}.\)

\paragraph{Online.} We set our total time horizon $T$ equal to $200, 400, 600, 800, 1000$ respectively. For the online optimal matching experiments, we set the number of exploration steps in our \textsf{CombLRB} to be $E_h=q T^{2/3}$ as advised by Theorem \ref{theorem: optimal_matching_bandit} for some hyperparameter $q$. We tune the hyperparameter $q$ and choose $q=1$; $\lambda$ is set in the same way as the offline setting. The implementations of both \textsf{CUCB} and \textsf{CTS} algorithms follow that described in Section 6 of \cite{cuvelier2021statistically}. 

For the online stable matching experiments, we generate the job preference rankings over workers through the matrix $\bPhi^*$. Each column of $\bPhi^*$ is a random permutation of $[N]$. 
For the 50 trials, we draw 10 such $\bPhi^*$ independently and then run 5 independent trials with each $\bPhi^*$. 
We set the length of exploration phase for our $\textsf{CompLRB}$ to be $E_h=q\log T$ for some hyperparameter $q$. We tune the hyperparameter $q$ and choose $q=40$.
$\lambda$ is set in the same way as the offline setting.

\subsection{Real Data of Labor Market}
\label{app: real_experiment_details}

We use an individual-level workforce dataset provided by Revelio Labs, which includes comprehensive employment histories of individuals, including their roles, skills, activities, education, seniority, geographic location, etc. For our analysis, we focus on software engineers with mid-level seniority employed in the United States between 2010 and 2015. This subset contains 684,696 observations from 468,807 software engineers employed by 89,365 different companies.

We group the engineers by their education background. Specifically, engineers who graduate from the same school are put in the same group. 
We group all the 37,926 schools into 236 clusters by their sizes, approximated by the total number of the graduates (i.e., employees) observed in the data. Schools with similar sizes are grouped into the same cluster.
Similarly, companies with similar sizes are grouped into 100 clusters.
 
We finally keep the top $N=50$ engineer clusters and $K=50$ company clusters with the highest number of observations to leave out pairs with missing data and very few observations. This procedure results in a $50\times 50$ ground-truth matrix $\bTheta^*$, where each entry represents the empirical probability that an engineer from a given cluster stays in a company from a corresponding cluster for more than six months.

Our matching observations are generated following \eqref{model: data-generating}, where the noises are i.i.d. Gaussian with mean $0$ and variance $0.1$. The variance is set to be the empirical variance across all entries. Similar to our synthetic experiments, we tune the hyperparameter $c_\lambda$ over the same predefined grid as our synthetic experiments and choose $c_\lambda = 3\times 10^{-3}.$ 
For online stable matching experiment, the job preference ranking matrix $\bPhi^*$ is generated in the same way as our synthetic experiment.

\end{APPENDICES}
\end{document}